\RequirePackage{fix-cm}
\documentclass[twocolumn]{svjour3}          
\smartqed  

\usepackage{amsmath}
\usepackage{graphics}
\usepackage{epsfig}
\usepackage{threeparttable}
\usepackage{xspace}
\usepackage{siunitx}
\usepackage{times}
\usepackage{graphicx}
\usepackage{subfig}
\usepackage{placeins}
\usepackage{blindtext}
\usepackage{amssymb}
\usepackage{threeparttable}
\usepackage{color}
\usepackage[normalem]{ulem}
\usepackage{multirow}
\usepackage{float}
\usepackage{array}
\usepackage{tabulary}	
\usepackage{microtype}
\usepackage{booktabs}
\usepackage{xspace}
\usepackage[table]{xcolor}
\usepackage{colortbl}
\usepackage{url}
\usepackage{cite}

\usepackage[pagebackref=true,breaklinks=true,colorlinks,citecolor=citecolor,bookmarks=false]{hyperref}

\newcommand{\tabincell}[2]{\begin{tabular}{@{}#1@{}}#2\end{tabular}}
\makeatletter
\newcommand{\thickhline}{%
    \noalign {\ifnum 0=`}\fi \hrule height 1pt
    \futurelet \reserved@a \@xhline
}

\DeclareRobustCommand\onedot{\futurelet\@let@token\@onedot}
\def\@onedot{\ifx\@let@token.\else.\null\fi\xspace}

\definecolor{citecolor}{HTML}{0071bc}
\definecolor{scorered}{HTML}{e4485a}
\definecolor{scoreblue}{HTML}{4a7ee8}
\definecolor{scoregreen}{HTML}{80ba0e}

\definecolor{purple0}{HTML}{e9e9f3}
\definecolor{purple}{HTML}{dcdaed}
\definecolor{purple1}{HTML}{bab6da}

\def\eg{\emph{e.g}\onedot} 
\def\ie{\emph{i.e}\onedot} 
 
\def\etc{\emph{etc}\onedot} 
 
\def\etal{\emph{et al}\onedot}

\makeatother

\journalname{IJCV}
\begin{document}
\title{A Survey on Long-tailed Visual Recognition}

\author{Lu Yang$*$, He Jiang$*$, Qing Song$\dagger$, Jun Guo}

\institute{Lu Yang, Beijing University of Posts and Telecommunications (soeaver@bupt.edu.cn) \\
           He Jiang, Beijing University of Posts and Telecommunications (JiangHe@bupt.edu.cn) \\
           Qing Song, Beijing University of Posts and Telecommunications (priv@bupt.edu.cn) \\
           Jun Guo, Beijing University of Posts and Telecommunications (guojun@bupt.edu.cn) \\
           $*$ Equal Contribution \\
           $\dagger$ Corresponding author: Qing Song
}

\date{Received: 11 November 2021 / Accepted: 21 April 2022}

\maketitle

\begin{abstract}
The heavy reliance on data is one of the major reasons that currently limit the development of deep learning. Data quality directly dominates the effect of deep learning models, and the long-tailed distribution is one of the factors affecting data quality. The long-tailed phenomenon is prevalent due to the prevalence of power law in nature. In this case, the performance of deep learning models is often dominated by the head classes while the learning of the tail classes is severely underdeveloped. In order to learn adequately for all classes, many researchers have studied and preliminarily addressed the long-tailed problem. In this survey, we focus on the problems caused by long-tailed data distribution, sort out the representative long-tailed visual recognition datasets and summarize some mainstream long-tailed studies. Specifically, we summarize these studies into ten categories from the perspective of representation learning, and outline the highlights and limitations of each category. Besides, we have studied four quantitative metrics for evaluating the imbalance, and suggest using the Gini coefficient to evaluate the long-tailedness of a dataset. Based on the Gini coefficient, we quantitatively study 20 widely-used and large-scale visual datasets proposed in the last decade, and find that the long-tailed phenomenon is widespread and has not been fully studied. Finally, we provide several future directions for the development of long-tailed learning to provide more ideas for readers.
\end{abstract}

\section{Introduction}
\label{sec:intro}

\begin{figure}
	\begin{center}
		\includegraphics[width=0.9\linewidth]{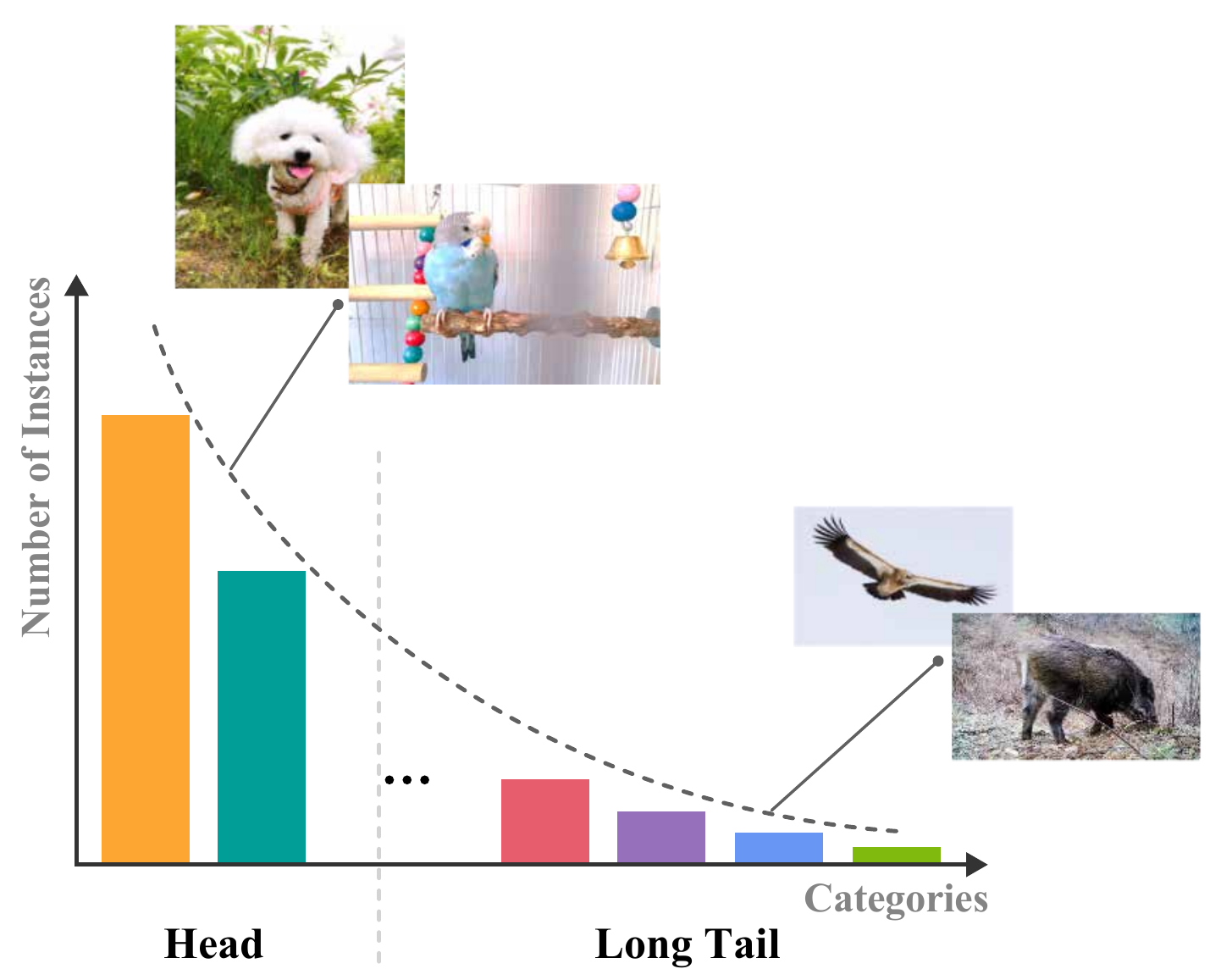}
	\end{center}
	\vspace{-1mm}
	\caption{\textbf{Distribution of Long-tailed dataset}. In nature, there are cases where a few individuals make a large contribution and data tend to show a long-tailed distribution. For example, dog and budgie are common classes, while most other classes such as alpine vulture, tetra are uncommon classes.}
	\vspace{-1mm}
	\label{title}
\end{figure}

\emph{``From politics to public relations, from music scores to college sports, the long tail is everywhere.''}
\begin{flushright}
      -- Chris Anderson, The Long Tail~\cite{chris2006longtail}
\end{flushright}

The advent of deep neural networks has led to remarkable breakthroughs in many fields such as computer vision~\cite{girshick2015fast, wang2018video, shaham2019singan, li2019learning, kim2020detecting}, natural language processing~\cite{lample2018phrase, lan2019albert, jacob2019bert}, and reinforcement learning~\cite{van2019perspective}. However, deep learning models learn features from large amounts of data, and thus inevitably have a heavy dependence on it. Therefore, deep learning faces the challenges brought by the existence of problems in the data itself.

\begin{figure}
	\begin{center}
		\includegraphics[width=0.98\linewidth]{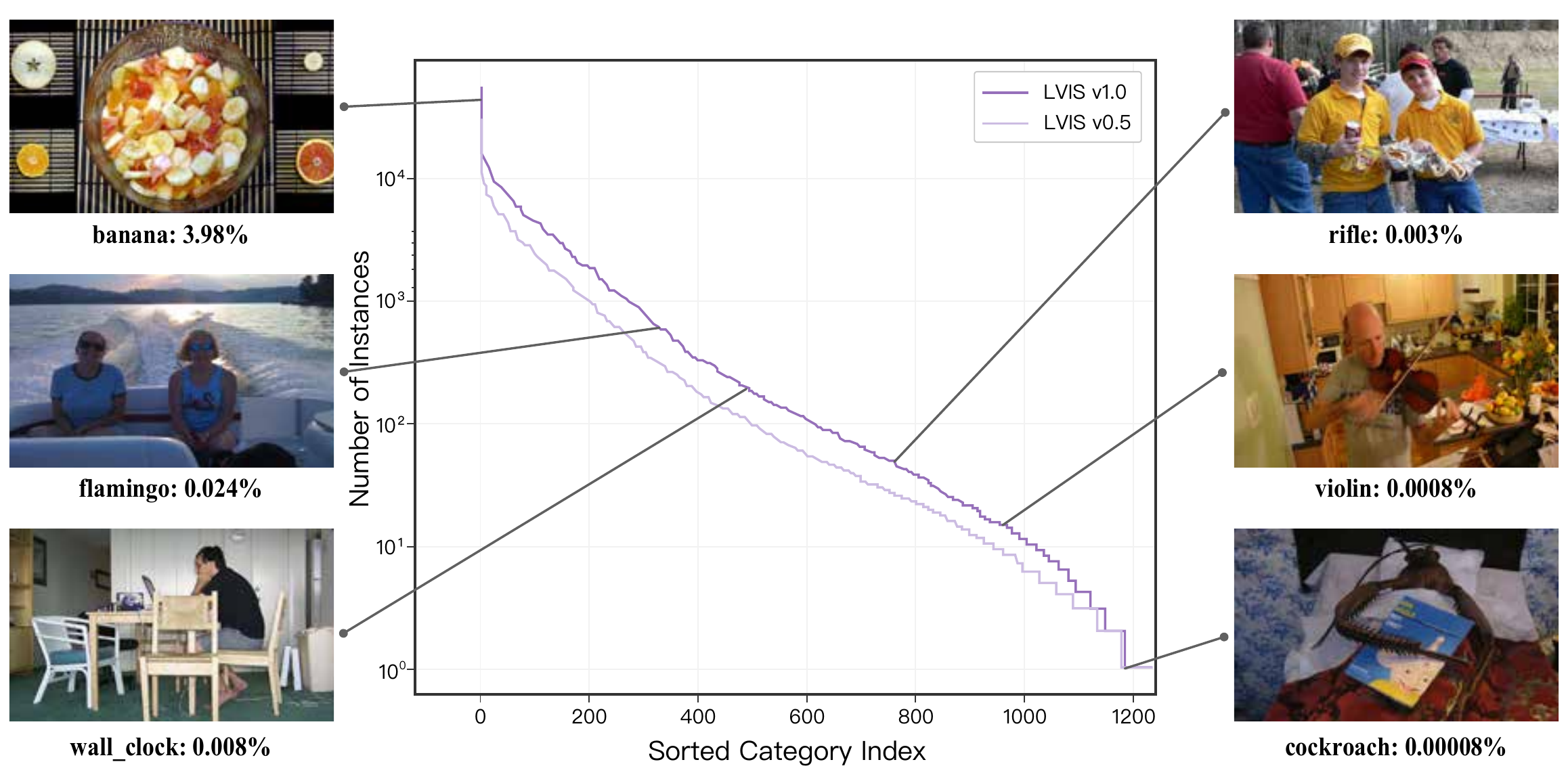}
		\includegraphics[width=0.90\linewidth]{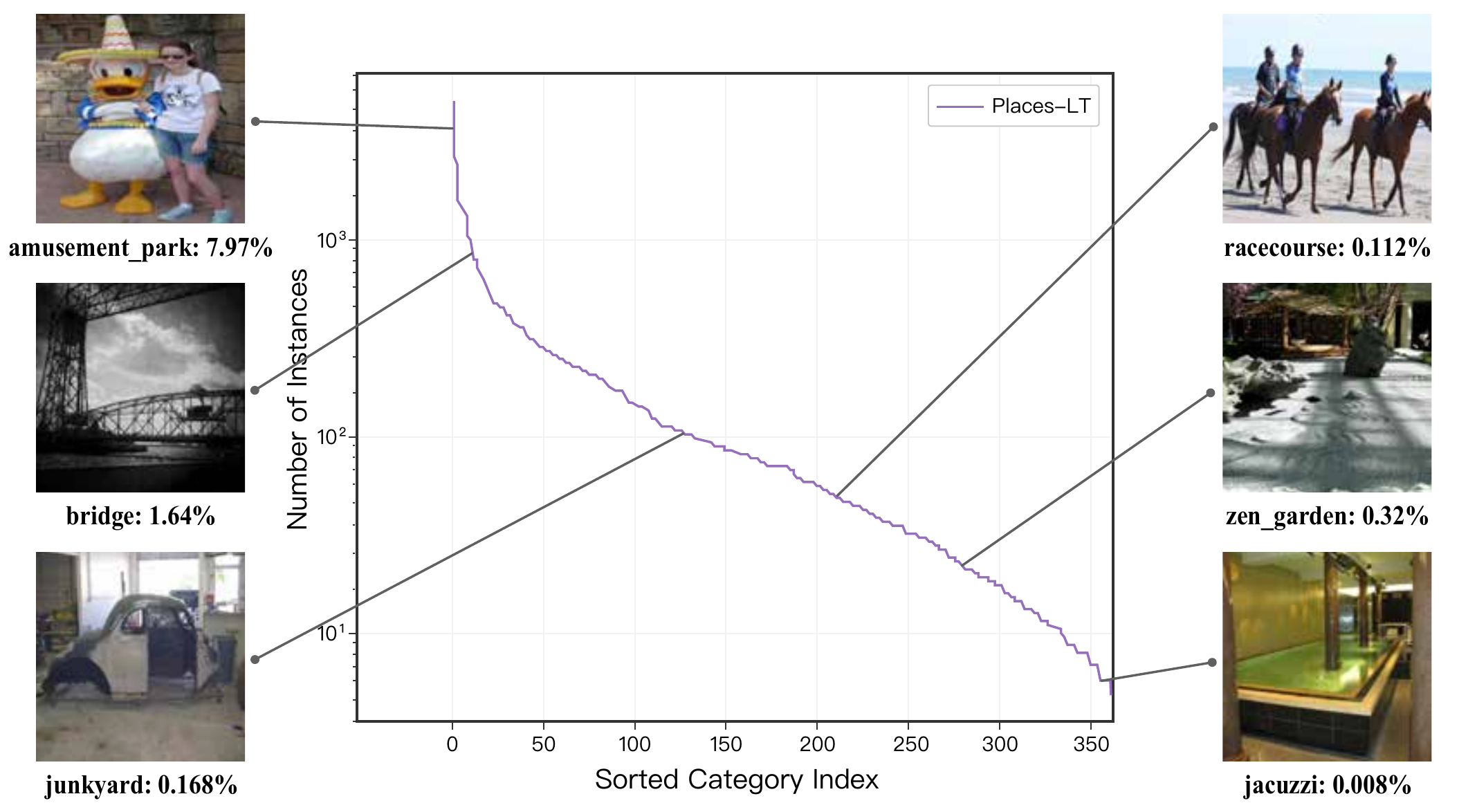}
	\end{center}
	\vspace{-1mm}
	\caption{\textbf{Some common long-tailed distribution datasets as well as their long-tailedness}. \textbf{Top}: examples of LVIS~\cite{gupta2019lvis}, which is a large-scale fine-grained vocabulary annotation dataset for instance segmentation. \textbf{Bottom}: examples of Places-LT~\cite{liu2019large}, which is a scene-centric recognition dataset.}
	\vspace{-1mm}
	\label{data_smples}
\end{figure}

In real life, there exists a distribution of random variables that is more extensive than the positive-terrestrial distribution, \ie the long-tailed distribution. It is mainly reflected in the fact that a small number of individuals usually make a large number of contributions, where a few classes occupy the majority of the dataset (\ie head classes), while the majority of classes have very little data samples (\ie tail classes), as shown in Fig.~\ref{title}. Long-tailed distribution can be reflected in many cases. For example, in the field of economics where the long-tailed theory first emerged, head and tail are used to distinguish between red ocean markets and blue ocean markets. In business sales, "best-selling goods" have fewer classes but high sales volume, which belonging to the head classes, while the "cold goods" have a huge variety, but the sales volume of each class is low, which belonging to the tail classes. In visual recognition, there are also many subfields that involve long-tailed problems, such as instance segmentation, scene classification, \etc as shown in Fig.~\ref{data_smples}. 

Chris Anderson~\cite{chris2006longtail}, who first proposed the long-tailed theory, suggested that the future of business and culture lies not in the popular products but in the infinitely long-tailed demand curve, which shows the importance of the research for the tail classes. From the perspective of machine learning research objectives and application implications, we should not only focus on the head classes but also give equal attention to the tail classes in data research.

We must admit that the success of deep learning is inseparable from the large-scale well-annotated datasets, such as ImageNet-1K~\cite{russakovsky2015imagenet}, COCO~\cite{lin2014microsoft}, and Places365 ~\cite{zhou2017places}, \etc These large datasets are artificially balanced, and the classes approximately obey a uniform distribution. In deep learning, we need to be able to learn well for all classes, so artificially balanced data will undoubtedly drive the development of deep learning. Therefore, we should realize that some of the progress made in the field of deep learning is partly driven by this artificial balance by force. In reality, however, this forced balancing of data is inappropriate. On the one hand, forcing class balancing within the dataset by hand is not in line with the natural conditions of data distribution. On the other hand, making the data distribution as balanced as possible by collecting more tail examples is a notoriously difficult task~\cite{lin2014microsoft, van2017devil, everingham2010pascal, krishna2017genome}, and the naturally existing power law can be a huge challenge in constructing a balanced dataset. Thus, the solution to the long-tailed problem is imperative.

\subsection{Previous surveys and our contributions}
To our knowledge, this work is not the first review to summarize the long-tailed phenomenon in visual recognition. Zhang \etal~\cite{zhang2021bag} summarized the technical guide for long-tailed visual recognition earlier this year, aiming to improve the performance of some long-tailed benchmarks through a reasonable combination of existing tricks. Although Zhang \etal's work did not comprehensively introduce and analyze the long-tailed phenomenon in visual recognition, their quantitative analysis of some methods can still be regarded as an early overview of this field. In addition, Zhang \etal~\cite{zhang2021survey} conducted a survey on the topic of deep long-tailed learning in the same period of our work, which grouped the existing deep long-tailed learning studies into three categories (class re-balancing, information augmentation and module improvement), and proposed a new evaluation metric (relative accuracy). In contrast, our work analyzes the long-tailed visual recognition more deeply, divides the existing methods more finely, and quantitatively analyzes the long-tailed phenomenon of mainstream large-scale visual datasets. We recommend that readers also read the above two works~\cite{zhang2021bag, zhang2021survey} in order to have a more comprehensive understanding of long-tailed visual recognition.

This review aims to comprehensively analyze the long-tailed problem in visual recognition, summarize the highlights and limitations of mainstream methods, and provide an outlook on future research directions. At the technical level, we not only sort out some general long-tailed problem solving methods. We also recommend to use the \emph{Gini coefficient}~\cite{gini1912gini} as a measure of the datasets' long-tailedness. At the application level, through the general research on the long-tailed phenomenon, we find that the long-tailed problem is common in the mainstream large-scale visual datasets~\cite{zheng2015market1501, guo2016ms, sami2016youtube8m, zhou2017ade20k, zhao2018mhp, shao2019objects365, weyand2020gldv2}, which reveals that the research on the long-tailed phenomenon in many fields is not enough. In general, the contribution of this survey can be summarized as follows:
\vspace{0.0em}
\begin{itemize}
\item
We conduct a comprehensive review of the advanced long-tailed studies, finely summarize them into ten categories from the perspective of representation learning, and outline the highlights and limitations of each category. \vspace{0.3em}
\item
We have studied four quantitative metrics to evaluate imbalance, deeply compared their characteristics, and proposed to use Gini coefficient to evaluate the long-tailedness of a dataset. \vspace{0.3em}
\item
Beyond the existing research scope, we further study the long-tailed phenomenon of 20 widely-used and large-scale visual datasets proposed in the last decade, and reveal that this problem has not been fully studied in some fields. \vspace{0.3em}
\item
We elaborate on open problems and opportunities in this field to facilitate future research. \end{itemize}

\subsection{Organization}
The rest of this paper is organized as follows. In Sec.~\ref{sec:overview}, we provide the definition of the long-tailed problem and compare the similarities and differences between it and related research fields. In Sec.~\ref{sec:longtail datasets}, we introduce some commonly used long-tailed datasets as well as their evaluation metrics, and use Gini coefficient to quantitatively evaluate the long-tailedness of datasets. In Sec.~\ref{sec:LT_obj-rec}, we give an overview of approaches to solving the long-tailed problem and summarize them based on the existing studies. In Sec.~\ref{sec:performance_comp}, we report the performance of some popular studies on CIFAR-10/100-LT, ImageNet-LT, Places-LT, iNaturalist 2017 \& 2018 as well as LVIS v0.5 \& v1.0. In Sec.~\ref{sec:phenomenon}, we further study the long-tailed phenomenon of mainstream large-scale visual datasets proposed in the last decade. Future directions for the long-tailed problem are given in Sec.~\ref{sec:guidelines}, and Sec.~\ref{sec:concl} concludes the whole paper.

\section{Overview}
\label{sec:overview}
To provide readers with the necessary background knowledge, in Sec.~\ref{sec:problem}, we formulate the task, and analyze the key challenges as well as the driven factors of the long-tailed distribution. And in Sec.~\ref{sec:relevant}, we establish linkages to other relevant fields, and compare their similarities and differences.

\subsection{Problem Definition}
\label{sec:problem}
In nature or real life, there exists a distribution of random variables that is more widespread than the positive-terminus distribution, \ie the long-tailed distribution. It is actually a colloquial expression for the power laws and Pareto characteristics in statistics. The protruding part in the curve is called "head", and the class corresponding to this part is called head class or frequent class. The relatively flat part on the right is called "tail", and the corresponding class is called tail class or rare class. Currently, some CNN-based models~\cite{ren2015faster, liu2016ssd, he2016deep, tian2019fcos} perform well on balanced datasets, but these networks tend to perform poorly on long-tailed datasets.

The long-tailed phenomenon is inherently present in large vocabulary scenarios, making model learning with long-tailed distributed data challenging in a number of ways:

From the perspective of model learning. First, since the data in the tail classes is usually insufficient to represent its true distribution, this poses a significant challenge for classifiers: a good classifier aims to provide a good decision boundary for the model, yet when a class is severely under-represented, it becomes more difficult to determine the location of the decision boundary, which can affect the performance of the model in the dataset. Besides, due to the rich training samples of the head classes, the head classes will be more adequately studied. Based on the case of tail classes severely under-learned, the positive gradient generated by the tail classes will inevitably be overwhelmed by the head classes, which makes it more difficult to learn effective feature extractors and classifiers for the tail classes.

\begin{figure}
	\begin{center}
		\includegraphics[width=0.9\linewidth]{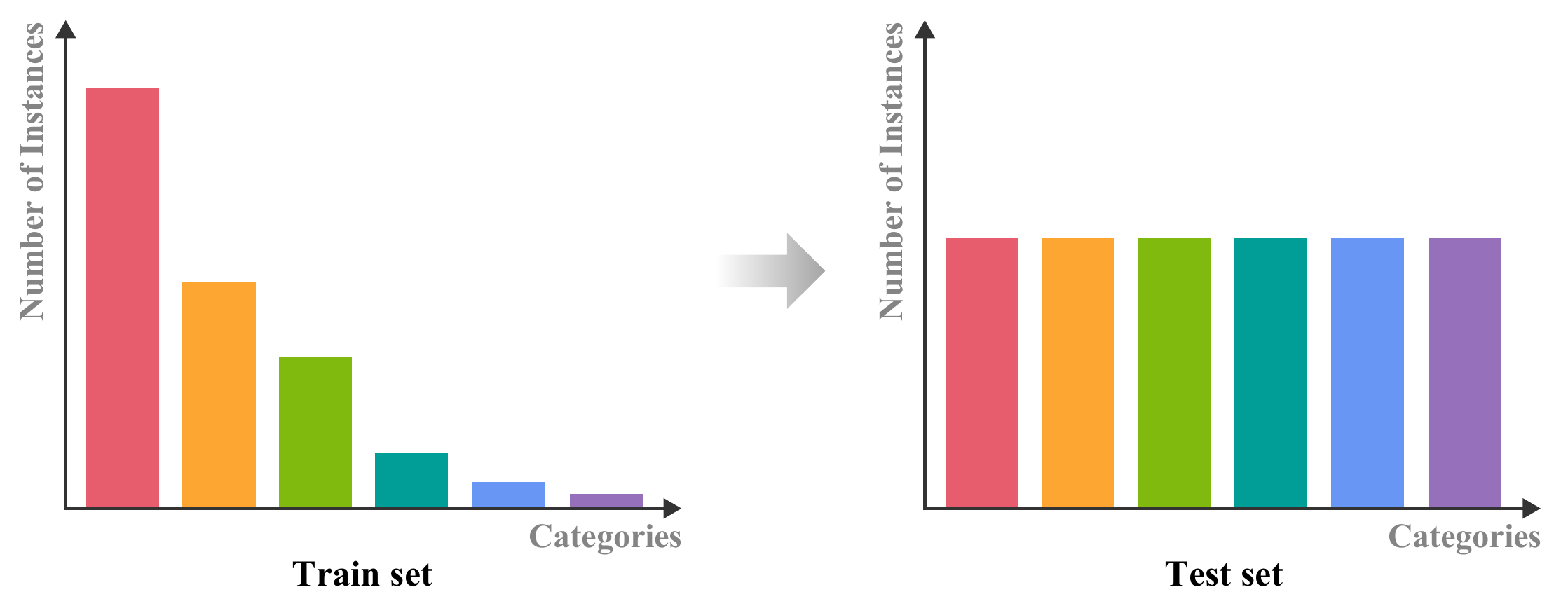}
	\end{center}
	\vspace{-1mm}
	\caption{\textbf{Differences in data distribution between the training and test sets}. For long-tailed dataset, the training set satisfies the long-tailed distribution, while the test set usually conforms to a balanced distribution.}
	\vspace{-1mm}
	\label{train_val}
\end{figure}

From the perspective of transfer learning, we take training data as the source domain and inference-time data as the target domain. For the long-tailed data, the training set satisfies the long-tailed distribution, while the test set usually satisfies the uniform distribution as shown in Fig.~\ref{train_val}. There is no guarantee of having similar data distributions between the source and target tasks due to the large gap between the head and tail item distributions~\cite{wang2017learning, zhang2021model}. Using conventional methods (\eg, Cross-Entropy loss, or simple fine-tuning, \etc) will result in the poor performance of the tail classes. This is because the traditional deep learning methods assume that the training data and the test data satisfy the independently and identically distributed condition. Therefore the quality of knowledge transfer can also be greatly affected. And the problem of \emph{target shift} can arise because the features learned on the training set are different from the features belonging to the corresponding labels in the test set. As the number of tail classes' representative examples are insufficient, which are susceptible to noise and other factors.

In addition to the challenges posed by the long-tailed in the classification task described above, we also investigate the challenges raised by the long-tailed in the object detection task as well as in the instance segmentation task.

For the long-tailed object detection task, there is competition for the category of the boxes between the tail and head classes. In the long-tailed distribution, the data of the tail class is often much smaller than that of the head class, which makes it likely that the sampling of the tail class is directly lost or the tail class is classified as background in the box sampling phase. In addition, for the NMS phase, the long-tailed distribution of data may cause a large number of missed detection~\cite{dave2021eval}, which leads to poor detection results.

For the long-tailed instance segmentation task, the two-stage method is widely used, such as Mask R-CNN, which needs to execute the detection first, so it also faces the limitations of the long-tailed object detection task. In addition, the sparse tail samples make it difficult to go through the learning to distinguish it well from the background, resulting in the inaccurate masks of the tail classes~\cite{zhang2021refinemask}.

\subsection{Relevant Learning Problems}
\label{sec:relevant}
Among the research in machine learning, there are some areas with strong relevance to long-tailed visual recognition, such as imbalance learning and few-shot learning. In this section, we compare long-tailed visual recognition with these two domains (Sec.~\ref{sec:imba} and Sec.~\ref{sec:fsl}), respectively. And we also illustrate the similarities and differences between the three in Sec.~\ref{sec:dif_sim}.

\begin{table*}[htbp]
  \centering
  \caption{\textbf{Similarities and differences between Long-tailed Recognition, Imbalance Learning, and Few-shot Learning in terms of their data and tasks}. Part of the table is extracted from~\cite{liu2019large}. We compared these three areas in terms of the training set (base set), the imbalance of the test set, the sample number of tail classes, the comparison of the class numbers, and evaluation range, respectively.}
  \begin{threeparttable}
   \resizebox{0.99\textwidth}{!}{
    \setlength\tabcolsep{5pt}
    \renewcommand\arraystretch{1.05}
    \begin{tabular}{|c|c|c|c|c|c|}
     \hline\thickhline
     \rowcolor{purple0}
     Task Setting  & \tabincell{c}{Imbalanced Train / Base Set} & \tabincell{c}{Balanced Test Set} & \tabincell{c}{Samples in Tail Classes} & \tabincell{c}{Number of Classes} & Evaluation: Accuracy Over?     \\ 
     \hline \hline
     Imbalanced Learning                   & \checkmark                                                       & x                                               & 20 $\sim$ 50      & less                                             & all classes                                 \\ 
     Few-Shot Learning                      &  x                                                                       & \checkmark                              & 1 $\sim$ 20         & --                                          & novel classes                            \\ 
     \textbf{Long-Tailed Recognition}  & \checkmark                                                       & \checkmark                               & 1 $\sim$ 20        & much                                          & all classes                                 \\ 
     \hline
    \end{tabular}
   }
  \end{threeparttable}
  \vspace{-1mm}
 \label{table:similarities}
\end{table*}

\vspace{6pt}
\subsubsection{Imbalanced Learning}
\label{sec:imba}
Imbalance learning is a widespread problem in deep learning, and it does not only refer to the imbalance of training data. Kemal \etal~\cite{oksuz2020imbalance} proposed that imbalance problems are divided into four types, namely \emph{class imbalance}, \emph{scale imbalance}, \emph{spatial imbalance} and \emph{objective imbalance}. For the long-tailed visual recognition, the current study is mainly based on the image long-tailed distribution level. For the imbalanced distribution of training data, Buda \etal~\cite{buda2018systematic} define and investigate two types of imbalance namely \emph{step imbalance} and \emph{linear imbalance}, which can represent most of the real-world cases.
	
The long-tailed distribution has a strong correlation with the imbalance problem. Specifically, the long-tailed distribution is an extreme case of imbalance. As shown in Tab.~\ref{table:similarities}, generally speaking, it is considered that the imbalance learning is usually reflected in the situation where there are fewer learning classes such as the two-classification problem, while for the long-tailed visual recognition, the number of classes is larger. When the number of classes increases to a certain level, the dataset tends to favor the long-tailed distribution. More importantly, for the long-tailed visual recognition, the tail classes are likely to lack a comprehensive data distribution due to the sparse training examples, and thus the model decision boundaries are more ambiguous, combined with the fact that the number of tail classes occupies most of the dataset, so the training of the model is more challenging, making it difficult to solve the long-tailed problem.
	
\vspace{6pt}
\subsubsection{Few-Shot Learning}
\label{sec:fsl}
In many application scenarios, it is very difficult to collect labeled data, so people want to be able to learn a well-performing model with only a small amount of data. In addition, humans have the ability to learn quickly from a small number of samples, and machine learning was desired to give such a property, which gave birth to few-shot learning (FSL). 
Wang \etal\cite{wang2019few} propose that FSL is a type of machine learning problem, specified by experience $E$, task $T$, and performance measure $P$, where $E$ contains little supervised information for the target $T$. For the $C$-way $K$-shot problem in FSL, it simply means that we need to learn $C$ classes with only $K$ training images (typically no more than 20) in each class, \ie, we are required to learn how to distinguish these $C$ classes in these $C\times K$ images.
	
The tail classes of long-tailed dataset have little supervisory information, which is similar to FSL. But the difference is that the base set of FSL is much more balanced, and head classes of long-tailed datasets are rich in supervised information. Although, generally speaking, the more data is available, the more beneficial it is for deep model learning, but this will inevitably inhibit or over-whelm the tail classes. Therefore, how to balance the relationship between the head classes and the tail classes is also an important point that needs additional consideration in long-tailed learning.

\vspace{6pt}
\subsubsection{Differences and similarities}
\label{sec:dif_sim}
Long-tailed visual recognition has a strong relationship with imbalance learning and few-shot learning. The head and body classes of the long-tailed dataset can be regarded as the traditional imbalance problem~\cite{liu2019large}. Besides, long-tailed data has the characteristics of "long" tails, and each tail class has very little data. Therefore, long-tailed data has a few-shot problem that cannot be ignored. In general, the long-tailed phenomenon is an extreme case of the imbalance problem, and it is also a combination of data imbalance and few-shot learning. Tab.~\ref{table:similarities} summarizes their differences according to ~\cite{liu2019large}.

\section{Long-tailed Datasets and Metrics}
\label{sec:longtail datasets}

Over time, several researchers have proposed some mainstream long-tailed datasets which facilitate the development of long-tailed studies. In this section, we focus our analysis around the long-tailed datasets, starting with introducing some generic long-tailed datasets in Sec.~\ref{sec:longtail_datasets}, and then we constructively analyze four quantitative metrics to measure the long-tailedness of datasets in Sec.~\ref{sec:long-tailedness}. Finally, we list the performance evaluation metrics of some long-tailed benchmarks in Sec.~\ref{sec:eval_metrics}.

\subsection{Long-tailed Benchmark}
\label{sec:longtail_datasets}
\begin{table*}[htbp]
	\centering
	\caption{\textbf{Statistics of representative long-tailed visual recognition datasets}. See Sec.~\ref{sec:longtail_datasets} for more detailed descriptions.}
	\begin{threeparttable}
		\resizebox{0.99\textwidth}{!}{
			\setlength\tabcolsep{5pt}
			\renewcommand\arraystretch{1.05}
			\begin{tabular}{|c|c|c|c|c|c|c|c|c|}
				\hline\thickhline
				\rowcolor{purple0}
				Dataset                                    & Venue                        &  Fields                        & \tabincell{c}{Annotation Types} & Training Samples & Classes & Max Size & Min Size         & Imba. Factor $\beta$ \\ 
				\hline \hline
				CIFAR-10-LT~\cite{cui2019class}                            & CVPR 2019               & Object-centric            & Classification                             & 50,000 - 11,203                          & 10                             & 5,000             & 500 - 25    & 10 - 200               \\ 
				CIFAR-100-LT~\cite{cui2019class}                          & CVPR 2019               & Object-centric            & Classification                             & 50,000 -  9,502                          & 100                            & 500             & 500 - 2       & 1 - 250               \\ 
				ImageNet-LT~\cite{liu2019large}                            & CVPR 2019               & Object-centric            & Classification                             & 115,846               & 1,000                        & 1,280     & 5       & 256           \\
				Places-LT~\cite{liu2019large}                                 & CVPR 2019               & Scene-centric            & Classification                             & 62,500                & 365                           & 4,980     & 5    & 996                \\ 
				iNaturalist 2017~\cite{van2018inaturalist}               & CVPR 2018               & Species-centric         & \tabincell{c}{Classification \\ Bounding-box}  & 579,184 & 5,089                  & 3,919    & 9     & 435               \\ 
				iNaturalist 2018~\cite{inaturalist2018}                        & -                                 & Species-centric         & \tabincell{c}{Classification \\ Bounding-box} & 437,513 & 8,142                   & 1,000     & 2     & 500              \\ 
				MS1M-LT~\cite{liu2019large}                                  & CVPR 2019               & Face-centric             & Classification                             & 887,530                & 74,532                      & 598            & 1        & 598             \\ 
				LVIS v0.5~\cite{gupta2019lvis}                                  & CVPR 2019               & Object-centric           & \tabincell{c}{Bounding-box \\ Instance-mask}    & 56,740   & 1,230  & 26,148  &  1   &  26,148            \\ 
				LVIS v1.0                                  & -                                & Object-centric            & \tabincell{c}{Bounding-box \\ Instance-mask}    & 99,388   & 1,203 & 50,552  & 1    & 50,552             \\ 
				\hline
			\end{tabular}
		}
	\end{threeparttable}
	\vspace{-1mm}
	\label{table:long-tailed_datasets}
\end{table*}

\begin{figure*}[htbp]
	\centering
	\subfloat[ImageNet-LT]{
		\begin{minipage}[t]{0.19\linewidth}
			\centering
			\includegraphics[height=1.0in]{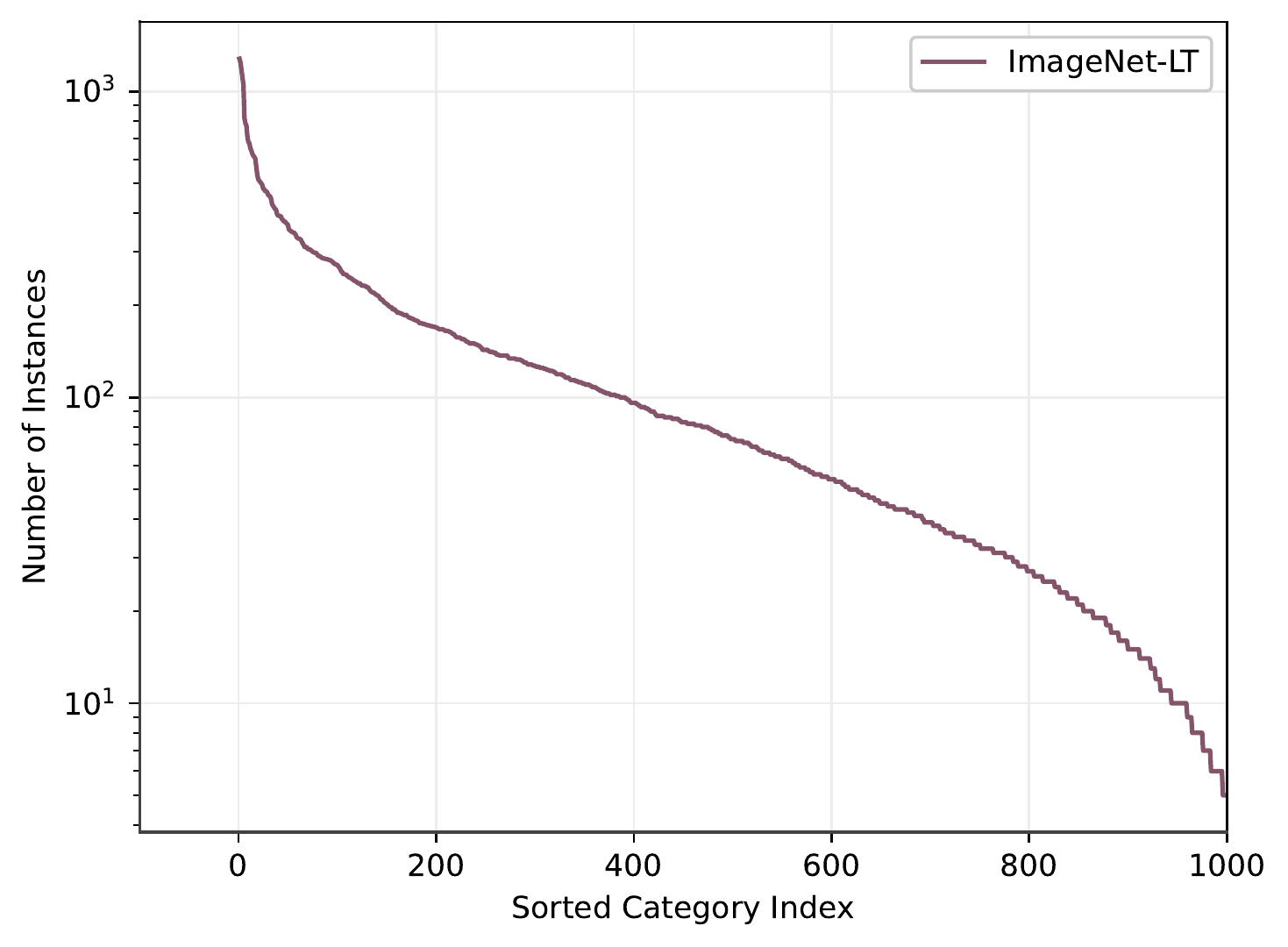}
	\end{minipage}}
	\subfloat[Places-LT]{
		\begin{minipage}[t]{0.19\linewidth}
			\centering
			\includegraphics[height=1.0in]{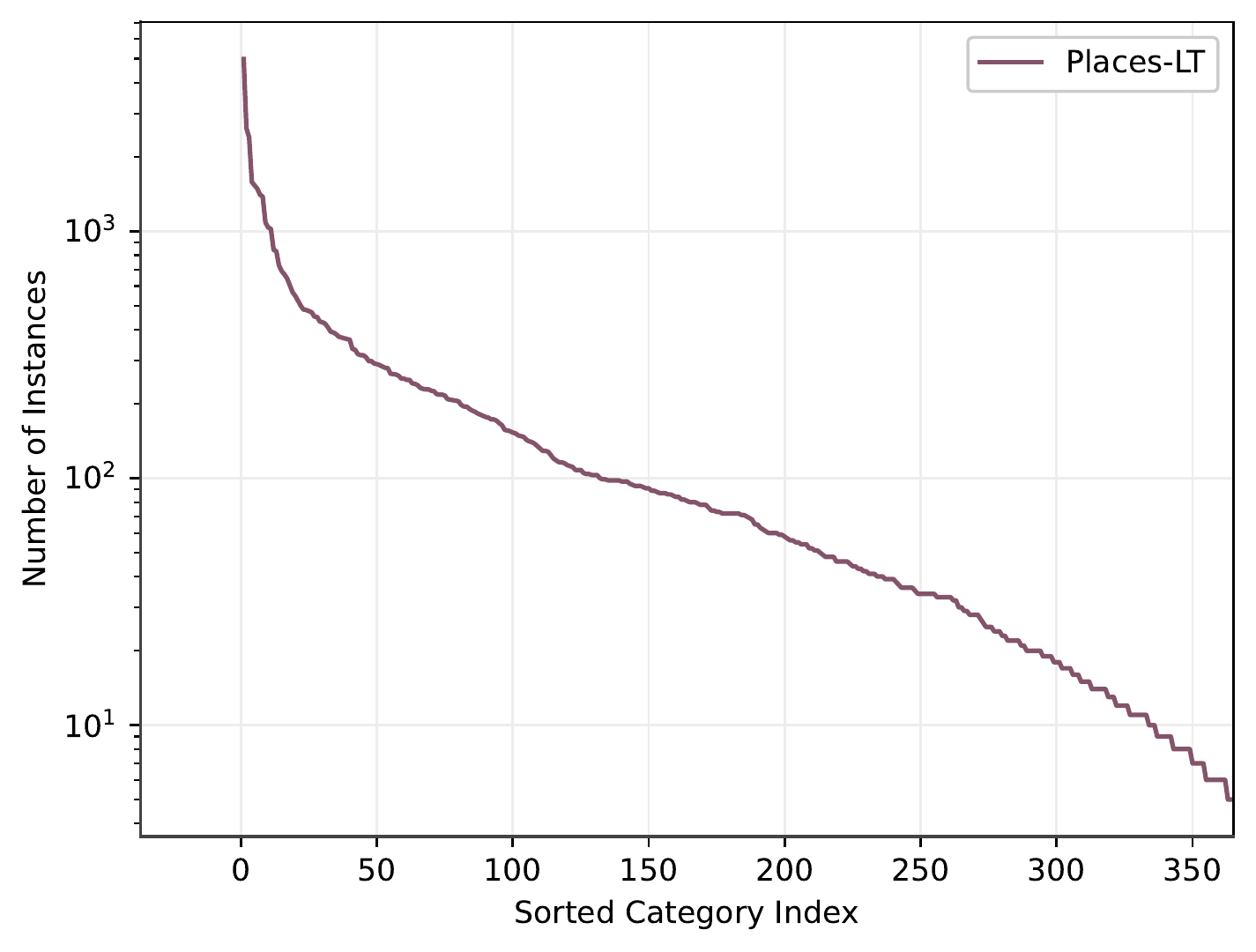}
	\end{minipage}}
	\subfloat[iNaturalist]{
		\begin{minipage}[t]{0.19\linewidth}
			\centering
			\includegraphics[height=1.0in]{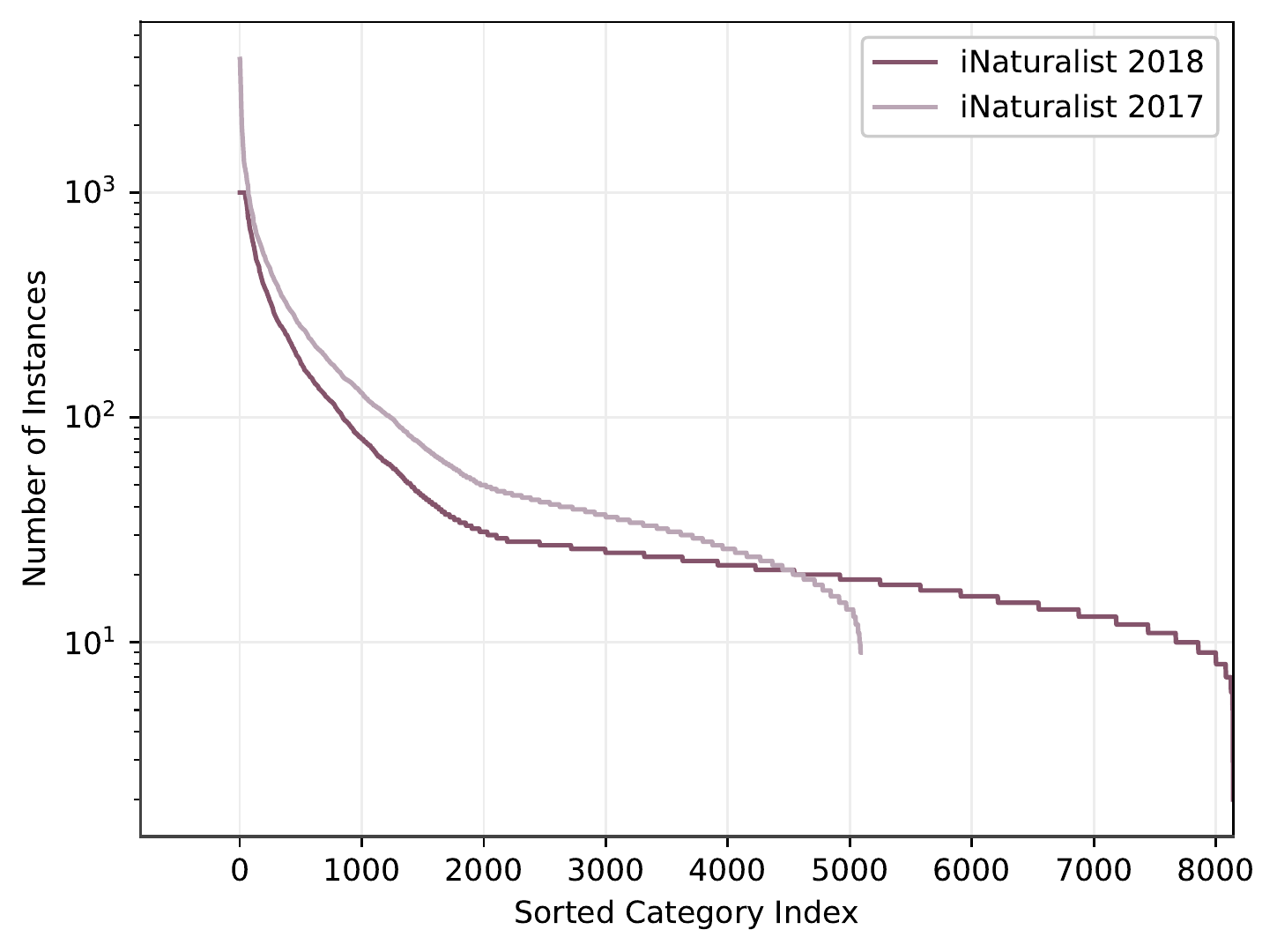}
	\end{minipage}}
	\subfloat[MS1M-LT]{
		\begin{minipage}[t]{0.19\linewidth}
			\centering
			\includegraphics[height=1.0in]{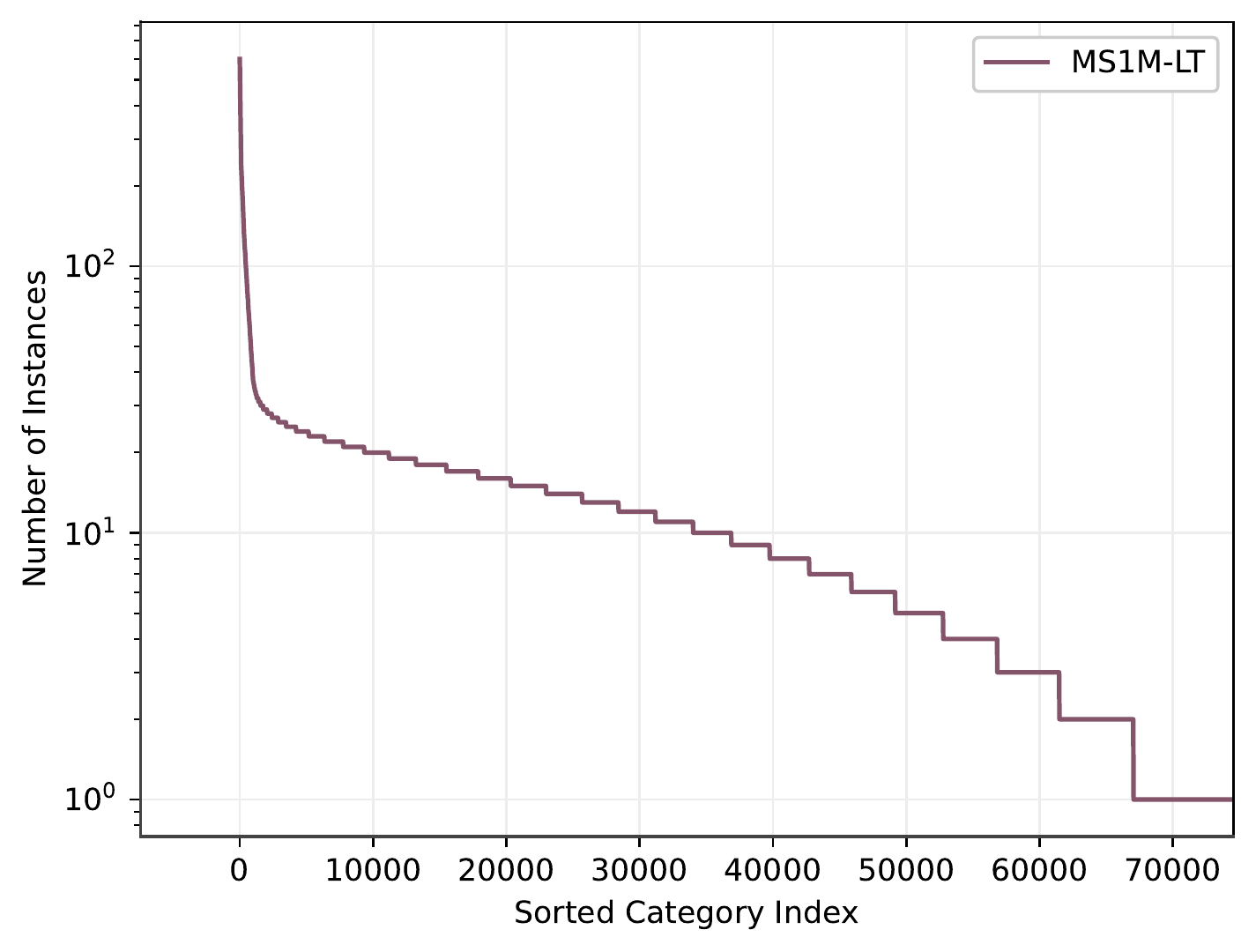}
	\end{minipage}}
	\subfloat[LVIS v0.5 \& v1.0]{
		\begin{minipage}[t]{0.19\linewidth}
			\centering
			\includegraphics[height=1.0in]{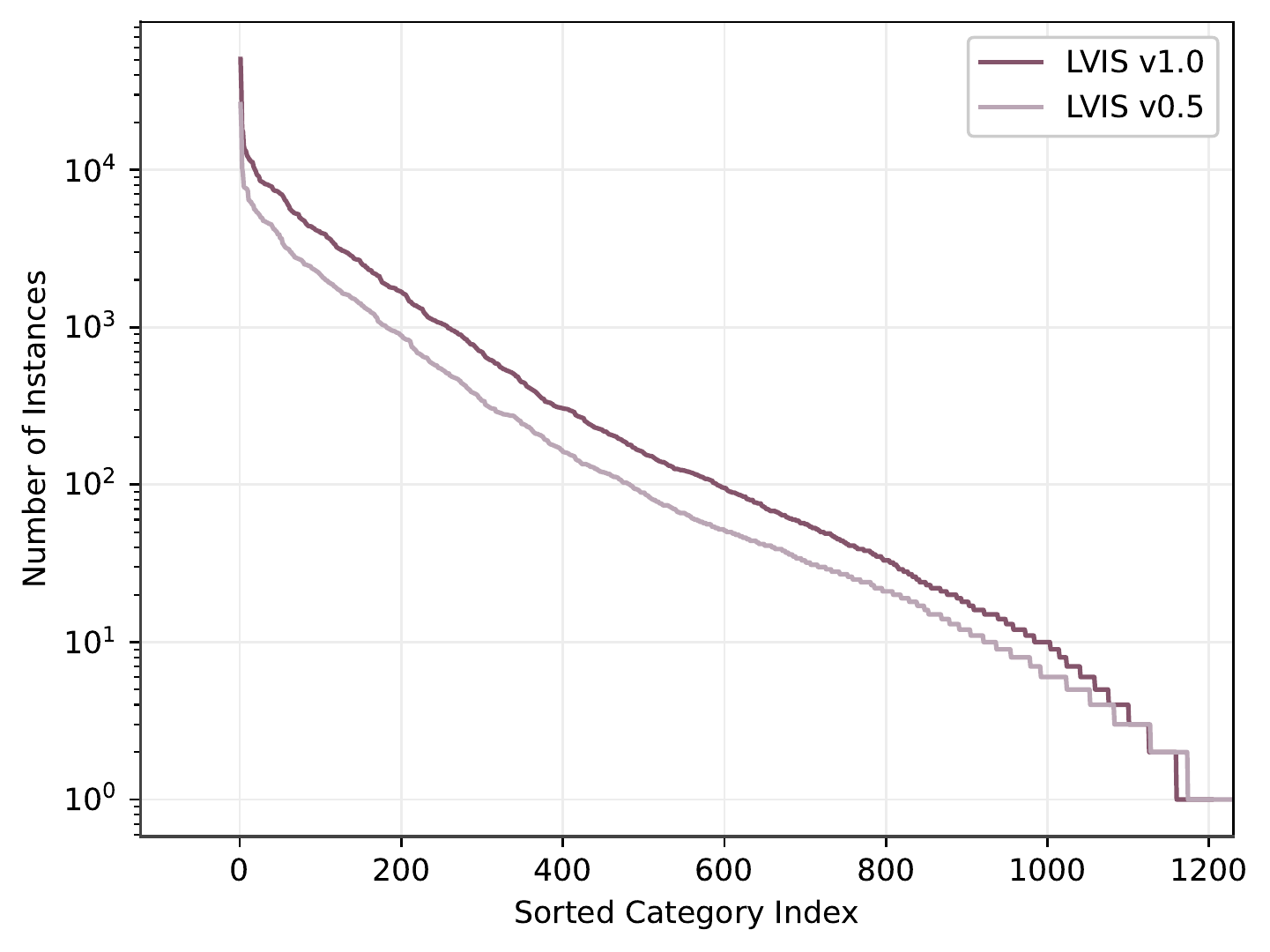}
	\end{minipage}}
	\caption{\textbf{Distributions of common long-tailed datasets}. Figure (a)-(e) show the long-tailed distributions of ImageNet-LT, Places-LT, iNaturalist 2017 \& 2018, MS1M-LT and LVIS v0.5 \& v1.0, respectively.}
	\vspace{-1mm}
	\label{long-tailed_curves}
\end{figure*}

\vspace{6pt}
To better study the long-tailed problem, several long-tailed datasets have been proposed over the past decades. We summarize the commonly used long-tailed datasets in Tab.~\ref{table:long-tailed_datasets}, depicts their category-instances distribution curves in Fig.~\ref{long-tailed_curves}, and give detailed review below.  

\begin{table}[htbp]
	\centering
	\caption{\textbf{Different metrics are used to quantify the long-tailedness}.The upper part of the table lists four common balanced visual datasets, and the lower part is the long-tailed datasets. The Gini coefficient is recommended in this review.}
	\begin{threeparttable}
		\resizebox{0.48\textwidth}{!}{
			\setlength\tabcolsep{5pt}
			\renewcommand\arraystretch{1.05}
			\begin{tabular}{|c|c|c|c|c|}
				\hline\thickhline
				\rowcolor{purple0}
				Dataset                                    &  Imba. Factor $\beta$  &  Std. $\sigma$             & $\dfrac{mean}{median}$ $\gamma$   & \textbf{Gini Coef. $\delta$}  \\ 
				\hline \hline
				CIFAR~\cite{alex2009cifar}             & 1.0                    & 0.0               & 1.0                                  & 0.0            \\
				ImageNet-1K~\cite{deng2009imagenet} & 1.77                    & 70               & 0.98                                  & 0.013            \\
				Places365~\cite{zhou2017places} & 1.62                    & 259               & 0.98                                  & 0.011            \\
				COCO~\cite{lin2014microsoft}     & 1,325                    & 29,360               & 1.76                              & 0.564            \\
				\hline
				ImageNet-LT                            & 256                    & 139               & 1.58                                  & 0.524            \\
				Places-LT                                 & 996                    & 382               & 2.37                                  & 0.671             \\ 
				iNaturalist 2017                        & 435                    &  241              & 2.77                                  & 0.634         \\ 
				iNaturalist 2018                        & 500                    & 117               & 2.44                                  & 0.620    \\ 
				MS1M-LT                                  & 598                   & 18                 & 1.32                                   & 0.473                  \\ 
				LVIS v0.5                                  & 26,148              & 1,516             & 11.7                                 & 0.825             \\ 
				LVIS v1.0                                  & 50,552               & 2,789            & 11.1                                  & 0.820            \\ 
				\hline
			\end{tabular}
		}
	\end{threeparttable}
	\vspace{-1mm}
	\label{table:long_tail_coff}
\end{table}

\vspace{6pt}
\noindent$\bullet$~\textbf{CIFAR-10/100-LT~\cite{cui2019class}.}
CIFAR-10-LT and CIFAR-100-LT are the long-tailed versions of the CIFAR-10 and CIFAR-100~\cite{alex2009cifar}. Both CIFAR-10 and CIFAR-100 contain 60,000 images, 50,000 for training and 10,000 for validation with class number of 10 and 100, respectively.

\vspace{6pt}
\noindent$\bullet$~\textbf{ImageNet-LT~\cite{liu2019large}.} 
ImageNet-LT is a long-tailed version of ImageNet-1K~\cite{deng2009imagenet}, created by Liu \etal~\cite{liu2019large}, including 115.8K images from 1,000 classes, with maximally 1,280 images and minimally 5 images per class.

\vspace{6pt}
\noindent$\bullet$~\textbf{Places-LT\cite{liu2019large}.} Places-LT is a long-tailed version of Places365~\cite{zhou2017places}, which contains 184.5K images from 365 classes, with maximum of 4,980 images and minimum of 5 images per class.

\vspace{6pt}
\noindent$\bullet$~\textbf{iNaturalist 2017 \& 2018~\cite{van2018inaturalist}.}
iNaturalist (iNat) is a real-world fine-grained species classification and detection dataset, covering several domains such as birds, dogs, airplanes, flowers, leaves, food, trees, cars, \etc iNat 2017~\cite{van2018inaturalist} contains 579,184 training images of 5,089 classes, and its 2018 version~\cite{inaturalist2018} has 437,513 training samples in 8,142 classes. 

\vspace{6pt}
\noindent$\bullet$~\textbf{LVIS v0.5 \& v1.0.}
LVIS is proposed by Gupta \etal ~\cite{gupta2019lvis}, which is a large-scale fine-grained vocabulary instance segmentation dataset that is based on the COCO dataset and is annotated with instances for over 1,000 classes of objects. 

\vspace{6pt}
\noindent$\bullet$~\textbf{MS1M-LT ~\cite{liu2019large}.}
MS1M-LT is a face recognition dataset, a long-tailed version of MS1M-ArcFace dataset~\cite{guo2016ms, deng2019arcface}. In MS1M-LT, each identity is sampled with a probability proportional to its number of images, which lead MS1M-LT to a long-tailed distribution with 887,530 images and 74,532 identities.




\subsection{Long-tailedness Metrics}
\label{sec:long-tailedness}

Accurate and objective measurement of the long-tailedness of data is an important prerequisite to solve the long-tailed visual recognition problem. Therefore, in this section, we compare four commonly used quantitative metrics in statistics, and critically analyze their advantages and disadvantages in measuring the long-tailedness.

\vspace{6pt}
\subsubsection{Four Quantitative Metrics in Statistics}
\label{sec:four_metrics}

\noindent$\bullet$~\textbf{Imbalance Factor.}
In ~\cite{cui2019class}, Cui \etal defined the imbalance factor (denoted as $\beta$) of a dataset as the number of training samples in the largest class divided by the smallest:
\begin{equation}
	\beta = max\left \{n_{1},n_{2},...,n_{k}\right \} /  min\left \{n_{1},n_{2},...,n_{k} \right \}
\end{equation}
where $n_{1},n_{2},...,n_{k}$ represents the number of samples in different classes. Although the imbalance factor is widely-used as a measurement of the long-tailedness~\cite{cui2019class, liu2019large, van2018inaturalist}, it is easily affected by extreme classes and can not reflect the overall characteristics of the dataset.

\vspace{6pt}
\noindent$\bullet$~\textbf{Standard Deviation.}
Standard deviation (denoted as $\sigma$) is frequently used in probability statistics as a measurement of statistical dispersion~\cite{david1954dist, robert2005stati}, and can also reflect the uncertainty of sampling in some cases~\cite{louise1999un, howard2001dist}, it can be expressed as:
\begin{equation}
	\sigma=\sqrt{\frac{1}{k}\sum\limits_{i=1}^{k}(n_{i}-\mu)^{2}}
\end{equation}
where $k$ represents the number of classes; $n_{i}$ represents the instance number of class $i$, and $\mu$ represents the average number of instances. Although standard deviation quantifies the dispersion degree between the number of classes within a dataset, it is also affected by the absolute number of samples, so it is difficult to objectively express the long-tailedness of data. In the third column of the Tab.~\ref{table:long_tail_coff}, we counted the standard deviations of some balanced datasets as well as long-tailed datasets, and it can be found that the balanced dataset, COCO ($\sigma$=29,360), has the largest standard deviation, and the long-tailed dataset, MS1M-LT ($\sigma$=18) has the smallest one, which shows that the standard deviation can not well identify the long-tailedness.

\vspace{6pt}
\noindent$\bullet$~\textbf{Mean / Median.}
Median is a proper term in statistics and is widely used in economics~\cite{malay1996estimation}, sociology~\cite{andrew2000inter} and medicine~\cite{greene2020covid}. Compared with the mean, the median is not affected by the maximum or minimum of data, and can better represent the distribution of data to a certain extent. Therefore, the ratio of mean to median (denoted as $\gamma$) can also reflect the skew distribution of data, which can be expressed by: 
\begin{equation}
	\gamma = \frac{mean(n_{1},n_{2},...n_{k})}{median(n_{1},n_{2},...n_{k})}
\end{equation}
When $\gamma$ is closed to 1, it indicates that the dataset is uniformly distributed, and when $\gamma$ is significantly greater than 1, it indicates that the dataset is of imbalance, including a large gap between the instance number in head and tail classes. As shown in Tab.~\ref{table:long_tail_coff}, although $\gamma$ accurately distinguishes between balance datasets and long-tailed datasets. However, like imbalance factor, it is easily affected by individual cases and cannot reflect the overall distribution. And the value range of $\gamma$ is an open interval (no upper limit), which is not a good characteristic for a measure.

\begin{figure}
	\begin{center}
		\includegraphics[width=0.50\linewidth]{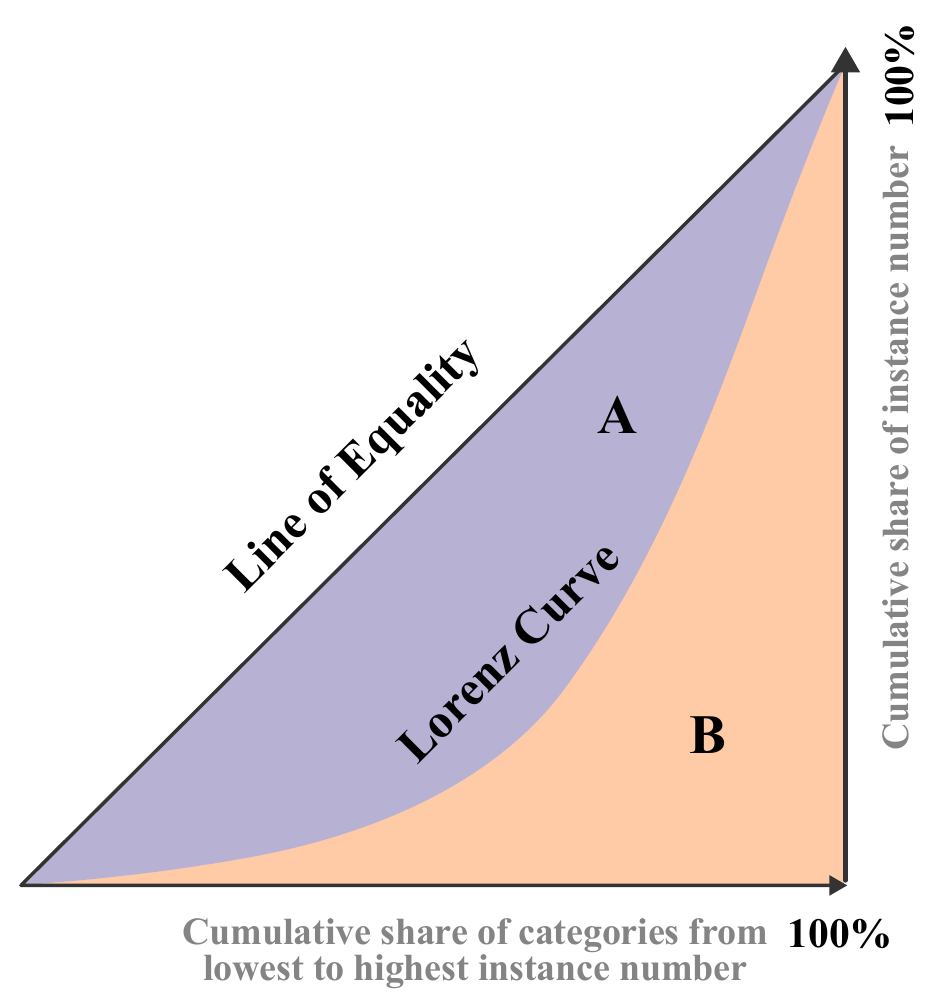}
	\end{center}
	\vspace{-1mm}
	\caption{\textbf{Calculation of the Gini coefficient}. The horizontal axis is the cumulative distribution of the class proportion and the vertical axis is the cumulative distribution of the instance number proportion.}
	\vspace{-1mm}
	\label{Gini}
\end{figure}

\vspace{6pt}
\noindent$\bullet$~\textbf{Gini Coefficient.}
Gini coefficient (denoted as $\delta$) was originally proposed by the Italian economist Gini in 1912~\cite{gini1912gini} as an indicator to judge the degree of distribution equality based on Lorentz curve. It is always used to represent income inequality or wealth inequality~\cite{kakwani1977lorenz, shlomo2013more}. Since long-tailedness is similar to inequality between each category, Gini coefficient can serve as a long-tailed metric. As shown in Tab.~\ref{table:long_tail_coff}, Gini coefficient can effectively distinguish balanced datasets and long-tailed datasets.

The calculating process of Gini coefficient consists of three steps. First, we suppose that the set of number samples of $k$ classes dataset ${n_i}, (i = 1, 2, ..., k)$ is in ascending order, we calculate the normalized cumulative distribution $\left\{C_{i}\right\}$ by:
\begin{equation}
        C_{i}=\frac{1}{k} \sum_{j=1}^{i} n_{j}
\end{equation}

Intuitively, $C_i$ indicates the probability of the $i$s smallest categories. By defining $C_0=0$, we can normalize the x-axis to the share of total categories and interpolate linearly to obtain continuous Lorentz curve $L(x), x\in[0,1]$ (as shown in Fig.~\ref{Gini}). The Lorentz curve $L(x)$ follows:
\begin{equation}
       L(x)=\left\{\begin{aligned}
                &C_{i}, &x=\frac{i}{k} \\
                &C_{i}+(C_{i+1}-C_{i})(kx-i), &\frac{i}{k}<x<\frac{i+1}{k}
       \end{aligned}\right.
\end{equation}
where $i=1,2,\dots,k$. $B$ represents the area to the lower right of the actual instance number distribution curve. Since Lorentz curve is linearly interpolated by $\{C_i\}$, we can calculate the area by trapezoids:
\begin{equation}
        B = \int_{0}^{1}L(x)dx=\sum_{i=1}^k \frac{C_i + C_{i-1}}{2} \cdot \frac{1}{k}
\end{equation}

For balanced dataset, Lorentz curve is an identity line and $A$ presents the area between the identity line and the Lorentz curve of an actual dataset. Thanks to the normalization of Lorentz curve, $A$ can be simply calculated by: $A = 0.5 - B$ . Finally, the Gini coefficient can be expressed as:
\begin{equation}
       \delta=\frac{A}{A+B}
\end{equation}

The Gini coefficient conforms to the closed interval distribution of (0,1), so it can better quantify the degree of imbalance and make the datasets more comparable with each other. Usually, the smaller the Gini coefficient $\delta$ of a dataset is, the more imbalanced the dataset is, and vice versa.

\vspace{6pt}
\subsubsection{Long-tailedness Analysis}
\label{sec:four_metrics}

Since the Gini coefficient $\delta$ is not affected by extreme samples, is not affected by the absolute number of data, and has a bounded distribution, we recommend using the Gini coefficient to measure the long-tailedness of data. Based on Gini coefficient, we quantify the long-tailedness of some commonly used balanced datasets~\cite{deng2009imagenet, lin2014microsoft} and long-tailed datasets~\cite{cui2019class, liu2019large, van2018inaturalist, gupta2019lvis}, as shown in Tab.~\ref{table:long_tail_coff}.

As the most widely used visual dataset, CIFAR~\cite{alex2009cifar} has a perfect manual balance, and the number of instance in each class is equal, so its Gini coefficient is 0. Another widely used visual dataset, ImageNet-1K~\cite{deng2009imagenet} contains 1,000 classes, and the instances in each category are also manually balanced. The Gini coefficient of ImageNet-1K is 0.013. In contrast, the Gini coefficients of long-tailed datasets are generally above 0.5, and some can even reach 0.8. It can be seen that using Gini coefficient to measure the long-tailedness of data is reasonable and effective, and there are great differences in the long-tailedness of existing long-tailed datasets.

On the other hand, COCO~\cite{lin2014microsoft} is the most common dataset to evaluate the performance of object detection and instance segmentation methods, and it is considered to be balanced. The Gini coefficient of COCO is 0.564, which is much larger than the balanced datasets CIFAR, ImageNet-1K and Places365~\cite{zhou2017places}, and even larger than the long-tailed datasets ImageNet-LT and MS1M-LT. However, the annotation of these datasets is image-level, and it is much simpler to manually control the data distribution than the instance-level annotation datasets. As long-tailed object detection / instance segmentation datasets, LVIS v0.5 and LVIS v1.0, their Gini coefficients are 0.825 and 0.820 respectively, which are still much larger than COCO. Therefore, for different visual tasks, different standards should be used to measure the long-tailedness of data distribution.

With the above analysis, we can quantitatively analyze the long-tailedness of most visual datasets. This provides guidance for long-tailed visual recognition, that is, different solutions are adopted through the long-tailedness of data. In Sec.~\ref{sec:phenomenon}, we will also use this standard to study 20 mainstream large-scale long-tailed visual recognition datasets, so as to deeply reveal the research status and future direction of this field.

\begin{figure*}[htbp]
\begin{center}
	\includegraphics[width=0.95\linewidth]{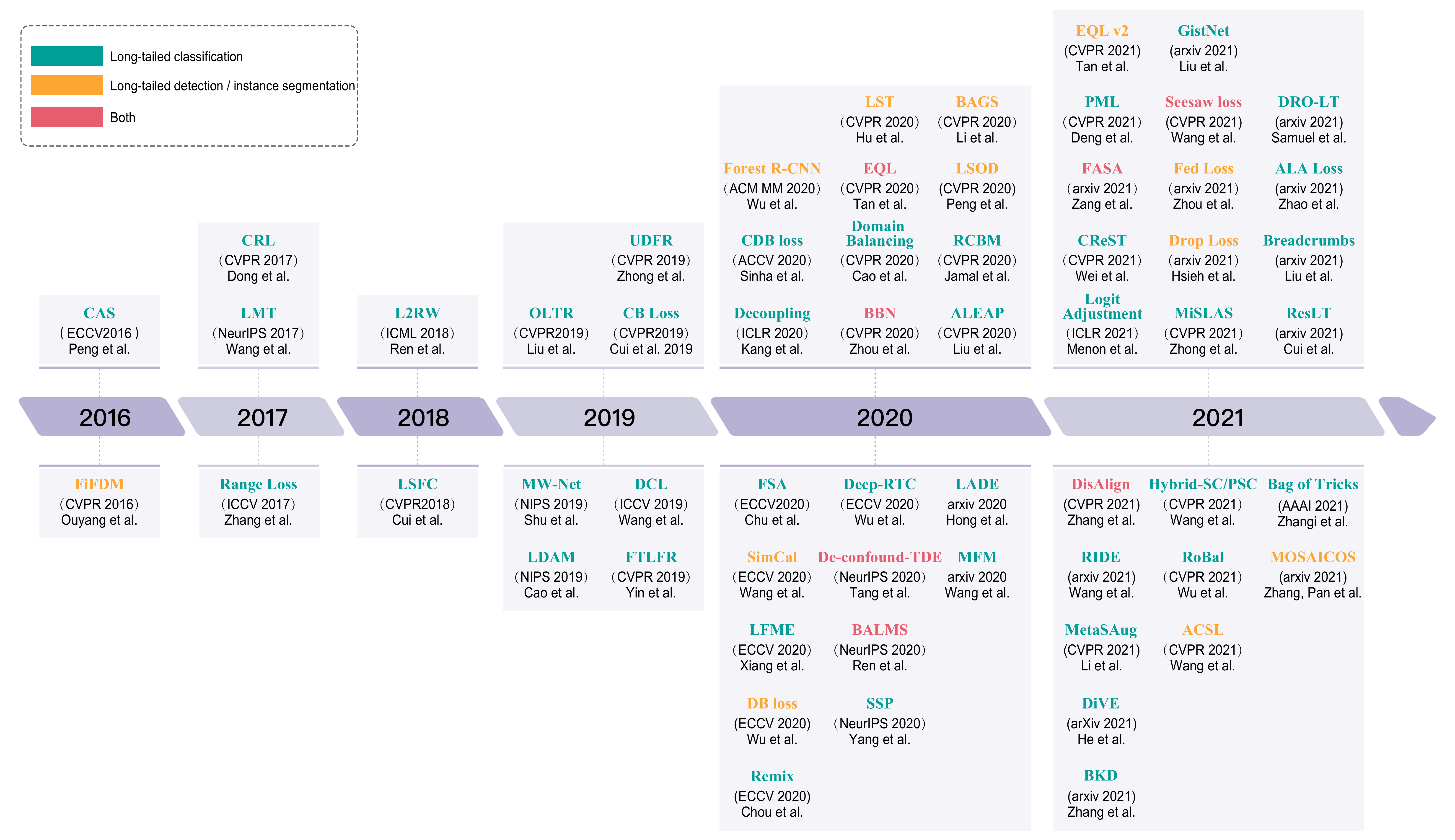}
\end{center}
\vspace{-1mm}
\caption{\textbf{A chronological overview of recent representative work in long-tailed recognition}. Visual recognition study on long-tailed distribution started with ~\cite{ouyang2016factors} in 2016 and has become more and more abundant since then. Work marked in green represents long-tailed classification, marked in orange represents long-tailed object detection / instance segmentation, marked in red represents both. We abbreviate some studies for ease of presentation.}
\vspace{-1mm}
\label{timeline}
\end{figure*}

\vspace{6pt}
\subsection{Performance Evaluation Metrics}
\label{sec:eval_metrics}

Presently, there are several general metrics that are widely-used to measure how long-tailed methods perform on classification and detection tasks. In this section, we provide a review of the major performance evaluation metrics for long-tailed recognition.

In terms of evaluation metric for classification, the top-1 accuracy is frequently adopted in the research community. For ImageNet-LT, Places-LT, and iNaturalist 2018 datasets, the accuracy can be split into four types based on the set of classes followed Liu \etal~\cite{liu2019large}: \emph{Many-shot} (classes each with over training 100 images), \emph{Medium-shot} (classes each with 20$\sim$100 training images), \emph{Few-shot} (classes under 20 training images) and the \emph{Overall} accuracy. In CIFAR-10/100-LT, datasets with different long-tailedness is be sampled according to the imbalance factor $\beta \in \{200,100,50,20,10,1\}$, and their evaluation metric is to measure the top-1 accuracy of the datasets under different imbalance factors respectively. To make a full comparison between different methods, we report benchmarks on CIFAR-10/100-LT (Tab.~\ref{table:cifar_lt_bench}),  ImageNet-LT and Places-LT (Tab.~\ref{table:im_places_lt_bench}), as well as iNaturalist (Tab.~\ref{table:inat_bench}),  in Sec.~\ref{sec:performance_comp}, respectively.

For object detection and instance segmentation tasks, there are several evaluation metrics that are used in LVIS v0.5 and v1.0, such as AP$ _{r} $(mask AP for rare classes), AP$ _{c} $(mask AP for common classes) and AP$ _{f} $(mask AP for frequent classes). To make a full comparison, we keep the common evaluation metrics in detection and instance segmentation tasks, like mask AP for \{AP, AP$_{50}$, AP$_{75}$\}, \{AP$_{r}$, AP$_{c}$, AP$_{f}$\} and bounding-box AP, on LVIS v0.5 and v1.0 benchmark in Sec.~\ref{sec:performance_comp}  (see Tab.~\ref{table:lvis_v05_bench} and Tab.~\ref{table:lvis_v10_bench}).

\section{Long-tailed Visual Recognition}
\label{sec:LT_obj-rec}

In the past few years, a growing number of research has investigated the long-tailed distribution of data as shown in Fig.~\ref{timeline}. In this section, we review deep learning based methods for long-tailed visual recognition from 2016 to present and introduce the related earlier work in context. Although many studies have mixed a variety of methods to solve the long-tailed problem, in order to highlight the contribution of each study, we mainly reviewed their core methods, finely summarized them into ten categories from the perspective of representation learning, and outlined the highlights and limitations of each category (Tab.~\ref{table:highlights}). 

\begin{figure}[htbp]
\begin{center}
	\includegraphics[width=0.65\linewidth]{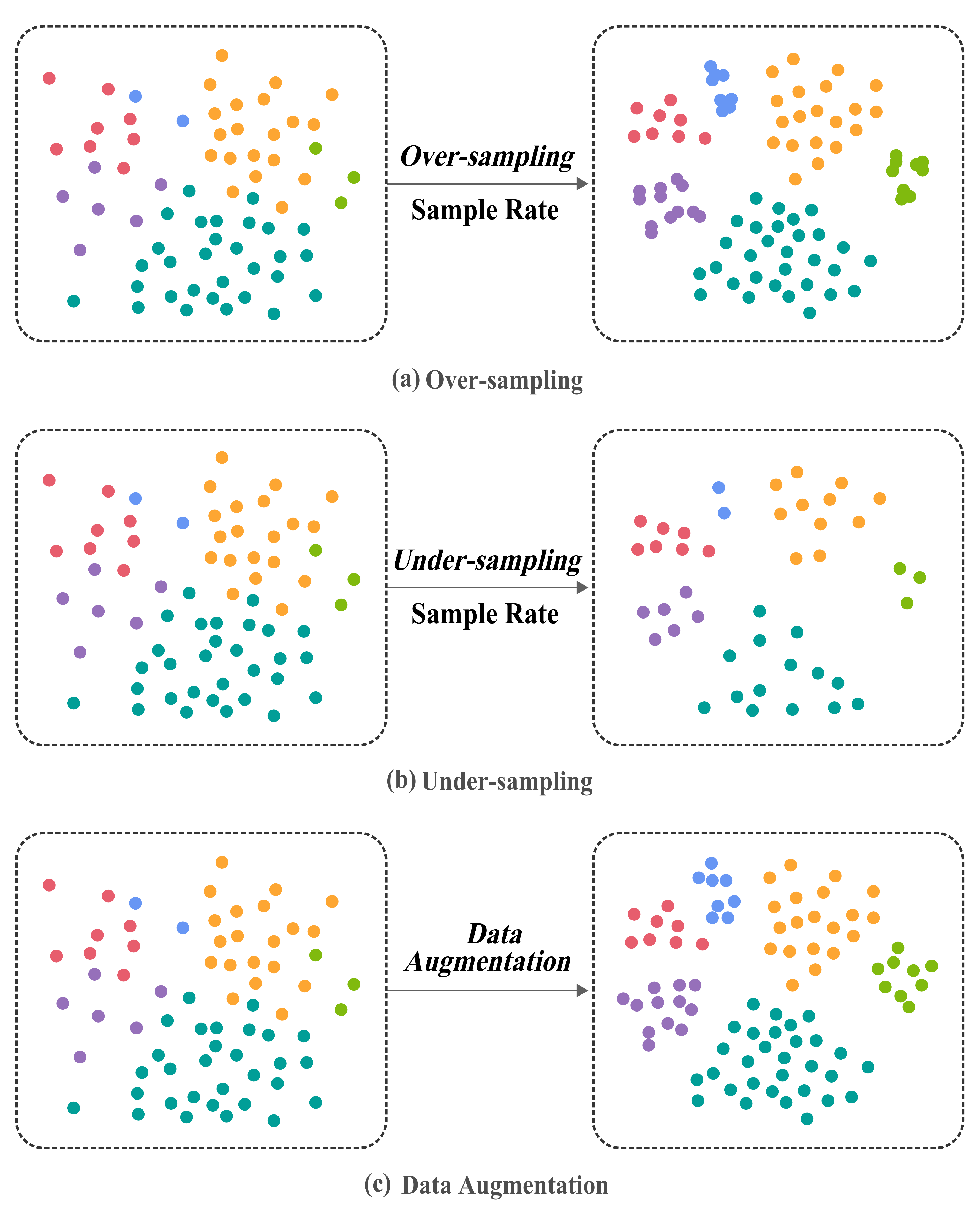}
\end{center}
\vspace{-1mm}
\caption{\textbf{Data processing methods for long-tailed problem}. Data points with the same color represent the same class. (a) and (b) represent data over-sampling and under-sampling, respectively. \ie, assigning different sampling rates for head/tail classes. (c) represents data synthesis, \ie, synthesizing new data for tail classes to increase their weighting.}
\vspace{-1mm}
\label{data_processing}
\end{figure}

\begin{figure}[htbp]
\begin{center}
	\includegraphics[width=0.9\linewidth]{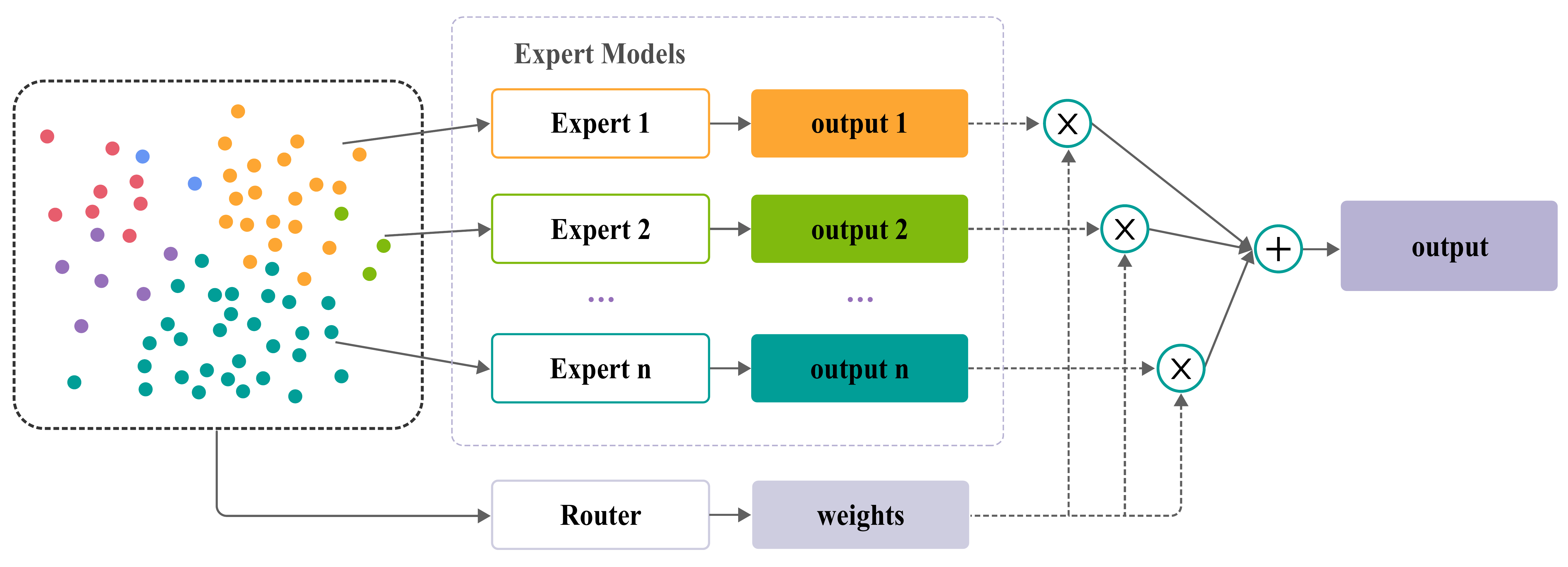}
\end{center}
\vspace{-1mm}
\caption{\textbf{Mixture-of-Experts for long-tailed problem}. Training multiple expert networks with a routing network. Each expert network in MoE has a data region in which it excels and on which it performs better than the other experts.}
\vspace{-1mm}
\label{experts_system}
\end{figure}

\begin{figure}[htbp]
\begin{center}
	\includegraphics[width=0.65\linewidth]{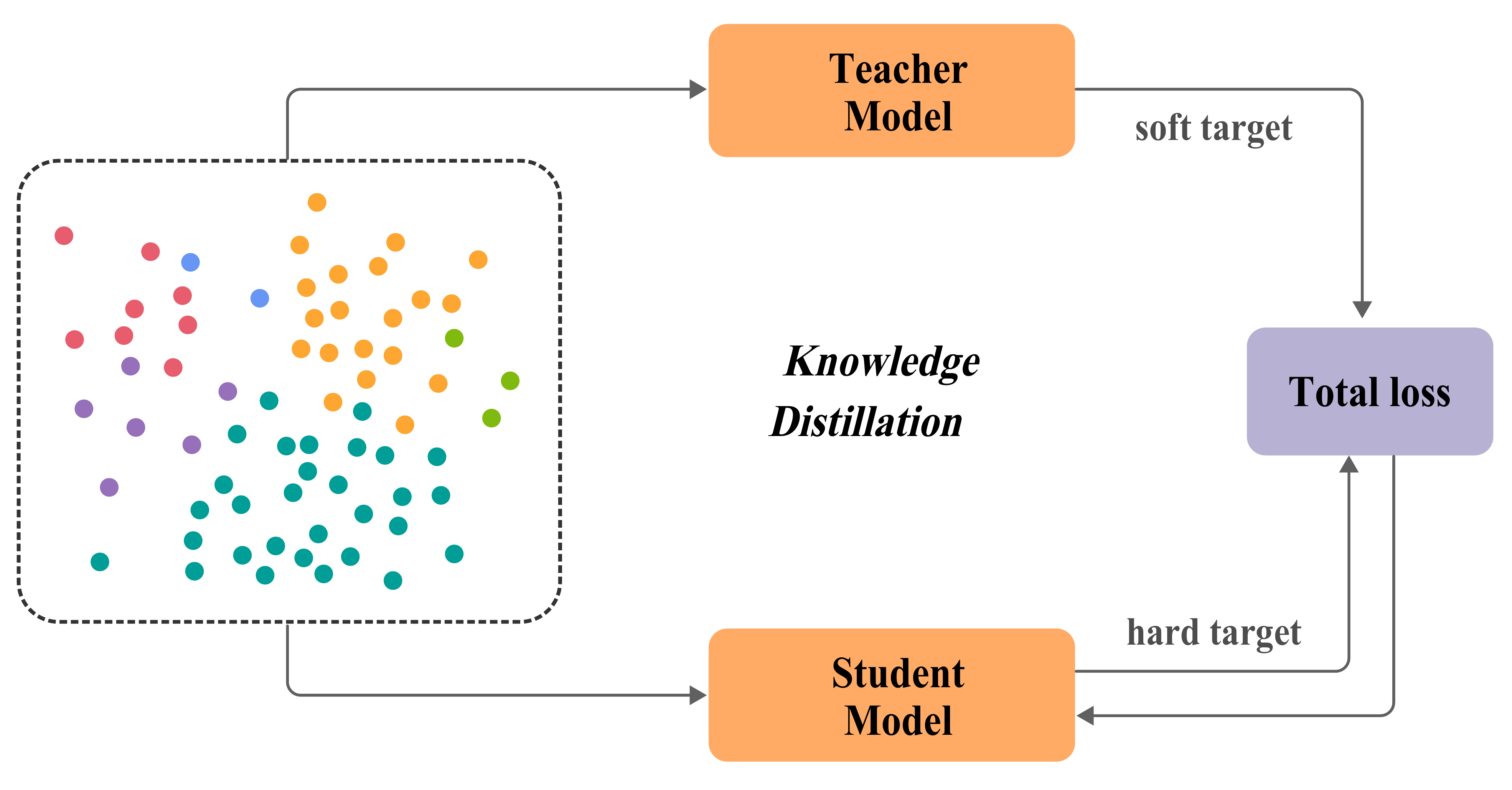}
\end{center}
\vspace{-1mm}
\caption{\textbf{Knowledge distilling for long-tailed problem}. With the guidance of the teacher model, the student model can learn a more balanced sample.}
\vspace{-1mm}
\label{knowledge_distilling}
\end{figure}

\begin{figure}[htbp]
\begin{center}
	\includegraphics[width=0.78\linewidth]{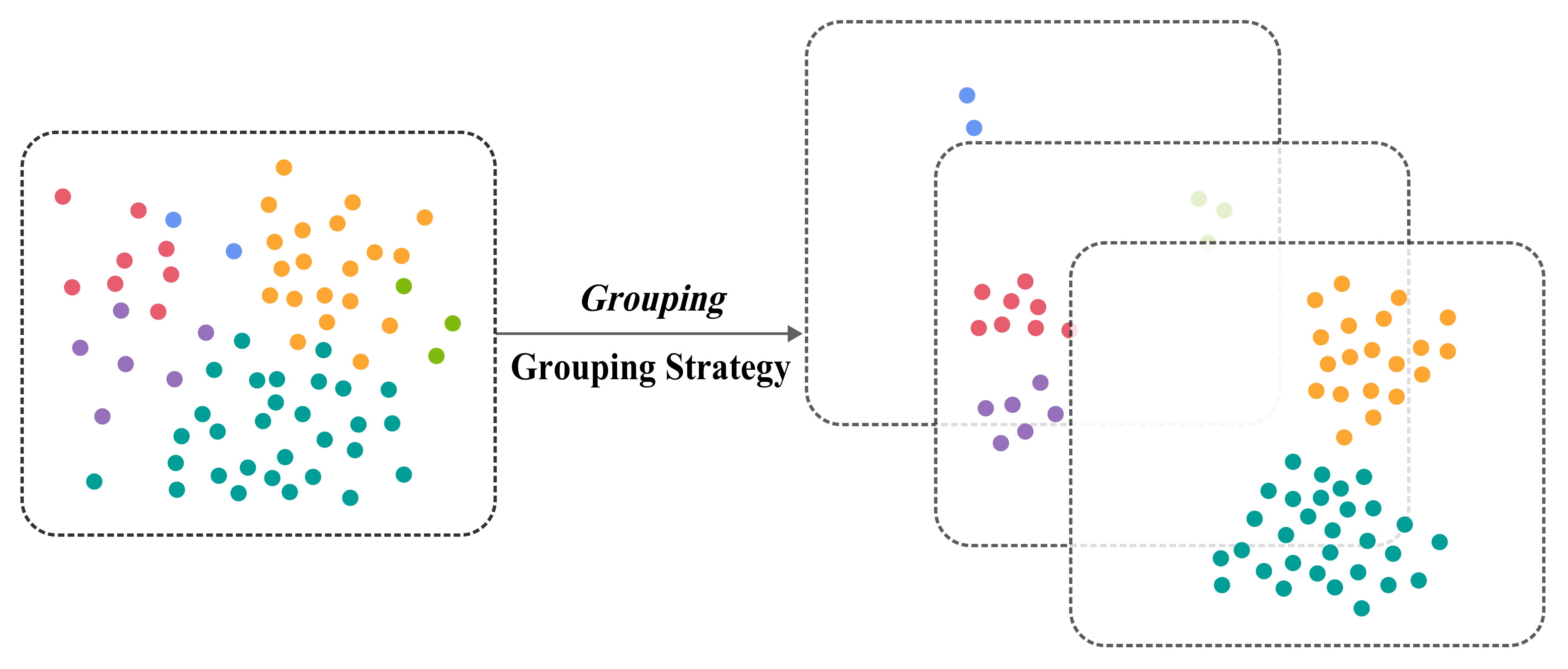}
\end{center}
\vspace{-1mm}
\caption{\textbf{Grouping method for long-tailed problem}. Group the data according to certain strategy and make the model learn separately for each group of data. The figure shows the group assignment according to the number of instances.}
\vspace{-1mm}
\label{grouping}
\end{figure}

\begin{figure}[htbp]
\begin{center}
	\includegraphics[width=0.65\linewidth]{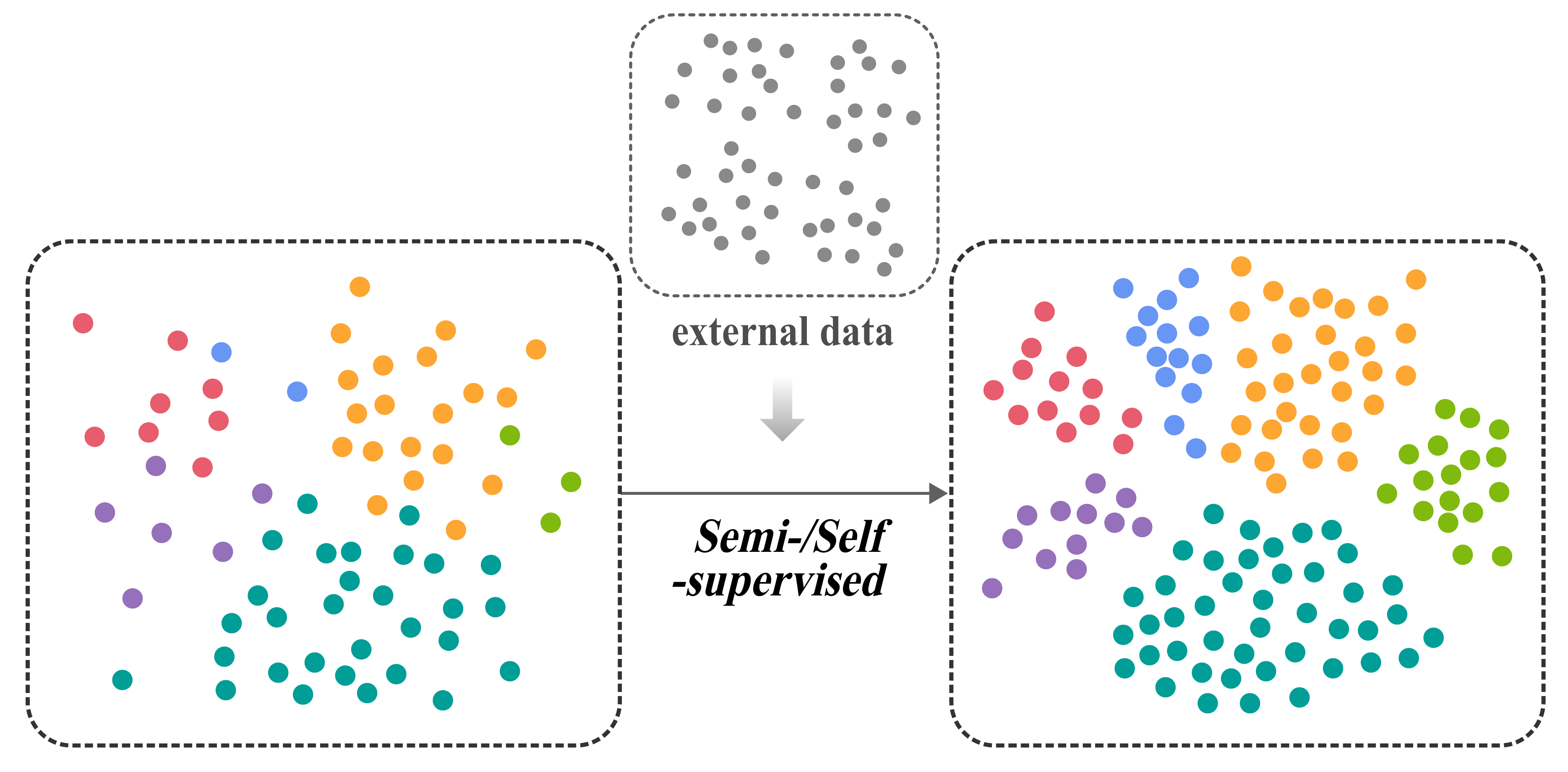}
\end{center}
\vspace{-1mm}
\caption{\textbf{Semi-supervised for long-tailed problem}. Adding external data through semi-supervised strategy, thus increasing the weight of the tail classes in the overall training data.}
\vspace{-1mm}
\label{semi_supervised}
\end{figure}

In order to make it easier for readers to understand the characteristics and differences of various methods, we use color scatter diagram to express the principle of each category (Fig.~\ref{data_processing}  - Fig.~\ref{decoupling}, Fig.~\ref{metric_learning}  - Fig.~\ref{semi_supervised}). Among them, dots of different colors represent different classes, gray dots represent unlabeled data, and the number of dots represents the instance number in that category. The circle outside the dot indicates that this data is sampled multiple times.

\subsection{Data Processing}

For dataset's long-tailed distribution, an intuitive idea is to make the model learn relatively balanced classes from the perspective of data. There are three ways to handle the data, namely over-sampling, under-sampling and data augmentation.

\subsubsection{Over-sampling}
Over-sampling is one of the most common methods in deep learning~\cite{haixiang2017learning, janowczyk2016deep, levi2015age}. As shown in Fig.~\ref{data_processing} (a), the over-sampling method emphasizes the tail classes and increases the instance number of the tail classes~\cite{byrd2019effect, buda2018systematic, shen2016relay} to reduce the imbalance between the head classes and the tail classes.
	
Shen \etal~\cite{shen2016relay} propose a sampling strategy \emph{Class-Aware Sampling (CAS)} to ensure that each class has the same probability of occurrence in each batch as much as possible. We denote the \emph{CAS} probability of the $i-th$ class as $P_{a}(i)$, \ie, for a total of $C$ classes, following the definition of~\cite{peng2020large}, the sampling probability for each class is:
\begin{equation}
	P_{a}(i)=\dfrac{1}{C}
\end{equation}
Dhruv \etal~\cite{mahajan2018exploring} compute a replication factor for each image based on the distribution of labels and repeated the images several times based on the replication factor. Inspired by this, the work of Gupta \etal~\cite{gupta2019lvis} propose \emph{Repeat Factor Sampling (RFS)} to perform rebalancing operations on training data by increasing the sampling frequency of images containing tail instances. \emph{Soft-balance Sampling with Hybrid Training}~\cite{peng2020large} combined the conventional sampling scheme and \emph{CAS}, which first trains the detector using the conventional strategy, and then introducing hyper-parameters to control the degree of ordinary sampling with $P_{o}(i)=\dfrac{n_{i}}{N}$, where $n_{i}$ represents the instance number of class $i$ and $N$ is the total number of instances.

In addition to the above-mentioned works that perform re-sampling at the image level, some works perform instance-level re-sampling to prevent unnecessary duplication of certain non-tail class instances in the oversampled images. Hu \etal~\cite{hu2020learning} presente \emph{Instance-level Data Balanced Replay} strategy. At stage $t$, for each class, a certain number of images containing that class are randomly sampled, and among these images, only annotations belonging to that class are considered valid during the training process. \emph{NMS Resampling}~\cite{wu2020forest} adaptively adjusts the \emph{non-maximum suppression (NMS)} threshold for different classes according to their label frequencies, so as to balance the data distribution by retaining more proposal candidates from the tail classes while suppressing those from the head classes.
	
\subsubsection{Under-sampling}
In contrast to the over-sampling method, the under-sampling method reduces the imbalance between the head classes and the tail classes by reducing the sample times of the head classes~\cite{buda2018systematic, japkowicz2002class, he2009learning, haixiang2017learning}, as shown in Fig.~\ref{data_processing} (b). 

Random under-sampling is performed by randomly deleting the head classes data until it has the same number of instances as the other classes ~\cite{buda2018systematic}. \emph{EasyEnsemble}~\cite{liu2008exploratory} divides the frequent classes into several subsets, each with the same number of instances as the rare classes, and combines them separately with the rare classes to train multiple classifiers and eventually combine the outputs of multiple classifiers in an ensemble fashion, thus dealing with the information loss problem in traditional random under-sampling. \emph{BalanceCascade} trains the classifiers sequentially, where in each step, the majority class examples that are correctly classified by the current trained learners are removed from further consideration. \emph{NearMiss}~\cite{mani2003knn} is another method to alleviate the information loss problem in random under-sampling, which is essentially a prototype selection method that uses \emph{KNN} to select the most representative samples from frequent class samples for training. 

There are also some data cleaning methods, which mainly clean overlapping data to achieve the purpose of under-sampling. In \emph{Edited Nearest Neighbours (ENN)}~\cite{wilson1972asymptotic}, for a sample belonging to a frequent class, if more than half of its $K$ nearest neighbors do not belong to the frequent class, this sample is eliminated. Modified \emph{Tomek Link} method~\cite{devi2017redundancy} performs data cleaning by finding pairs of samples whose nearest neighbors are each other and belong to different classes, and removing the one that belongs to the frequent class.

Over / under-sampling is one of the most common and easily considered operations. However, there are some drawbacks involved. For example, in the case of over-sampling the tail classes, it may lead to over-fitting~\cite{chawla2002smote, wang2014hybrid} the tail classes~\cite{cui2021reslt, sinha2020class, tan2020equalization, cui2019class} and if there are errors or noise in the samples of the tail classes, then over-sampling may aggravate these problems. Under-sampling may lead to under-learning of the head classes~\cite{sinha2020class, cui2019class}, and may potentially missing valuable data in the head classes. For extremely long-tailed data, the under-sampling method usually loses a lot of information because of the large difference in the amount of data between the head class and the tail class~\cite{tan2020equalization}.

\subsubsection{Data Augmentation}
Data augmentation is another way of data processing to solve the problem of long-tailed distribution, as shown in Fig.~\ref{data_processing} (c). Due to the small sample size of the tail classes, it is difficult to learn the complete features. Therefore the tail classes can be compensated by data augmentation methods such as generating and synthesizing new samples with the help of similar samples~\cite{chawla2002smote} or other data sources~\cite{he2008adasyn,gidaris2018dynamic}. There are some common ways of data augmentation such as random image flipping, scaling, rotating and cropping and so on. But these naive methods are not good enough for the tail classes where samples and features are extremely sparse. In order to reduce the over-fitting risk of the tail classes and to improve the generalization ability, some methods expand the tail classes by data synthesis, which can be divided into two approaches: image space and feature space.

For image space, some data augmentation methods help to improve the performance for tail classes. 
Zhang \etal~\cite{zhang2017mixup} proposed that although \emph{Empirical Risk Minimization (ERM)} allows large-scale neural networks to memorize (rather than generalize) training data, validation on samples outside the training distribution (adversarial samples) can dramatically change the prediction results, \ie, when the distribution of the test set are different from the training set, the \emph{ERM} method no longer has good interpretation and generalization performance. In contrast, data augmentation methods can improve the generalization of the model to the training data~\cite{simard1998transformation}. \emph{Mixup} is a data-independent data augmentation approach that performs data augmentation by constructing a virtual training sample, expressed by the formula:
\begin{equation}
	\begin{aligned}
		\tilde{x}=\lambda x_{i}+(1-\lambda)x_{j},\\
		\tilde{y}=\lambda y_{i}+(1-\lambda)y_{j}.
	\end{aligned}
\end{equation}
where $(x_{i},y_{i})$ and $ (x_{j},y_{j}) $ are two samples randomly selected from the training data and $ \lambda \in [0,1] $.
Recently, Chou \etal~\cite{chou2020remix} improved the mixup method and proposed a new data augmentation method \emph{Remix}. It assigns the label in favor of the minority class by providing a distributed higher weight to the minority class, which makes the classifier push the decision boundary to the majority class and balance the generalization error between the majority class and the minority class. The formulation of \emph{Remix} is:
\begin{equation}
	\begin{aligned}
		\tilde{x}^{RM}=\lambda_{x} x_{i}+(1-\lambda_{x})x_{j},\\
		\tilde{y}^{RM}=\lambda_{y} y_{i}+(1-\lambda_{y})y_{j}.
	\end{aligned}
\end{equation}
where $ \lambda_{x} $ is sampled from the beta distribution and $ \lambda_{y} $ takes the form of:
\begin{equation}
	\lambda_{y}=\left\{
	\begin{array}{lr}
		0,\ n_{i}/n_{j}\geq\kappa\ and\ \lambda<\tau;\\
		1,\ n_{i}/n_{j}\leq1/\kappa\ and\ 1-\lambda<\tau;\\
		\lambda,\ otherwise
	\end{array}\right.
\end{equation}
where the hyper-parameters $\kappa$ and $\tau$ are used to set synthetic labels by comparing the number relationship between samples and to control the degree of synthetic labels, respectively.

Some studies are based on the feature level for feature synthesis, which can enrich the features of the tail classes and create clearer decision boundaries. At the beginning of this century, some classic data synthesis work provided ideas for some current advanced research.\emph{SMOTE}~\cite{chawla2002smote} is an effective method for data synthesis. For each minority class sample, a sample is randomly selected from its nearest neighbors, and then a point on the line between these two samples is randomly selected as the newly synthesized minority class sample. However, \emph{SMOTE} method has some drawbacks, it has some blindness in the selection of nearest neighbors, and it is also prone to the problem of distribution marginalization due to the imbalance of the data. Therefore there are many works to improve it. \emph{Borderline-SMOTE}~\cite{han2005borderline} judges the boundary samples for the minority class samples and generates new samples for the boundary samples, and the rule for judging the boundary samples is that more than half of the K-nearest neighbors of the sample are majority class samples. Another classical approach is \emph{ADASYN}~\cite{he2008adasyn}, which can adaptively decide how many synthetic samples to generate for each minority class based on the distribution of the samples. First the degree of imbalance as well as the number of new synthetic samples to be generated are calculated, then the distribution of each minority class sample is calculated and the distribution is used to determine the number of synthetic samples for each class.

In recent years, some studies use data synthesis and feature synthesis to solve the long-tailed problem. Over-sampling the tail classes is a very common strategy. However, this can easily lead to tail classes over-fitting. Towards this question, Kim \etal~\cite{kim2020m2m} utilize an optimization phase so that the head class samples are modified into tail class samples and then added to the original dataset for the purpose of balancing the dataset. Chu \etal~\cite{chu2020feature} decompose the class activation map into class-generic features and class-specific features. In order to make up for the missing information of the tail classes, the specific features of the tail classes are fused with the common features of the head classes, which can expand the feature space and generate augmented samples to recover the base distribution of the tail classes. Finally, the online generated augmented samples are used to fine-tune the network trained in the first stage to improve the performance of the tail classes. \emph{FASA}~\cite{zang2021fasa} generates virtual features by obtaining the mean and standard deviation of the corresponding class features. And the number of generated virtual features is dynamically decided by an adaptive feature sampling scheme, thus effectively avoiding over-fitting and under-fitting triggered by feature augmentation. \emph{Breadcrumbs}~\cite{liu2021bread} proposes a new feature augmentation strategy that tracks features backwards to access the large number of feature vectors available for each training image from previous epochs, in a way that is more diverse than the features obtained by simply copying and pasting. To overcome the lack of discriminative information in existing re-sampling methods, Zhang \etal propose a novel data augmentation approach based on \emph{Class Activation Maps (CAM)}~\cite{zhou2016learning}, which is tailored for two-stage training and generates discriminative images by transferring foregrounds while keeping backgrounds unchanged.

Data augmentation aims at applying enhancement techniques in the image space or feature space to synthesize new samples, thus expanding the data in knowledge-poor tail classes~\cite{zang2021fasa}. We need to acknowledge that the data produced by the data synthesis approach is more economical and efficient in many cases, especially when the data is difficult to obtain, and can also complement the real-world data. However, this artificial way of creating data does not come from real scenarios, and its impact on the model lacks theoretical guidance, and the synthesis of data close to real scenarios is a highly complex operation, such as VAE~\cite{kingma2013vae, aaron2017vgvae} and GAN \etc~\cite{ian2014gan, andrew2018large, karras2019stylegan}. At the same time, the simple data synthesis method cannot accurately avoid the adverse effects of noise and other undesirable factors from the original dataset.

\subsection{Cost Sensitive Weighting}

\begin{figure}[htbp]
\begin{center}
	\includegraphics[width=0.65\linewidth]{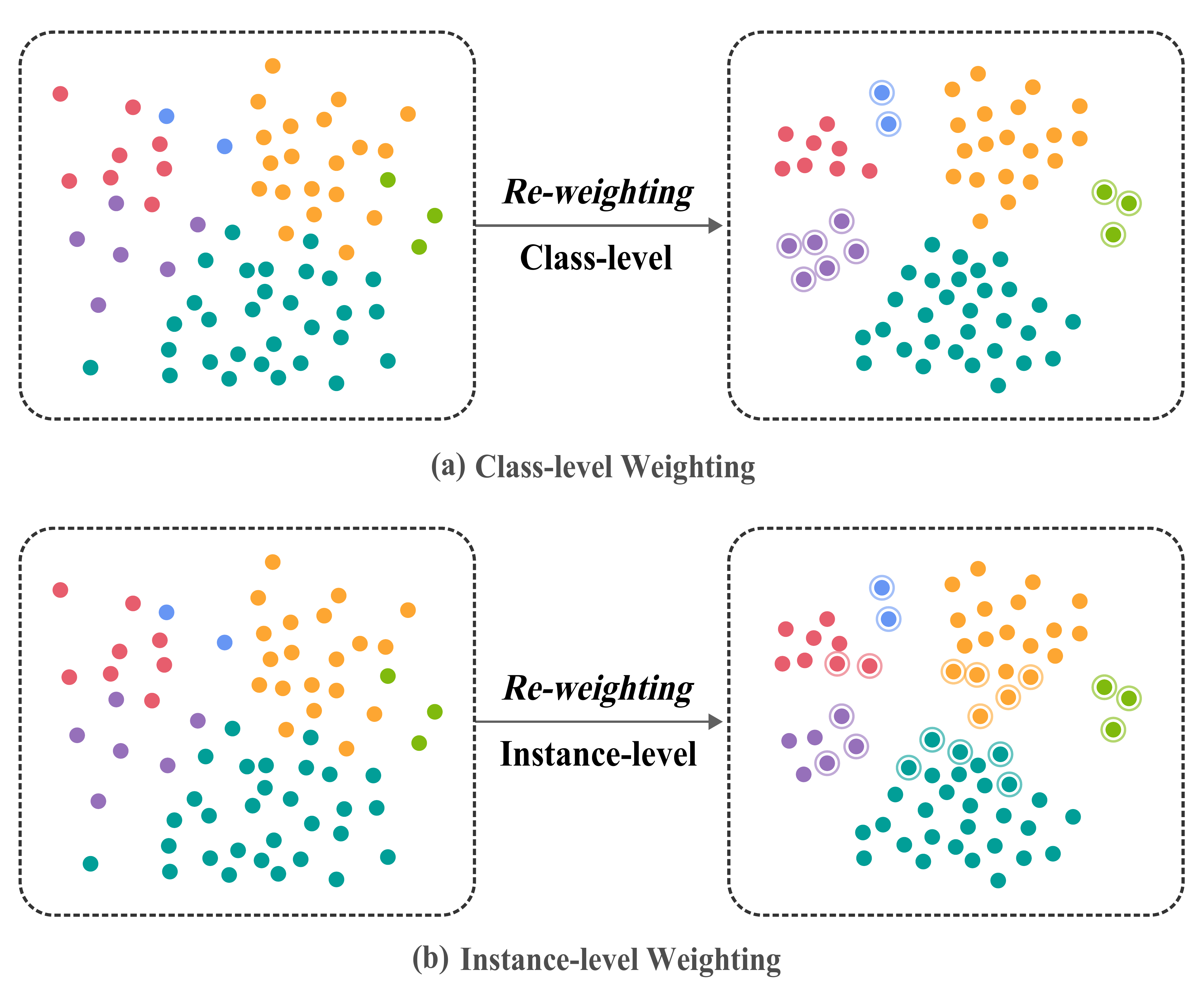}
\end{center}
\vspace{-1mm}
\caption{\textbf{Cost sensitive weighting methods for long-tailed problem}. We use a ring to represent increasing the learning weight of data points. (a) Represents class-level re-weighting, \ie, assigning different learning weights to different classes so that the model enhances the learning of the tail classes. (b) represents instance-level re-weighting, \ie, adjusting the sample weights by controlling at a more fine-grained level to make model focus more on learning hard samples.}
\vspace{-1mm}
\label{cost_sensitive_weighting}
\end{figure}

Cost sensitive learning can be traced back to a classical approach in statistics that considers the cost of misclassified samples, some studies refer to this as importance sampling~\cite{kahn1953methods, cui2019class}. It is shown that there is a strong link between cost sensitive learning and imbalance learning~\cite{he2009learning}, so this approach can be naturally used to deal with extremely unbalanced data like long-tailed data. As shown in Fig.~\ref{cost_sensitive_weighting}(a), cost sensitive weighting assigns different weights to different classes in an explicit or implicit way, so that the influence of the tail samples can be improved. Cost sensitive weighting can also assign weights at the sample level for more fine-grained control, as shown in Fig.~\ref{cost_sensitive_weighting}(b). This strategy can be applied in many directions besides long-tailed learning, such as the classification of foreground and background for detection task.

\subsubsection{Class-level Re-weighting}
For cost-sensitive weighting, one of the most intuitive approaches is to re-weight classes proportionally by the inverse of their frequencies~\cite{huang2016learning, wang2017learning}. But this naive method tends to perform poorly, some works tend to use a “smoothed” version of weights that are empirically set to be inversely proportional to the square root of label frequency~\cite{mikolov2013distributed,mahajan2018exploring}.

However, it is often inefficient to set weights directly based on the instances number of the classes for class-level re-weighting methods, and it is difficult to find an appropriate weight that is valid. Therefore, some methods implicitly distinguish between head and tail classes, and turn their attention to other factors such as effective number of samples, contribution gradient, sample difficulty, \etc

Cui \etal~\cite{cui2019class} argue that there is information overlap among data, as the number of samples increases, the marginal benefit a model can extract from the data diminishes. Based on this view the concept of effective number of samples was proposed. Where \emph{Effective Number} $ E_{n} $ is defined as 
\begin{equation}
E_{n}=(1-\beta^{n})/(1-\beta), where\  \beta=(V-1)/V,
\end{equation}
$n$ is sample number, $ V $ is the total volume of all possible data in the class. \emph{Class-Balanced Loss} solves the training problem for unbalanced data by adding a class-balanced weighting term $ \alpha_{i} \propto 1/E_{n_{i}} $ to the loss function for class $ i $ that is inversely proportional to the number of valid samples, where $n_{i}$ is the number of samples for class $i$.

Tan \etal \cite{tan2020equalization} argue that negative gradients from the head classes severely inhibit the learning of tail classes during training. For the tail classes, the gradients from negative samples are larger than those from positive samples. Therefore, \emph{Equalization Loss (EQL)} sets a weight term $w$ for class $j$:
\begin{equation}
	w_{j}=1-E(r)T_{\tau}(f_{j})(1-y_{j}).
\end{equation}
For a region proposal $r$, $E(r)$ outputs 1 means $r$ is a foreground region proposal and otherwise 0. $f_{j}$ means frequency of class $j$, and $T_{\tau}(x)$ is a threshold function. $y$ is the ground truth distribution with one-hot representation. It aims to ignore the gradients from frequent classes on rare classes while preserving the gradients from background samples, thus ensuring a fair training for each class. 
However, \emph{DropLoss}~\cite{hsieh2021droploss}  observes that most of the gradients that inhibit the tail class actually come from correct background classification rather than incorrect foreground prediction, and therefore \emph{DropLoss} is proposed to adaptively rebalance the ratio of background prediction loss between the rare/common class and the frequent class.
Tan \etal ~\cite{tan2021equalization} propose \emph{EQL v2} from the gradient perspective. It chooses the gradient statistic as an indicator to indicate whether a task is in balanced training. The ratio of accumulated positive gradients to negative gradients for each classifier is used to independently increase the weight of positive gradients and reduce the weight of negative gradients for each classifier. Li \etal ~\cite{Li2022eod} extend the idea of equalization loss to the single-stage object detector~\cite{lin2017focal}, independently rebalances the loss contribution of positive and negative samples of different categories according to their imbalance degrees, and effectively solve the long-tailed problem under the imbalance of positive and negative samples.

Wu \etal~\cite{wu2020distribution} propose that for the multi-label recognition problem under long-tailed distribution, the general re-sampling scheme leads to undesirable effects due to the presence of label co-occurrence. Therefore, \emph{Distribution-Balanced Loss} is proposed to address the undesirable effects caused by label co-occurrence, while over-suppression of negative labels is overcome by regularization to mitigate the tail classes over-fitting problem. 
Peng \etal~\cite{peng2020large} address the case where multiple labels explicitly exist for an object. In order to avoid the traditional softmax function suppressing coexisting classes, concurrent softmax is proposed to avoid unnecessarily large losses due to the multi-label problem, and the gradient could focus on more valuable knowledge.

Sinha \etal~\cite{sinha2020class} propose \emph{Class-Wise Difficulty-Balanced Loss (CDB loss)} to assign loss weights by measuring the learning difficulty for each class. \emph{Seesaw Loss}~\cite{wang2021seesaw} sets a mitigation factor, and the penalty for the class with fewer instances is dynamically adjusted according to the ratio of the number of instances in the tail class to the instance number in the head classes. \emph{Federated loss}~\cite{zhou2021probabilistic} is proposed for solving the federal annotation of LVIS. It selects a subset of classes for each training image, including all positive annotations as well as a random negative subset. A binary Cross-Entropy loss is used for all classes in this subset during training, and classes outside are ignored. Wang \etal~\cite{wang2021adaptive} propose to treat all object classes as tail classes regardless of the instance number of each class. In addition, \emph{Adaptive Class Suppression Loss (ACSL)} is introduced to adaptively balance the negative gradients between different classes, which can effectively improve the discriminative power for the tail classifier. \emph{ResLT}~\cite{cui2021reslt} is rebalanced from a parameter space perspective. The shared part of the model parameters is used to learn the classes' common features. The dedicated part retains the specific capacities of the head, middle, and tail classes through three branches, where the main branch learns to recognize images from all classes, and then augments images from the middle+tail and tail classes through two other residual branches to progressively augment the classification results on the tail classes, respectively. Finally, the branches are aggregated into a final result by additive shortcuts, which is an adaptive, incremental learning method.

\subsubsection{Instance-level Re-weighting}
Models have difficulty in learning on hard examples as well as their features, so some studies have identified these hard samples for targeted treatment. Although hard example mining are not specifically designed for the long-tailed problem, some studies~\cite{cui2019class, dong2017class, jamal2020rethinking} illustrate the effectiveness of instance-level re-weighting for long-tailed learning. For long-tailed data, the extreme imbalance of the data can lead to the tail classes learning fewer iterations and gradually becoming a kind of hard-to-score sample, so the instance-level reweighting methods can be effective in improving the performance of the tails.

In \emph{OHEM}~\cite{shrivastava2016training}, each example is scored by its loss, non-maximum suppression (NMS) is then applied, and a mini-batch is constructed with the highest-loss examples. Positive and negative samples are taken as 3:1 to calculate the loss, and the other negative sample weights are set to 0. Although the \emph{OHEM} algorithm increases the weights of hard samples, it ignores the samples that are easy to classify. Lin \etal~\cite{lin2017focal} propose \emph{Focal Loss} to solve the problem of severe imbalance in the ratio of positive and negative samples in one-stage object detection. It improves on the Cross-Entropy loss by adding a modulating factor, which distinguishes between simple and hard samples. The weight of the simple samples is reduced, while paying more focus on the hard samples. As hard samples are mostly composed of tail classes for the long-tailed data, so \emph{Focal Loss} can effectively improve the learning of tail classes.

The essential effect of sample imbalance from the perspective of gradient distribution is explored by \emph{Gradient Harmonizing Mechanism (GHM)}~\cite{li2019gradient}, which points out that in one-stage detectors, the number of simple samples is very large, so they tend to dominate the model update, and since they are already well discriminated by themselves, the parameter update caused by this part does not improve the model much and the gradients generated by the samples are small. The class imbalance can thus be attributed to an imbalance in the degree of difficulty, and the imbalance in the degree of difficulty can be attributed to an imbalance in the distribution of the gradient parametrization. Specifically, by counting the gradients of the samples and designing \emph{GHM} based on this distribution, the gradients generated by different samples are weighted so as to change the amount of their contributions and eliminate the negative effects of outliers. Zhao \etal designe an \emph{adaptive logic adjustment (ALA) loss}~\cite{zhao2021ala}, which contains an instance-specific adjustment term that adapts to the logic of each sample and can make the model more focused on hard samples.

The re-weighting approach is an important strategy to solve the long-tailed problem by giving different learning weights to different categories or samples. However, it is not feasible to set simple learning weights, such as the inverse of the category frequency or a smoothed version, based only on the size of the category data. Therefore finding a loss weight that fits the model and the data takes effort. Some work assigns weights at the instance level for more fine-grained control. However, some researchers~\cite{sinha2020class} point out that this training strategy still results in a focus on learning the head classes because the absolute number of head data is dominant so that the number of hard samples in the head classes is still more than the number of hard samples in the tail classes. For large-scale real-world data, re-weighting tends to make the deep model difficult to optimize during training~\cite{huang2016learning, huang2019deep}. Moreover, re-weighting methods are susceptible to sensitive hyper-parameters, and the optimal settings may vary widely from dataset to dataset~\cite{tan2021equalization}.

\subsection{Decoupling Methods}

\begin{figure}[htbp]
\begin{center}
	\includegraphics[width=0.99\linewidth]{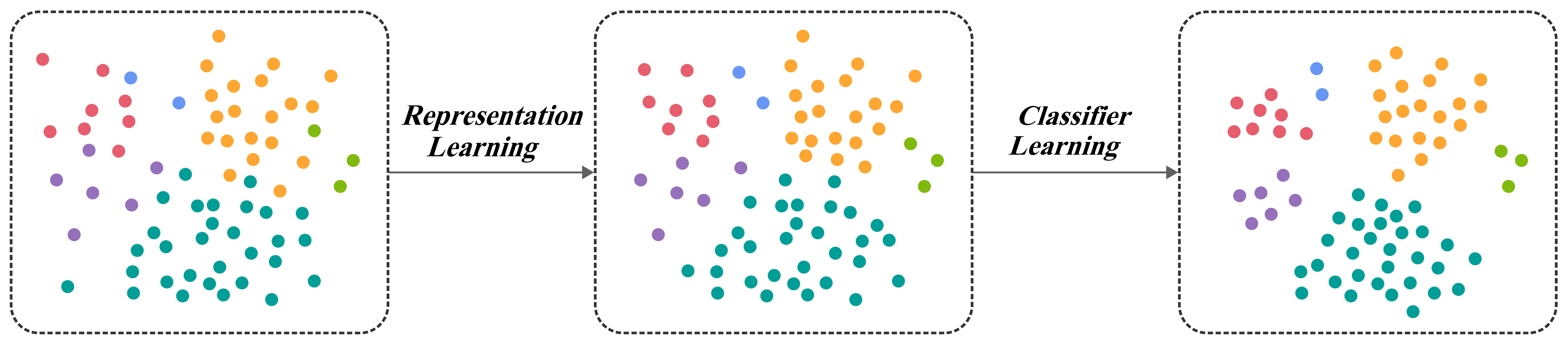}
\end{center}
\vspace{-1mm}
\caption{\textbf{Decoupling methods for long-tailed problem}. Because the rebalancing approach can severely damage the learned representations, some studies decouple representation learning and classifier learning. In the first stage, ordinary learning is used for representation learning. In the second stage, the network parameters of the representation learner are frozen and using a rebalancing approach to learn a good classifier.}
\vspace{-1mm}
\label{decoupling}
\end{figure}

Some studies found that although rebalancing strategies are important for long-tailed data, manipulating the data by re-sampling or re-weighting methods can harm the feature representation during the representation learning phase, while regular sampling tends to give more general representations. Therefore, uniform sampling is used to train the deep learning model to obtain the features of the data, and then class-balanced sampling is performed on the classifier to balance the head and tail classes, as shown in Fig.~\ref{decoupling}.

Kang \etal~\cite{kang2019decoupling} propose that unbalanced data would not be a problem in learning high-quality feature representation, while strong long-tailed recognition could be achieved by tuning the classifier only. The learning process is decoupled into representation learning and classifier learning. The former is learned by standard instance-balanced sampling, class-balanced sampling, or a mixture of the two, the results shows that instance-balanced sampling yields better results, which shows that the rebalance methods can impair the learned representations. The latter learns the classifier by three methods, namely \emph{Classifier Re-training} (\emph{cRT}), \emph{Nearest Class Mean classifier} (\emph{NCM}), \emph{$\tau$-normalized classifier} (\emph{$\tau$-normalized}). 

Inspired by~\cite{kang2019decoupling}, \emph{BAGS}~\cite{li2020overcoming} also decouple the representation learner and classifier. The balanced group softmax module is introduced in the classification head of the detection framework. Grouping according to the instance number of the classes, and respectively executing softmax operation. The group-by-group training is used to separate the classes with disparate numbers of instances, thus balancing the classifiers in the detection framework and effectively reducing the control of the head classes over the tail classes.

\emph{BBN}~\cite{zhou2020bbn} designed a conventional learning branch as well as a re-balancing branch, where the former learns the generic pattern of the original distribution from the original long-tailed data, while the latter models the tail data in a back-sampling manner. Finally, the weights of the feature vectors of the two branches $ f_{c} $ and $ f_{r} $ are controlled by the adaptive trade-off parameter $ \varphi $ and input to the two classifiers $ W_{c} $ and $ W_{r} $, respectively, and combined to obtain the final logits results $z$:
\begin{equation}
	z=\varphi {W_{c}}^\top f_{c} +(1-\varphi) {W_{r}}^\top f_{r}.
\end{equation}

\emph{SimCal}~\cite{wang2020devil} corrects for biases in the classification head through a decoupled learning scheme. The model is first trained normally. A \emph{bi-level sampling} scheme combining image-level and instance-level sampling is then used to collect class-balanced training instances. These samples are then used to calibrate the classification head to improve the tail classes performance. To mitigate the adverse effects of the above calibration on the head classes, \emph{SimCal} also proposes a \emph{Dual Head Inference} architecture that selects predictions for the tail and head classes directly from the new balanced classifier head and the original head.

\emph{DisAlign}~\cite{zhang2021distribution} believes that existing two-stage learning methods usually rely on heuristic design to adjust the initially learned classifier, which requires lengthy hyper-parameter tuning. At the same time, the bias of the decision boundary in the feature space can become a bottleneck. To this end, an adaptive method is designed to calibrate the output of the classifier, and a generalized weighting method is used to balance the class prior, so that the classifier output is matched to the reference distribution of the class that is beneficial to balance the prediction to calibrate the output of the classifier.

\emph{MiSLAS}~\cite{zhong2021improving} proposes label-aware smoothing to deal with different degrees of class over-confidence in order to solve the mis-calibration problem of the two-stage method. Shifted batch in the decoupling framework is further proposed for the deviation of the dataset between the two stages due to different samplers normalization.

\emph{LADC}~\cite{wang2021ladc} suppose that tail classes can be enriched by similar head classes and proposes a novel distribution calibration approach, which transfers the statistics from relevant head classes to infer the distribution of tail classes in the second stage.

Recently, decoupling representation learner and classifiers method has been shown to be effective in long-tailed data distributions, and it has become one of the mainstream research directions of long-tailed recognition. Moreover, decoupling method is relatively convenient to use in combination with data processing and cost sensitive weighting methods, which can obtain better model learning effect~\cite{li2020overcoming, zhang2021bag}. But the two-stage learning strategy defies the expectation of end-to-end training sought in deep learning. At the same time, the resampling or re-weighting method adopted in the second stage still has the limitations mentioned above.

\subsection{Other Long-tailed Visual Recognition Methods}
\begin{figure}[htbp]
\begin{center}
	\includegraphics[width=0.75\linewidth]{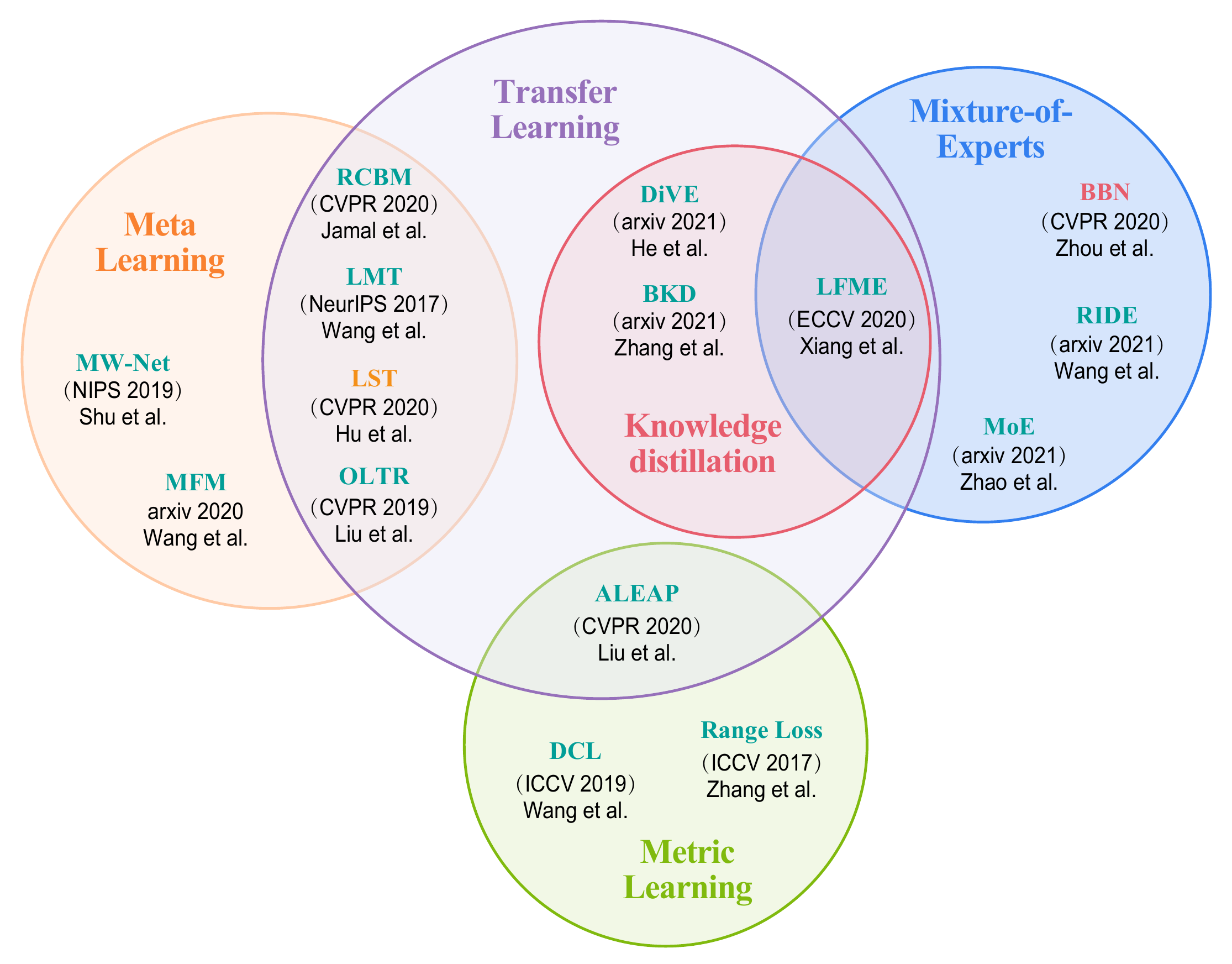}
\end{center}
\vspace{-1mm}
\caption{\textbf{Domain relevance of some long-tailed methods}. To illustrate the domain relevance of the long-tailed approach, we list some of the mainstream studies, which draw on knowledge from multiple research domains.}
\vspace{-1mm}
\label{additional}
\end{figure}

In addition to the above methods, many studies use one or mixed multiple machine learning methods (metric learning, transfer learning, meta learning, knowledge distilling, mixture-of-experts, \etc) to solve the long-tailed problem, as shown in Fig.~\ref{additional}.

\subsubsection{Metric Learning}

\begin{figure}[htbp]
\begin{center}
	\includegraphics[width=0.65\linewidth]{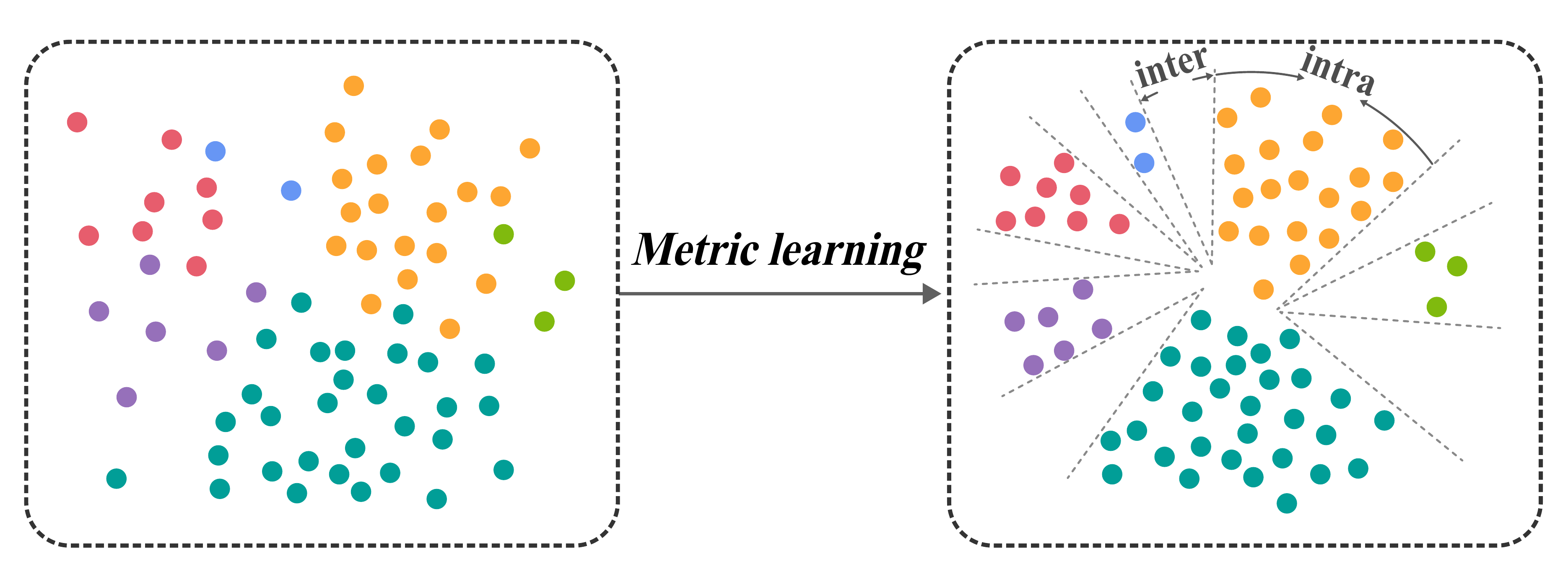}
\end{center}
\vspace{-1mm}
\caption{\textbf{Metric learning for long-tailed problem}. The method aims to clarify the decision boundary between classes while expanding the inter-class distance and reducing the intra-class distance.}
\vspace{-1mm}
\label{metric_learning}
\end{figure}

Long-tailed data can be further clarified by performing feature similarity metrics in feature space with the help of metric learning to specify the boundaries between classes. Metric learning~\cite{oh2016deep, sohn2016improved}, also known as \emph{Distance Metric Learning (DML)}, which aims to learn an embedding function that can embed data to a feature space where the inter-data relationships are preserved ~\cite{sinha2020class}. Specifically, metric learning is a spatial mapping method. By learning to an embedding space in which all data converted to feature vectors are measured for similarity and distances are reduced for similar samples and expanded for dis-similar samples, as shown in Fig.~\ref{metric_learning}. This is helpful for classifiers to classify certain tail classes where the decision boundary is not clear enough.

Zhang \etal~\cite{zhang2017range} propose \emph{Range Loss}, the intra-class compactness constraint minimizes the two maximum intra-class distances for each class. The inter-class separation constraint calculates the centers of each class and makes the two classes with the smallest class center distances greater than a set margin. \emph{Class Rectification Loss (CRL)}~\cite{dong2018imbalanced} argued that the traditional Cross-Entropy loss is not applicable to data with high imbalance and can cause model suffers from generalizing inductive decision boundaries biased towards majority classes. Therefore, \emph{CRL} is proposed to gradually enhance the decision boundaries of minority classes. Specifically, the triplet ranking loss~\cite{liu2011learning} is considered to model the relative relationship constraints within and between classes. In order to fully learn and exploit the minority classes, each minority class sample is considered as an "anchor" in the triplet structure to compute the batch loss balance regularization. \emph{Dynamic Curriculum Learning (DCL)}~\cite{wang2019dynamic} argues that this way of setting anchors tends to cause feature push-pull confusion.Therefore only simple samples of minority classes are set as anchors, instead of all samples of minority classes. Thus, the problem of sample feature push and pull confusion can be avoided to a certain extent when the decision boundary is blurred. 

Cao \etal~\cite{cao2019learning} believe that encouraging a large margin can be viewed as regularization, and propose to regularize the minority classes more strongly than the frequent classes. Therefore, minority classes are encouraged to obtain higher margins. Assume that $ \gamma_{i} $ is the margin of class $ i $ and $ n_{i} $ is the number of samples of class $ i $. \emph{LDAM} designs a \emph{label-distribution-aware loss} function that finds the best trade-off between the margins of the classes: 
\begin{equation}
	\gamma_{i} \propto {n_{i}}^{-1/4}
\end{equation}
and forces it to be a multi-class class-dependent margin.

\emph{Hybrid SC/PSC}~\cite{wang2021contrastive} is designed to include a \emph{supervised contrastive learning (SCL)-based} feature learning branch and a Cross-Entropy loss based classifier learning branch. The hybrid network structure progressively adjusts the weight of the two branches in the learning process, and jointly performs feature learning and classifier learning. \emph{Prototypical supervised contrastive loss} is designed to learn the prototype of each class for comparative learning and to force the different augmented views of each sample to be close to the prototype of their class and away from the prototype of the remaining classes.

Wang \etal~\cite{wang2021marc} study the relationship between the margins and logits (classification scores) and empirically observe the biased margins and the biased logits are positively correlated. Based on this observation, they propose \emph{Margin Calibration (MARC)}, a simple yet effective margin calibration function to dynamically calibrate the biased margins for unbiased logits.

Cui \etal~\cite{cui2021paco} observe supervised contrastive loss tends to bias on high-frequency classes and thus increases the difficulty of imbalance learning. Therefore, \emph{Parametric Contrastive Learning (PaCo)} is proposed to solve the problem of long-tailed recognition. \emph{PaCo} introduces a set of parametric class-wise learnable centers to rebalance from an optimization perspective. When more samples are pulled together with their corresponding centers, \emph{PaCo} can adaptively enhance the intensity of pushing the same class of samples closer, and is conducive to hard example learning. In addition, this study also found that RandAugment~\cite{cubuk2020randaug} and longer training epochs can further improve the effect of long-tailed learning, and achieved very competitive accuracy on multiple benchmarks combined with \emph{PaCo}.

The basic purpose of metric learning is to make the sample features of similar classes closer together and those of different classes farther apart. Metric-based learning methods are usually based on loss functions to perform metrics between features, so it is necessary to consider the appropriate way for the combination of training samples as well as to choose the appropriate loss function. And for the head classes with large absolute numbers, it is still necessary to consider how to avoid the bias of the model for the head classes~\cite{wang2021contrastive}.

\subsubsection{Transfer Learning}

\begin{figure}[htbp]
\begin{center}
	\includegraphics[width=0.99\linewidth]{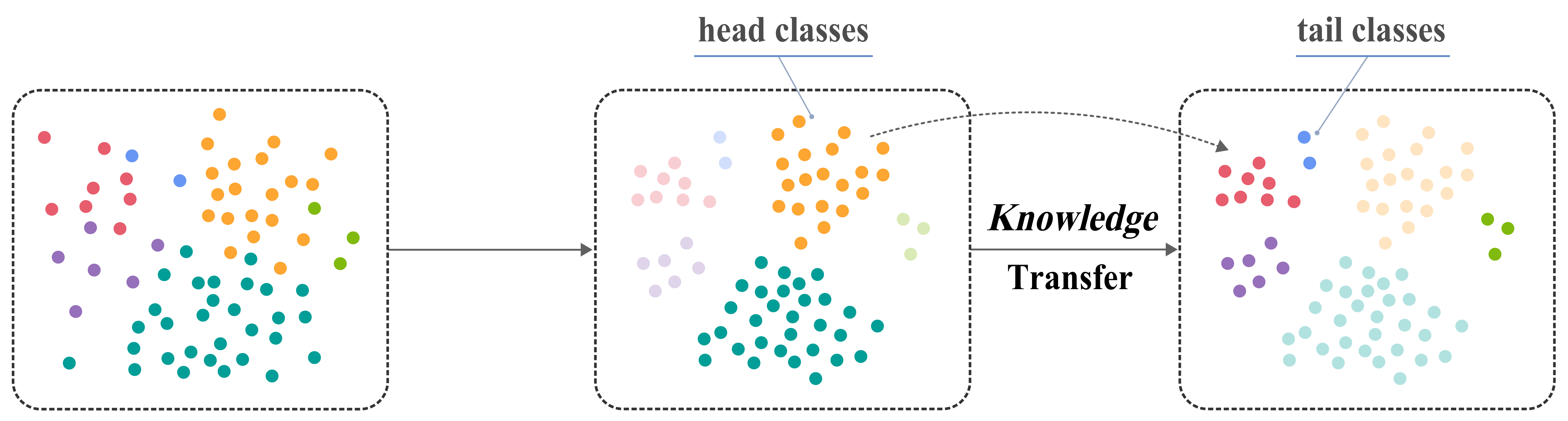}
\end{center}
\vspace{-1mm}
\caption{\textbf{Transfer learning for long-tailed problem}. The method aims to make full use of the sufficient head classes and to transfer the knowledge acquired from the head classes to the tail classes.}
\vspace{-1mm}
\label{head_to_tail}
\end{figure}

Due to the large absolute amount of data in the head classes of the dataset, it has a richer and more complete training resource compared to the tail classes. Therefore some studies hope the head knowledge can be fully utilized to guide the learning of feature under-represented tail classes, as shown in Fig.~\ref{head_to_tail}.

Wang \etal~\cite{wang2017learning} emphasis the strong correlation between meta-networks and model parameters. The basic assumption is proposed that model parameters share similar dynamics across classes, laying the foundation for transferring model parameters between classes. Specifically, this research constructs a meta-network with a deep residual network as the basic unit to learn the head classes on the model parameter space and gradually transfer the meta-knowledge to the tail classes.

Hu \etal~\cite{hu2020learning} divide the LVIS dataset of more than one thousand classes into sections according to the number of instances within the classes. The learning and merging of each part was performed by a divide-and-conquer strategy. A knowledge distillation strategy is used in order to retain the knowledge learned from the head classes. And a \emph{Meta Weight Generator (MWG)} is designed to dynamically generate the weight matrix of the current stage using the fundamental knowledge learned and inherited from the head classes.

Due to the small spatial span of the tail classes and the large spatial span of the head classes, in order to compensate for the intra-class diversity of the tail classes, Liu \etal~\cite{liu2020deep} construct to a feature cloud for each feature, transferring from the head classes to extend the distribution of the tail classes. The ideas of \emph{CosFace}~\cite{wang2018cosface} and \emph{ArcFace}~\cite{deng2019arcface} are borrowed to learn corner features. The overall variance of the head classes is obtained by computing the angle distributions and means between the head class features and their corresponding class centers. And the angular variance of the head classes is passed to the tail classes by constructing additional distributions, thus constructing feature clouds for each tail instance and expanding the space of the tail classes. \emph{GistNet}~\cite{liu2021gist} implements geometric structure transfer by implementing constellations of classification parameters, transferring the geometric structure of the head classes to the tail classes.

In order to make full use of the knowledge in the head classes and compensate for the lack of knowledge in the tail classes, some work uses transfer learning to transfer knowledge from the head classes to the tail classes. How to improve the performance of the tail classes without damaging the performance of the head classes is one of the issues to be considered. Meanwhile, some works usually carry out complex model and module design for knowledge transfer~\cite{wang2017learning}, which is not conducive to the training of the model and the convergence of the network.
 
\subsubsection{Meta Learning}

\begin{figure}[htbp]
\begin{center}
	\includegraphics[width=0.65\linewidth]{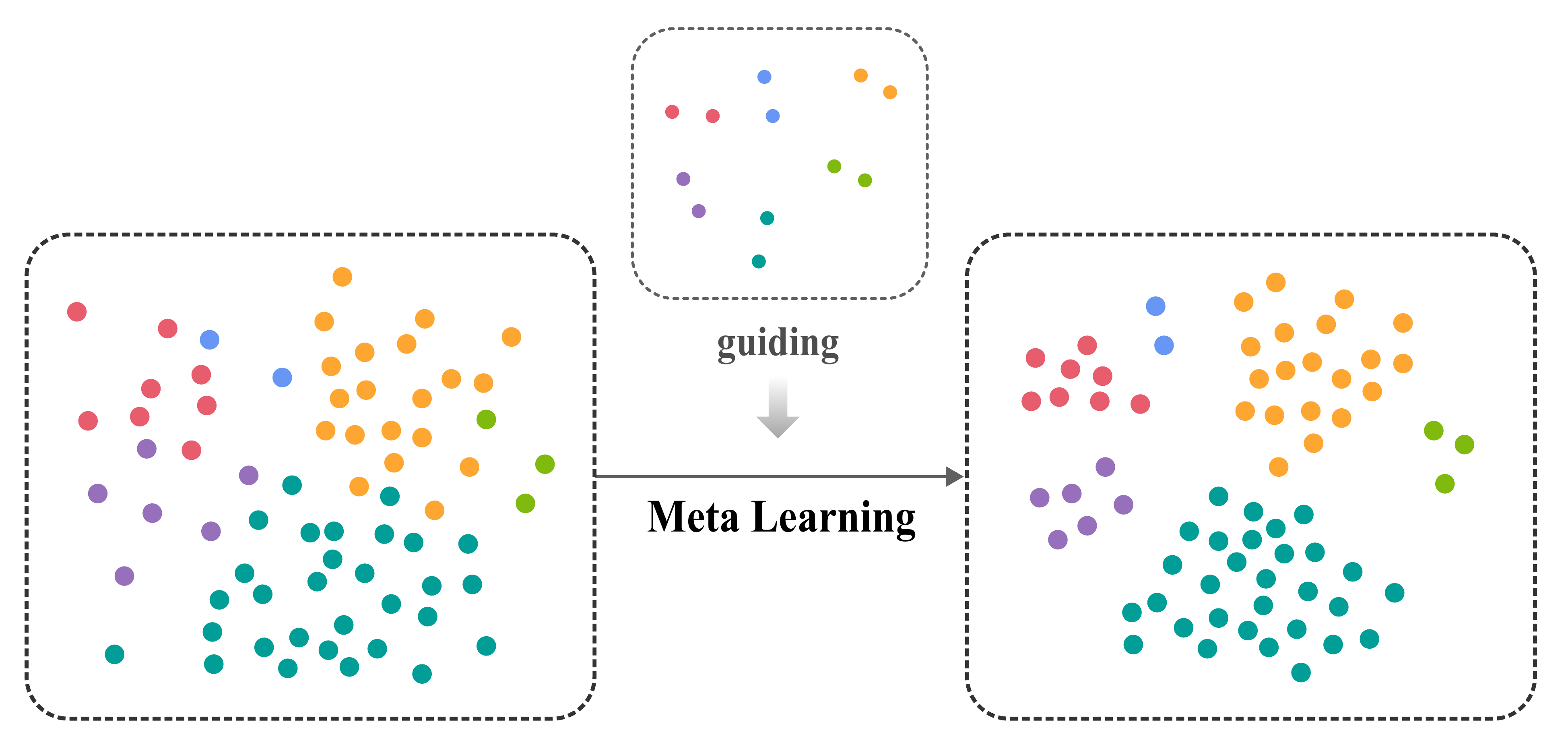}
\end{center}
\vspace{-1mm}
\caption{\textbf{Meta Learning for long-tailed problem}. Deep models are trained adaptively by constructing a small amount of balanced meta-data, building meta-models, or setting adaptive parameters.}
\vspace{-1mm}
\label{meta_learning}
\end{figure}

Meta Learning, also called Learning to Learn, is another important branch of research after Reinforcement Learning~\cite{sutton2018reinforcement, kaelbling1996reinforcement}. It aims to learn a model that acquires general knowledge from different tasks and equips the model with the ability to learn in order to quickly adapt to new tasks. In long-tailed problem, meta-learning can guide the model to train, construct metamaps from head classes to tail classes, learn model parameters adaptively, assign sample weights, adjust classification network features, \etc, as shown in Fig.~\ref{meta_learning}. The difficulty and uncertainty of human intervention to guide the model to learn are alleviated.

Ren \etal~\cite{ren2018learning} propose that there is a tendency to select samples with small training losses in noisy images, and a tendency to select examples with large training losses in data imbalance problems. To address these two conflicting views, \emph{L2RW} guides the updating of the weights of training losses by constructing a small unbiased validation set (\ie, meta-data). To reduce the computational cost, an online approximation method is used to adjust the weights of the training losses online in a mini batch in an approach similar to SGD. Shu \etal~\cite{shu2019meta} argued that the possible way of learning weights implicitly in \emph{L2RW} leads to unstable weighting behavior and non-generalizability during the training process, and therefore adopted the meta-network mechanism of learning weights explicitly, and proposed an adaptive sample weighting strategy for setting up sample loss. Specifically, a simple meta-network, \emph{MW-Net}, is set up so that it adaptively learns the weights of the $ i-th $ sample loss and uses meta-data to guide all parameters.

Jamal \etal~\cite{jamal2020rethinking} propose that the conditional distribution $P_{s}(x|y)=P_{t}(x|y)$ may not hold, which can result in target-shift. For this reason, this research proposes to relax the assumption that the source and target domains share the same conditional distribution $P_{s}(x|y)$ and $P_{t}(x|y)$ to enhance class-balanced learning. Specifically, this research relates the expected error in the target domain to the error in the source domain, and sets a balanced meta-data set to guide the meta-framework in estimating the conditional weights which is set for the target-shift part. Also, Ren \etal~\cite{ren2020balanced} find that in the long-tailed case, according to Bayes theorem, the regular softmax regression is affected by the label distribution shift, which will make the classifier more inclined to consider the samples as belonging to the head class. For this purpose, the work explicitly considers the label distribution shift, and re-derive the softmax function. The final \emph{Balanced Softmax} $\hat{\phi}_{j}$ can be expressed as:
\begin{equation}
	\hat{\phi}_{j}=\dfrac{n_{j}e^{\eta_{j}}}{\sum_{i=1}^{k}n_{i}e^{\eta_{i}}},
\end{equation}
where $n_{i}$ denotes the number of class $i$, and $\eta$ denotes model output.

\emph{Meta feature modulator (MFM)}~\cite{wang2020meta} is proposed to model the difference between long-tailed data and balanced meta-data from a representational learning perspective. Specifically, modulation parameters are introduced to channel-wisely scale or shift the intermediate features of the classification network. The modulation parameters and the classification network parameters are gradually optimized under the guidance of the balanced meta-data. Eventually, the model is made to have similar preferences for all classes.

\emph{MetaSAug}~\cite{li2021metasaug} performs data augmentation with the help of \emph{Implicit Semantic Data Augmentation (ISDA)} technique to obtain more semantically informative features. For tail classes, a reasonable covariance matrix cannot be obtained. Therefore, \emph{MetaSAug} validates a small balanced validation set in each training iteration, minimizing the validation loss to update and learn the appropriate class-level covariance to achieve more meaningful augmentation results.

The meta-learning based approach allows the model to be more automated for adaptive learning. Some work guides the training of models for balance by employing meta-data~\cite{li2021metasaug}, or employs meta-models to learn model parameters adaptively~\cite{shu2019meta}. However, the guidance of meta-data is weak, and the model will inevitably be more inclined to learn head data. At the same time, the design of some meta-models or modules is complicated.

\subsubsection{Mixture-of-Experts}

Some studies use Mixture-of-Experts (MoE)~\cite{jacobs1991adaptive, jordan1994hierarchical} model to train multiple neural networks (\ie, multiple experts), each of which is specialized to be applied to a different part of the dataset. Based on the divide-and-conquer principle, and training multiple expert networks with a routing network~\cite{masoudnia2014mixture}. Each neural network in MoE (\ie, each expert) will have a data region in which it excels and on which it performs better than the other experts. Thus, it precisely solves the problem that needing to treat the head and tail classes differently under large-scale long-tailed datasets.

\emph{RIDE}~\cite{wang2020long} analyzes the prediction of long-tailed classifiers in terms of bias and variance, proposing that model bias measures the prediction accuracy relative to the true value; variance measures the stability of the prediction. Routing diverse experts is proposed to reduce the model variance of the long-tailed classifier by employing multiple experts, and a distributed perceptual diversity loss is set to reduce the model bias. The accuracy of both the head classes and the tail classes can be improved.
Similarly, \emph{Mixture-of-Experts (MoE)}~\cite{zhao2021ala} uses multiple experts to learn diverse results, and then the routing module dynamically integrates the results of multiple experts based on each input instance and trains them jointly with the expert network in an end-to-end manner.

MoE strategies usually requires reliable expert models for better learning of long-tailed data by means of model ensemble strategies. MoE can often achieve very high accuracy, but it always implicitly or explicitly integrates multiple models~\cite{wang2020long}, which brings specific computational load and is a problem that needs to be improved in the follow-up work.

\subsubsection{Knowledge Distilling}

Hinton~\cite{hinton2015distilling} first propose the concept of knowledge distillation. By introducing a teacher network with superior inference performance, a streamlined and low-complexity student model is continuously induced for training, allowing the student model to continuously approach the predictions of the accurate network. For the long-tailed data distribution, knowledge distillation is always used to balance the predictions of head and tail classes.

In~\cite{xiang2020learning}, \emph{LFME} framework is proposed as a self-paced knowledge distillation method. The long-tailed dataset is split into several balanced subsets, and trained with expert models to guide the learning of the student models. A weighting scheme with automatic speed is introduced to train the training data in an easy-to-hard way, which eventually makes the student models outperform the expert models.

\emph{DiVE}~\cite{he2021distilling} uses the output of the teacher model as virtual examples to share knowledge among different classes by knowledge distillation, and proposes that \emph{DiVE} can explicitly adjust the virtual example distribution to become flatter, thus improving the under-represented tail classes.

\emph{Balanced knowledge distillation (BKD)}~\cite{zhang2021bkd} first uses Cross-Entropy loss to train a common teacher model. The student model is then trained by minimizing the combination of instance-balanced classification loss and class-balanced distillation loss.

Knowledge distillation strategies also require reliable expert models for better learning of long-tailed data by means of expert-student model knowledge transfer. Moreover, for knowledge distillation strategies, existing long-tailed approaches usually no longer choose large models as teacher models, but learn better representation by teacher models~\cite{xiang2020learning} or balance data by adding virtual data~\cite{he2021distilling}, but this approach needs to choose appropriate knowledge distillation strategies with reasonable hyper-parameters.

\subsubsection{Grouping}

Some studies group vocabularies according to their relations, in addition to the most general grouping based on the number of instances of the class (\eg, ~\cite{li2020overcoming}, ~\cite{xiang2020learning}, ~\cite{hu2020learning}, \etc), \emph{Forest R-CNN}~\cite{ wu2020forest} uses a priori knowledge of lexical, visual, and geometric relationships to construct classification trees, each of which will contribute to fine-grained classification. For example, when considering lexical relations, "school bus" and "car" have the same parent class "vehicle", while when considering geometric relations, "steering wheel" and "basketball" have the same parent class "roundness". The \emph{Forest R-CNN} is designed to reduce the confidence scores of those classes misclassified by the fine-grained classifier, thus making the model more fault-tolerant in terms of the noise logarithm of the fine-grained classifier.

The grouping strategy requires to group the long-tailed data, and some researchers divide the categories according to the instance number, but this approach is likely to block the knowledge interaction between groups. \emph{Forest R-CNN}~\cite{ wu2020forest} divides the categories based on some semantic relations, but this requires the help of additional knowledge learning. Therefore, the automatic learning grouping strategy based on data itself is a problem worthy of research.

\begin{table*}[htbp]
  \centering
  \caption{\textbf{Highlights of the main methods for solving the long-tailed problem as well as their limitations}. For summary purposes, some methods are not covered in this table.}
  \begin{threeparttable}
   \resizebox{0.99\textwidth}{!}{
    \setlength\tabcolsep{5pt}
    \renewcommand\arraystretch{1.05}
    \begin{tabular}{|c|c|c|c|c|}
     \hline\thickhline
     \rowcolor{purple0}
     Methods                                                          & Strategies                &  Representative Work                                                                &           Highlights             &            Limitations       \\ 
     \hline \hline
     \multirow{6}{*}{\tabincell{c}{Data Processing \\Methods}}   & Over-sampling           & ~\cite{shen2016relay, mahajan2018exploring,  gupta2019lvis, peng2020large, kim2020m2m, liu2021gist}  & \tabincell{c}{Increase the number of samples for \\tail data.}                        & \tabincell{c}{a) Causes over-fitting of the tail class. \\b) Easy to amplify errors or noise present \\in the tail class.}                \\ 
     \cline{2-5}      
                                                                                                 & Under-sampling         & ~\cite{liu2008exploratory, wilson1972asymptotic, devi2017redundancy}  & \tabincell{c}{Deletion of head data.}                        & \tabincell{c}{a) Causes under-fitting of the head class. \\b) It is possible to delete valuable data of \\head class by mistake.}                \\ 
     \cline{2-5}      
                                                                                                 & Data Augmentation         & \tabincell{c}{~\cite{chawla2002smote, han2005borderline, chu2020feature, he2008adasyn, chou2020remix} \\ ~\cite{li2021metasaug, zang2021fasa}}  & \tabincell{c}{The tail data / feature is extended by data \\augmentation.}                        & \tabincell{c}{Inability to introduce new effective \\samples.}                \\                     
     \hline             
      \multirow{7}{*}{\tabincell{c}{Cost Sensitive \\Weighting}}   & \tabincell{c}{Class-level \\Re-weighting}        &  \tabincell{c}{~\cite{hong2021disentangling, huang2016learning, mikolov2013distributed, cui2019class, tan2021equalization} \\~\cite{tan2020equalization, wu2020distribution, wang2021seesaw, zhong2021improving, Li2022eod}}  & \tabincell{c}{Assigning weights to different classes and \\aggravating the learning of tail class.}                        & \tabincell{c}{a) It is difficult to choose appropriate \\weights for each class. \\b) Susceptible to the influence of sensitive \\hyper-parameters. \\c) There may be big differences for \\different data sets.}                \\ 
     \cline{2-5}      
                                                                                                 & \tabincell{c}{Instance-level \\Re-weighting}    & ~\cite{shrivastava2016training, lin2017focal, li2019gradient, zhao2021ala, hsieh2021droploss}  & \tabincell{c}{Assign learning weights to examples \\based on their difficulty.}                        & \tabincell{c}{There is a high probability that the \\number of hard samples in the \\head class will exceed the tail class. So in \\essence, more emphasis will still be \\given to the head class.}                \\      
      \hline             
      Decoupling   & --        & ~\cite{kang2019decoupling, chang2020alpha, wang2020devil, zhang2021distribution, zhong2021improving, wang2021ladc}  & \tabincell{c}{Decoupling representation learning \\and classifier learning.}                        & \tabincell{c}{a) Two-stage learning defies the end-to-end \\pursuit of deep learning. \\ b) In the rebalancing phase, the same problems \\ are faced as in other rebalancing methods.}              \\               
      \hline             
      \tabincell{c}{Metric \\Learning}   & --        & \tabincell{c}{~\cite{dong2018imbalanced, liu2011learning, zhang2017range, liu2020deep, wang2019dynamic} \\ ~\cite{wang2021marc, Li2021tsc, cui2021paco}}  & \tabincell{c}{Learn an embedding space in which \\to measure the similarity of embedded \\features or force a larger margin \\for the tail classes.}                        & \tabincell{c}{a) Select the appropriate distance function \\and measurement method. \\b) The learned embedding function still \\has the risk of biasing towards the \\head classes.}                \\   
      \hline       
       \tabincell{c}{Transfer \\Learning} & \tabincell{c}{--} & ~\cite{wang2017learning, liu2019large, hu2020learning, liu2021gist}  & \tabincell{c}{Transferring the knowledge of head \\class to tail class.}                        & \tabincell{c}{Requiring a more complex model or \\module design, which can make the \\model difficult to train.}                \\ 
     \cline{2-5}      

     \hline             
\tabincell{c}{Meta \\Learning}   & --        & ~\cite{ren2018learning, shu2019meta, ren2020balanced, wang2020meta, li2021metasaug}  & \tabincell{c}{Learn adaptive solutions from data \\or modules to make learning more \\automated.}                        & \tabincell{c}{a) The guidance of meta-data is weak. \\ b) More complex model or module design \\is required.}     \\                                      
   
     \hline
     \tabincell{c}{Mixture-of-Experts}   & --        & ~\cite{zhou2020bbn, zhao2021ala, wang2020long}  & \tabincell{c}{Multi-expert model ensemblling .}                        & \tabincell{c}{ Expert model ensemble requires more \\computational resources, }     \\                                                                         \hline
     \tabincell{c}{Knowledge Distilling}   & --        & ~\cite{xiang2020learning,he2021distilling, zhang2021bkd}  & \tabincell{c}{Guided by the expert model, the \\student model is able to learn the \\data in a balanced manner.}                        & \tabincell{c}{a) Requires reliable expert models for better \\ earning. \\b) Knowledge distillation needs to control the \\parameters of student model learning.}     \\ 
                       
     \hline
     \tabincell{c}{Grouping}   & --        & ~\cite{li2020overcoming, wu2020forest}  & \tabincell{c}{The data are trained in groups \\according to certain relationships.}                        & \tabincell{c}{A suitable grouping method needs to be \\found to ensure as much knowledge \\interaction between the groups as possible \\during training.} \\                    
     \hline
     \tabincell{c}{Semi-supervised}   & --        & ~\cite{yang2020rethinking, wei2021crest, ramanathan2020dlwl, zhang2021simple}  & \tabincell{c}{Semi-supervised learning on long-tailed \\data by introducing other data sources.}                        & \tabincell{c}{Additional data sources are required.}     \\                                                                                                                                                                                                                                   
     \hline
    \end{tabular}
   }
  \end{threeparttable}
  \vspace{-1mm}
 \label{table:highlights}
\end{table*}

\subsubsection{Semi-supervised}

\begin{table*}
	\centering
	\caption{
		{\textbf{Performance (\%) summarization of some representative methods on CIFAR-10/100-LT benchmarks}. The summation of values in Epoch is meant to be a two-stage training strategy. Most methods adopt ResNet-32 as backbone. (In the 2020 and 2021 studies, the three best scores of $\beta$ = 100 and $\beta$ = 10 are marked in \textcolor{scorered}{red}, \textcolor{scoreblue}{blue}, and \textcolor{scoregreen}{green}, respectively.)}
	}
	\vspace{-5pt}
	\label{table:cifar_lt_bench}
	\begin{threeparttable}
		\resizebox{0.98\textwidth}{!}{
			\setlength\tabcolsep{8pt}
			\renewcommand\arraystretch{1.05}
			\begin{tabular}{|c|c|c|c|c|c|c|c|c|c|c|c|c|c|c|}
				\hline\thickhline
				\rowcolor{purple0}
				& & & \multicolumn{6}{c|}{{CIFAR-10-LT (top-1)}} & \multicolumn{6}{c|}{{CIFAR-100-LT (top-1)}}  \\
				\cline{4-15}
				\rowcolor{purple0}
				& & & \multicolumn{6}{c|}{{Imbalance Factor $\beta$}}  & \multicolumn{6}{c|}{{Imbalance Factor $\beta$}} \\
				\cline{4-15}
				\rowcolor{purple0}
				\multirow{-3}{*}{{Year}} & \multirow{-3}{*}{{Method}}  & \multirow{-3}{*}{{Pub.}}  &   200   &   100   &   50   &   20    &   10   &   1   &   200   &   100   &   50   &   20    &   10   &   1   \\
				\hline
				\hline
				
				\hline
				\multirow{3}{*}{{}}
				& Softmax Loss~\cite{he2016deep} & CVPR   &  65.6  & 70.3  & 74.8  & 82.2  & 86.3  & 92.8  & 34.8  &  38.2  &  43.8  &  51.1  &  55.7  &  70.5  \\
				& Focal Loss~\cite{lin2017focal}   & ICCV  & 65.2  & 70.3  & 76.7  & 82.7 & 86.6 & 93.0 & 35.6  & 38.4  & 44.3  & 51.9  & 55.7 & 70.5 \\
				& L2RW~\cite{ren2018learning}    & ICML   & 66.5  & 74.1  & 78.9  & 82.1 & 85.1 & 89.2 & 33.3  & 40.2  & 44.4  & 51.6  & 53.7 & 64.1 \\
				
				\hline
				\multirow{2}{*}{{2019}}
				& CB Loss~\cite{cui2019class}       & CVPR  &  68.8  & 74.5  & 79.2  & 84.3 & 87.4 & 92.8 & 36.2  & 39.6  & 45.3  & 52.9  & 57.9 & 70.5  \\
				& MWNet~\cite{shu2019meta}        & NeurIPS   & 68.9  & 75.2  & 80.0  & 84.9 & 87.8 & 92.6 & 37.9  & 42.0  & 46.7  & 54.3  & 58.4 & 70.3 \\				
				\hline
				\multirow{10}{*}{{2020}}
				& CDB Loss~\cite{sinha2020class} & ACCV  & -      & -      & -      & -     & -     & -     & 37.4  & 42.5  & 46.7  & 54.2  & 58.7 & -   \\
				& RCBM-CE~\cite{jamal2020rethinking} & CVPR  & 70.6 & 76.4  & 80.5  & 86.4 & 88.8 & 92.7 & 39.3  & 43.3 & 48.5 & 55.6  & 59.5 & 71.8  \\
				& EQL~\cite{tan2020equalization} & CVPR & -      & -      & -      & -     & -     & -      & 43.3  & -      & -      & -      & -     & -     \\
				& BBN~\cite{zhou2020bbn}             & CVPR & -      & 79.8  & 82.1  & -     & 88.3 & -     & -      & 42.5  & 47.0  & -      & 59.1 & -    \\
				& De-c-TDE~\cite{tang2020long} & NeurIPS  & -      & 80.6   & 83.6   & -     & 88.5  & -     & -      & 44.1   & 50.3   & -      & 59.6  & -    \\
				& BALMS~\cite{ren2020balanced}  & NeurIPS & 81.5   & \textcolor{scorered}{\textbf{84.9}}   & -      & -     & \textcolor{scorered}{\textbf{91.3}}  & -     & 45.5   & \textcolor{scorered}{\textbf{50.8}}   & -      & -      & \textcolor{scoreblue}{\textbf{63.0}}  & -  \\
				& FSA~\cite{chu2020feature} & ECCV & 75.5  & \textcolor{scoreblue}{\textbf{82.0}}  & 84.4 & 89.2 & \textcolor{scoreblue}{\textbf{91.2}}  & -     & 41.4  & \textcolor{scoreblue}{\textbf{48.5}} & 52.1  & 59.7 & \textcolor{scorered}{\textbf{65.3}} & -  \\
				& Remix-DRW~\cite{chou2020remix}  & ECCV  & -      & 79.7  & -      & -     & \textcolor{scoregreen}{\textbf{89.0}} & -     & -      & \textcolor{scoregreen}{\textbf{46.7}}   & -      & -      & \textcolor{scoregreen}{\textbf{61.2}} & -      \\
				& LFME~\cite{xiang2020learning} & ECCV & -      & -      & -      & -     & -     & -     & 37.4  & 42.5  & 46.7  & 54.2  & 58.7 & -   \\
				& MBJ~\cite{liu2020memory} & ArXiv & -      & \textcolor{scoregreen}{\textbf{81.0}}   & 87.2   & -     & 88.8  & -     & -      & 45.8   & 57.5   & -      & 60.7  & -    \\
				
				\hline
				\multirow{14}{*}{{2021}}
				& RIDE(4 experts)~\cite{wang2020long} & ICLR  & -      & -      & -      & -     & -     & -     & -      & 49.1   & -      & -      & -     & -   \\
				& LADE~\cite{hong2021disentangling}& CVPR  & -      & -      & -      & -     & -     & -     & -      & 45.4   & 50.5   & -      & 61.7  & -   \\
				& MetaSAug-CE~\cite{li2021metasaug} & CVPR  & 76.8  & 80.5  & 84.0  & 87.6 & 89.4 & -     & 39.9  & 46.8  & 51.9  & 57.8 & 61.7 & -   \\
				& Hybrid-SC~\cite{wang2021contrastive}  & CVPR  & -      & 81.4  & 85.3  & -     & \textcolor{scorered}{\textbf{91.1}} & -     & -      & 46.7  & 51.8  & -      & 63.0 & -     \\
				& Hybrid-PSC~\cite{wang2021contrastive}  & CVPR  & -      & 78.8  & 83.8  & -     &  \textcolor{scoregreen}{\textbf{90.0}} & -     & -      & 44.9  & 48.9  & -      & 62.3 & -   \\
				& MiSLAS~\cite{zhong2021improving} & CVPR  & -      & \textcolor{scoregreen}{\textbf{82.1}}  & 85.7   & -     &  \textcolor{scoregreen}{\textbf{90.0}}  & -     & -      & 47.0   & 52.3   & -      & \textcolor{scoregreen}{\textbf{63.2}}  & -        \\
				& Bag of Tricks~\cite{zhang2021bag} & AAAI & -      & 80.0      & 83.5      & -     & -     & -     & -      & 47.8   & 51.6   & -      & -  & -   \\
				& LDA~\cite{peng2021lda}& ACM MM  & -      & -      & -      & -     & -     & -     & -      &  \textcolor{scoregreen}{\textbf{50.6}}   & 54.6   & -      & 61.9  & -   \\
				
				& TSC~\cite{Li2021tsc} & ArXiv & -      & 79.7      & 82.9      & -     & 88.7    & -     & -      & 43.8  & 47.4  & -      & 59.0 & -    \\
				& BKD~\cite{zhang2021bkd} & ArXiv & -      & 81.7      & 83.8      & -     & 89.2     & -     & -      & 45.0  & 49.6  & -      & 61.3 & -    \\
				& DRO-LT~\cite{dvir2021drl} & ArXiv  & -      & -      & -      & -     & -     & -     & -      & 47.3  & 57.5  & -      & \textcolor{scoreblue}{\textbf{63.4}} & -    \\
				& DiVE~\cite{he2021distilling} & ArXiv & -      & -      & -      & -     & -     & -     & -      & 45.3  & 51.1  & -      & 62.0 & -    \\
				& LADC~\cite{wang2021ladc} & ArXiv & 81.5      &  \textcolor{scoreblue}{\textbf{84.6}}      & 87.0     & -     &  \textcolor{scoreblue}{\textbf{90.8}}     & -     & 46.6      &  \textcolor{scoreblue}{\textbf{50.7}}  & 54.9  & -      &  \textcolor{scorered}{\textbf{64.6}} & -    \\
				& MARC~\cite{wang2021marc} & ArXiv & 81.1      &  \textcolor{scorered}{\textbf{85.3}}      & -      & -     & -     & -     & 47.4      &  \textcolor{scorered}{\textbf{50.8}}  & -  & -      & - & -    \\
				
				\hline
			\end{tabular}
		}
	\end{threeparttable}
	\vspace{-1mm}
\end{table*}

There are also some studies~\cite{yang2020rethinking} that use the strategy of generating pseudo labels to expand the data in semi-supervised manner~\cite{zou2018unsupervised}. Yang \etal~\cite{yang2020rethinking} investigate a self-training semi-supervised learning method on long-tailed data by acquiring a certain amount of unlabeled data and generating pseudo-label for it, which is combined with labeled data to learn the final model. Wei \etal~\cite{wei2021crest} perform data expansion of tail classes by predicting pseudo-labels for the tail classes of unlabelled data, thereby mitigating model over-fitting and alleviating data imbalance.

For object detection tasks, object-centered images may have richer data for tail classes. To make full use of these weakly labeled data and improve the tail classes performance, Ramanathan \etal~\cite{ramanathan2020dlwl} use a combination of weak supervision and full supervision. Weakly labeled images from YFCC100M \cite{thomee2016yfcc100m} are used to perform enhancement for rare classes. 

Also the scene-centric images have the potential to make it difficult for the detector to detect the tail objects, while the object-centric images make the detector give more attention to the target objects. Thus, Zhang \etal~\cite{zhang2021simple} use object-centric images to generate pseudo-scene-centric images to adjust the detector, thus eliminating the domain gap between the two image sources and also solving the problem of lacking bounding box labels for object-centric images. Finally, the scene centered images are combined to train and adjust the detector.

Semi-supervised learning generates pseudo-labels by learning from unlabeled data in order to expand the tail classes~\cite{yang2020rethinking}. This approach compensates for the problem of insufficient learning of tail representation, but requires additional training, and it is difficult to play a role when the unlabeled data is not easy to obtain.

\subsubsection{More Methods}

In addition, there are several approaches that employ \emph{Causal Inference}~\cite{tang2020long}, \emph{Adversarial Training}~\cite{wu2021adversarial}, \emph{Distributional Robust Optimization (DRO)}~\cite{dvir2021drl} \etc to solve the long-tailed problem.

Tang \etal~\cite{tang2020long} propose that the momentum in the SGD optimizer is a confounder of the sample features and the classification logits, which may lead to spurious correlation between them. Therefore, causal intervention is used for de-confounding training to cut off backdoor confounding path and retain mediation path.

Wu \etal~\cite{wu2021adversarial} find that long-tailed data have a negative impact on adversarial robustness and that the natural accuracy loss of the tail classes is further magnified in adversarial training. Meanwhile, they argue that suitable features as well as classifier embedding help to reduce the boundary error, and the combination of long-tailed recognition methods with the adversarial training framework helps to improve the natural accuracy. Therefore, the \emph{RoBal} framework is designed with scale-invariant classifiers and a two-stage rebalancing method, respectively, which are thus used to improve the adversarial robustness. 

\emph{DRO-LT}~\cite{dvir2021drl} aims to improve the representation learning layer, and in order not to compromise the original data representation, a new loss based on robustness theory is proposed, which encourages the model to learn high quality representations of both head and tail classes. 

On the whole, long-tailed learning continues to learn from other machine learning sub-fields, and more combinatorial methods will appear one after another.

\subsection{Summary}

We summarize the above long-tailed visual recognition approaches in this section. Highlights and limitations of these methods are shown in Tab.~\ref{table:highlights}. On the whole, the characteristics of these methods are very distinct, focusing on data processing, loss function, model architecture, training methods and so on. However, there is no method that can greatly solve the long-tailed problem. In practical application, it is often a combination of multiple methods, such as using semi-supervised learning to expand the tail data, using over-sampling to improve the sampling frequency of tail data, using class-level re-weighting to balance the gradient during training, and finally using decoupling strategy to enhance the representation ability of the model. Therefore, it can be predicted that the research on solving the long-tailed problem will still be in full bloom in the future.

\begin{table*}
	\centering
	\caption{
		{\textbf{Performance (\%) summarization of some representative methods on ImageNet-LT and Places-LT benchmarks}. For Places-LT dataset, we follow the experimental setup of most of the current work as a statistical benchmark, where ResNet-152 is uniformly selected as the backbone. (In the 2020 and 2021 studies, the three best scores of overall top-1 accuracy are marked in \textcolor{scorered}{red}, \textcolor{scoreblue}{blue}, and \textcolor{scoregreen}{green}, respectively.) PaCo~\cite{cui2021paco}$^\dagger$ models are trained with RandAugment~\cite{cubuk2020randaug} in 400 epochs.}
	}
	\vspace{-5pt}
	\label{table:im_places_lt_bench}
	\begin{threeparttable}
		\resizebox{0.98\textwidth}{!}{
			\setlength\tabcolsep{8pt}
			\renewcommand\arraystretch{1.05}
			\begin{tabular}{|c|c|c|c|c|c|c|c|c|c|c|c|}
				\hline\thickhline
				\rowcolor{purple0}
			         & & &  & \multicolumn{4}{c|}{{ImageNet-LT (top-1)}} & \multicolumn{4}{c|}{{Places-LT (top-1)}}  \\
			         \cline{5-12}
			         \rowcolor{purple0}
				 \multirow{-2}{*}{{Year}} & \multirow{-2}{*}{{Method}}  & \multirow{-2}{*}{{Pub.}}  &\multirow{-2}{*}{{Backbone}} &   \tabincell{c}{\textgreater{}100\\ Many-shot}   &   \tabincell{c}{$\leqslant100$ \& \textgreater{}20\\ Medium-shot}    &   \tabincell{c}{\textless{}20\\ Few-shot}    &   Overall    &   \tabincell{c}{\textgreater{}100\\ Many-shot}   &   \tabincell{c}{$\leqslant100$ \& \textgreater{}20\\ Medium-shot}    &   \tabincell{c}{\textless{}20\\ Few-shot}    &   Overall       \\
				\hline
				\hline
				
				\hline
				\multirow{3}{*}{{}}
				& Softmax Loss~\cite{he2016deep} & CVPR & ResNet-10   & 40.9  & 10.7  & 0.4  & 20.9  & 45.9  & 22.4  & 0.36  & 27.2   \\
				& Focal Loss~\cite{lin2017focal}   & ICCV   & ResNet-10   & 36.4  & 29.9  & 16.0  & 30.5  & 41.1  & 34.8  & 22.4  & 34.6      \\
				& Range Loss~\cite{zhang2017range}   & ICCV   & ResNet-10   & 35.8 & 30.3 & 17.6 & 30.7 & 41.1 & 35.4 & 23.2 & 35.1    \\
				
				\hline
				\multirow{1}{*}{{2019}}
				& OLTR~\cite{liu2019large} & CVPR & ResNet-10  & 43.2 & 35.1 & 18.5 & 35.6 & 44.7 & 37.0 & 25.3 & 35.9       \\
						
				\hline
				\multirow{12}{*}{{2020}}
				& CDB Loss~\cite{sinha2020class}   & ACCV   & ResNet-10  & -  & -  & -  & 38.4 & - & - & -  & -    \\
				& EQL~\cite{tan2020equalization}   & CVPR   & ResNet-10  & -  & -  & -  & 36.4 & - & - & -  & -    \\	
				& BALMS~\cite{ren2020balanced}   & NeurIPS   & ResNet-10   & 50.3 & 39.5 & 25.3 & 41.8 & 41.2 & 39.8 & 31.6 &  \textcolor{scorered}{\textbf{38.7}}             \\	
				& De-c-TDE~\cite{tang2020long}   & NeurIPS   & ResNeXt-50 & 62.7 & 48.8 & 31.6 &  \textcolor{scoreblue}{\textbf{51.8}} & - & - & - & -     \\		
				& FSA~\cite{chu2020feature}  & ECCV   & ResNet-10  & 47.3 & 31.6 & 14.7 & 35.2 & 42.8 & 37.5 & 22.7 &  \textcolor{scoregreen}{\textbf{36.4}}      \\	
				& LFME~\cite{xiang2020learning}   & ECCV   &  ResNet-10  & 47.1 & 35.0 & 17.5 & 37.2 & 38.4 & 39.1 & 21.7 & 35.2            \\	
				& Joint~\cite{kang2019decoupling}   & ArXiv   & ResNeXt-50 & -  & -  & -  & 44.4 & - & - & -  & -  \\
				& NCM~\cite{kang2019decoupling}   & ArXiv   & ResNeXt-50  & -  & -  & -  & 47.3 & - & - & -  & -    \\	
				& cRT~\cite{kang2019decoupling}   & ArXiv   & ResNeXt-50  & -  & -  & -  & 49.5 & - & - & -  & -   \\	
				& $\tau$-normalized~\cite{kang2019decoupling}   & ArXiv   & ResNeXt-50  & -  & -  & -  & 49.5 & - & - & -  & -   \\	
				& LWS~\cite{kang2019decoupling}   & ArXiv   & ResNeXt-50  & -  & -  & -  &  \textcolor{scoregreen}{\textbf{49.9}} & - & - & -  & -    \\	
				& MBJ~\cite{liu2020memory}   & ArXiv   & ResNeXt-50 & 61.6 & 48.4 & 39.0 &  \textcolor{scorered}{\textbf{52.1}} & 39.5 & 38.2 & 35.5 &  \textcolor{scoreblue}{\textbf{38.1}}          \\	
				
				\hline
				\multirow{19}{*}{{2021}}
				& RIDE(4 experts)~\cite{wang2020long} & ICLR & ResNet-50  & 66.2 & 52.3 & 36.5 & \textcolor{scoreblue}{\textbf{55.4}} & - & - & - & - \\
				& Logit adjustment~\cite{menon2020long} & ICLR & ResNet-50  & - & - & - & 51.1 & - & - & - & - \\
				& LADE~\cite{hong2021disentangling} & CVPR & ResNeXt-50  & 62.3 & 49.3 & 31.2 & 51.9 & 42.8 & 39.0 & 31.2 & 38.8                \\
				& Seesaw Loss~\cite{wang2021seesaw} & CVPR & ResNeXt-50  & 67.1 & 45.2 & 21.4 & 50.4 & - & - & - & -                \\        
				& MetaSAug-CE~\cite{li2021metasaug}  & CVPR & ResNet-50 & - & - & - & 47.3 & - & - & - & - \\
				& MiSLAS~\cite{zhong2021improving}  & CVPR & ResNet-50  & 61.7 & 51.3 & 35.8 & 52.7 & 39.6 & 43.3 & 36.1 & \textcolor{scoreblue}{\textbf{40.4}} \\
				& DisAlign~\cite{zhang2021distribution} & CVPR & ResNet-50  & 59.9 & 49.9 & 31.8 & 52.9 & 40.4 & 42.4 & 30.1 & 39.3 \\
				& Bag of Tricks~\cite{zhang2021bag} & AAAI & ResNet-10  & - & - & - & 43.1 & - & - & - & - \\
				& LDA~\cite{peng2021lda}  & ACM MM & ResNeXt-50  & 64.5 & 50.9 & 31.5 & 53.4 & 32.1 & 40.7 & 41.0 & 39.1 \\
				& PaCo~\cite{cui2021paco}$^\dagger$ & ICCV & ResNeXt-50  & -  & - & -  &  \textcolor{scorered}{\textbf{58.2}} & 36.1 & 47.9 & 35.3 & \textcolor{scorered}{\textbf{41.2}}  \\
				
				& TSC~\cite{Li2021tsc}  & ArXiv & ResNet-50 & 63.5  & 49.7 & 30.4  & 52.4 & -  & - & - & -  \\
				& GistNet~\cite{liu2021gist}  & ArXiv & ResNet-10  & 52.8 & 39.8 & 21.7 & 42.2 & 42.5 & 40.8 & 32.1 & 39.6 \\
				& BKD~\cite{zhang2021bkd} & ArXiv & ResNet-10  & 54.6 & 37.2 & 20.4 & 41.6 & 41.9 & 39.1 & 30.0 & 38.4 \\
				& DRO-LT~\cite{dvir2021drl} & ArXiv & ResNet-50  & 64.0 & 49.8 & 33.1 & \textcolor{scoregreen}{\textbf{53.5}} & - & - & - & - \\
				& ResLT~\cite{cui2021reslt}  & ArXiv & ResNeXt-50  & 63.0 & 50.5 & 35.5 & 52.9 & 39.8 & 43.6 & 31.4 & \textcolor{scoregreen}{\textbf{39.8}}  \\
				& DiVE~\cite{he2021distilling} & ArXiv & ResNeXt-50  & 64.0  & 50.4 & 31.4  & 53.1 & -  & - & - & -  \\
				& Breadcrumbs~\cite{liu2021bread} & ArXiv & ResNeXt-50  & 62.9  & 47.2 & 30.9  & 51.0 & 40.6  & 41.0 & 33.4 &  39.3  \\
				& ALA Loss~\cite{zhao2021ala}  & ArXiv & ResNeXt-50  & 64.1 & 49.1 & 34.0 & 52.8 & - & - & - & - \\
				& MARC~\cite{wang2021marc} & ArXiv & ResNeXt-50  & 60.4  & 50.3 & 36.6  & 52.3 & 39.9  & 39.8 & 32.6 & 38.4  \\
				
				\hline
			\end{tabular}
		}
	\end{threeparttable}
	\vspace{-1mm}
\end{table*}

\section{Performance Comparison}
\label{sec:performance_comp}
To provide readers with a straightforward statistic, we compare the performance of some mainstream long-tailed studies in this section. For classification task, we report some popular long-tailed studies via ImageNet-LT and Places-LT, CIFAR-10/100-LT, and iNaturalist 2017 \& 2018 benchmarks, respectively. For object detection and instance segmentation tasks, we report some popular long-tailed studies via LVIS benchmarks. It should be noted that the experimental settings of each study are not completely consistent. We try to eliminate these effects when comparing, but we still can't be absolutely fair. Therefore, we hope that readers can only take the comparison in this section as a reference, and the specific performance comparison still needs to be analyzed based on the implementation details of the original article.

\begin{table*}
	\centering
	\caption{
		{\textbf{Performance (\%) summarization of some representative methods on iNaturalist 2017 \& 2018 benchmarks.} All the methods adopt ResNet-50 as backbone. (In the 2020 and 2021 studies, the three best scores of overall top-1 accuracy on iNaturalist 2018 are marked in \textcolor{scorered}{red}, \textcolor{scoreblue}{blue}, and \textcolor{scoregreen}{green}, respectively.) PaCo~\cite{cui2021paco}$^\dagger$ is trained with RandAugment~\cite{cubuk2020randaug}.}
	}
	\vspace{-5pt}
	\label{table:inat_bench}
	\begin{threeparttable}
		\resizebox{0.98\textwidth}{!}{
			\setlength\tabcolsep{8pt}
			\renewcommand\arraystretch{1.05}
			\begin{tabular}{|c|c|c|c|c|c|c|c|c|}
				\hline\thickhline
				\rowcolor{purple0}
			         & & &  & iNat 2017 (top-1) & \multicolumn{4}{c|}{{iNat 2018 (top-1)}}  \\
			         \cline{5-9}
			         \rowcolor{purple0}
				 \multirow{-2}{*}{{Year}} & \multirow{-2}{*}{{Method}}  & \multirow{-2}{*}{{Pub.}}  &\multirow{-2}{*}{{Epoch}}  &   Overall    &   \tabincell{c}{\textgreater{}100\\ Many-shot}   &   \tabincell{c}{$\leqslant100$ \& \textgreater{}20\\ Medium-shot}    &   \tabincell{c}{\textless{}20\\ Few-shot}    &   Overall       \\
				\hline
				\hline
				
				\hline
				\multirow{1}{*}{{}}
				& Softmax Loss~\cite{he2016deep} & CVPR &         -   & 54.6  & -  & -  & -  & 57.1     \\
				
				\hline
				\multirow{2}{*}{{2019}}
				& CB Loss~\cite{cui2019class} & CVPR   & -             & 58.0  & -  & -  & -  & 61.1     \\
				& LDAM~\cite{cao2019learning} & NeurIPS   & 60+30    & -  & -  & -  & -  & 68.0     \\
						
				\hline
				\multirow{10}{*}{{2020}}
				& BBN~\cite{zhou2020bbn}     & CVPR    & 180        & 65.7  & -  & -  & -  &  \textcolor{scoregreen}{\textbf{69.6}}     \\
				& RCBM-CE~\cite{jamal2020rethinking} & CVPR   & -              & 59.3  & -  & -  & -  & 67.3     \\
				& FSA~\cite{chu2020feature} & ECCV     & 100         & 61.9  & -  & -  & -  & 65.9     \\
				& Remix-DRW~\cite{chou2020remix}  & ECCV   & 200   & -  & -  & -  & -  &  \textcolor{scorered}{\textbf{70.4}}     \\
				& Joint~\cite{kang2019decoupling} & ArXiv   & 90 / 200    & -    & 72.2 / 75.7  & 63.0 / 66.9  & 57.2 / 61.7  & 61.7 /  65.8     \\
				& NCM~\cite{cui2019class} & ArXiv   &  90 / 200        & -   & 55.5 / 61.0  & 57.9 / 63.5  & 59.3 / 63.3  & 58.2 / 63.1     \\
				& cRT~\cite{kang2019decoupling}  & ArXiv   & 90 / 200     & -   & 69.0 / 73.2  & 66.0 / 68.8  & 63.2 / 66.1  & 65.2 / 68.2     \\
				& $\tau$-normalized~\cite{kang2019decoupling} & ArXiv   & 90 / 200      & -   & 65.6 / 71.1  & 65.3 / 68.9  & 65.9 / 69.3  & 65.6 / 69.3     \\
				& LWS~\cite{cui2019class} & ArXiv   & 90 / 200    & -  & 65.0 / 71.0  & 66.3 / 69.8  & 65.5 / 68.8  & 65.9 / 69.5     \\
				& MBJ~\cite{liu2020memory} & ArXiv &  90 / 200        & -  & -  & -  & -  &  \textcolor{scoreblue}{\textbf{66.9 / 70.0}}     \\
				
				\hline
				\multirow{19}{*}{{2021}}
				& RIDE(4 experts)~\cite{wang2020long} & ICLR   & 100  & -  & 70.9  & 72.4  & 73.1  & \textcolor{scoreblue}{\textbf{72.6}}      \\
				& Logit adjustment~\cite{menon2020long} & ICLR &   90         & - & - & - & -  & 68.4 \\
				& Bag of Tricks~\cite{zhang2021bag} & AAAI &   90         & - & - & - & -  & 70.8 \\
				& LADE~\cite{hong2021disentangling} & CVPR   & 200 & -   & -  & -  & -  & 70.0     \\
				& MetaSAug-CE~\cite{li2021metasaug}  & CVPR   & -  & 63.2  & -  & -  & -  & 68.7     \\
				& Hybrid-SC~\cite{wang2021contrastive}  & CVPR   & 100   & -  & -  & -  & -  & 66.7     \\
				& Hybrid-PSC~\cite{wang2021contrastive} & CVPR   & 100 / 200  & -  & -  & -  & -  & 68.1 / 70.3     \\
				& MiSLAS~\cite{zhong2021improving} & CVPR   &  200       & -   & 73.2 & 72.4  & 70.4  & \textcolor{scoregreen}{\textbf{71.6}}      \\
				& DisAlign~\cite{zhang2021distribution} & CVPR   & 90 / 200 & -  & 61.6 / 68.0  & 70.8 / 71.3  & 69.9 / 69.4 & 69.5 / 70.2     \\
				& PaCo~\cite{cui2021paco}$^\dagger$ & ICCV & 400 & -   & -  & -  & -  & \textcolor{scorered}{\textbf{73.2}}     \\
				
				& TSC~\cite{Li2021tsc} & ArXiv & - & -   & 72.6  & 70.6  & 67.8  & 69.7      \\
				& GistNet~\cite{liu2021gist}  & ArXiv & 200 & -   & -  & -  & -  & 70.8     \\
				& BKD~\cite{zhang2021bkd} & ArXiv & 90 & -   & 67.1  & 66.1  & 67.6  & 66.8      \\
				& DRO-LT~\cite{dvir2021drl} & ArXiv  & - & -   & -  & -  & -  & 69.7     \\
				& ResLT~\cite{cui2021reslt}  & ArXiv & 200 & -   & -  & -  & -  & 70.2      \\
				& DiVE~\cite{he2021distilling}  & ArXiv   & 90       & -   & 70.6  & 70.0  & 67.5  & 69.1     \\
				& Breadcrumbs~\cite{liu2021bread} & ArXiv & 200 & -   & -  & -  & -  & 70.3     \\
				& ALA Loss~\cite{zhao2021ala}& ArXiv & 200   & -  & 71.3   & 70.8  & 70.4  & 70.7     \\
				& MARC~\cite{wang2021marc} & ArXiv & 200 & -   & -  & -  & -  & 70.4     \\
				
				\hline
			\end{tabular}
		}
	\end{threeparttable}
	\vspace{-1mm}
\end{table*}

\subsection{CIFAR-10/100-LT Performance Benchmarking}
The performance of some representative methods on CIFAR-10/100-LT benchmark is shown in Tab.~\ref{table:cifar_lt_bench}. In the 2020 and 2021 studies, we select two cases with imbalance factors of 100 and 10 on CIFAR-10/100-LT, respectively. The three best scores of each year are marked in red, blue, and green. In the 2020 studies, \emph{BALMS}~\cite{ren2020balanced} achieve the highest performance in three metrics. In addition, \emph{FSA}~\cite{chu2020feature} and \emph{Remix-DRW}~\cite{chou2020remix} are also competitive methods in that year. In the next year, \emph{MARC}~\cite{wang2021marc}, \emph{LADC}~\cite{wang2021ladc} and \emph{MiSLAS}~\cite{zhong2021improving} rank in the top three of the comprehensive performance.

Through the results, we can clearly observe that the performance of the method in 2021 is slightly improved compared with that in 2020. The 2021 best method \emph{MARC} is only 0.4 points higher (85.3\% vs 84.9\%) than the 2020 best method \emph{BALMS} in 2020 at $\beta$ = 100 of CIFAR-10-LT, and \emph{MARC} has not even improved in the other three metrics. This shows that CIFAR-LT datasets  have become saturated, and future research should focus on more difficult benchmarks.

\vspace{6pt}
\subsection{ImageNet-LT \& Places-LT Performance Benchmarking}
Tab.~\ref{table:im_places_lt_bench} shows the performance of some representative methods on ImageNet-LT and Places-LT benchmarks. The three best scores of overall top-1 accuracy on 2020 and 2021 are marked in red, blue, and green, respectively.

For ImageNet-LT, the 2021 best method \emph{PaCo}~\cite{cui2021paco} achieves 58.2\% overall top-1 accuracy, which is 6.1 points higher than the 2020 best method \emph{MBJ}~\cite{liu2020memory}. Although \emph{PaCo} adopts RandAugment and longer training epochs, this improvement is still very significant. In addition, Tab.~\ref{table:im_places_lt_bench} records 19 studies on ImageNet-LT in 2021, which has almost doubled compared with 2020. More researchers have joined the community and greatly promoted the development of long-tailed recognition. Of course, compared with the best result of ResNeXt-50 on balanced ImageNet-1K (80.5\%~\cite{ross2021strikes}), there are still many problems to be solved in the research of long-tailed recognition.

For Places-LT, \emph{PaCo} is still the best method in 2021, with an overall top-1 accuracy of 41.2\%, which is 2.5 points higher than the 2020 best method \emph{BALMS}~\cite{ren2020balanced}. Although the methods in 2021 are better than those in 2020 on the whole, the improvement is still relatively small compared with ImageNet-LT. We consider that most of the current long-tailed recognition methods are mainly designed for object-centric data, while the scene-centric long-tailed problem needs specific solutions.

\vspace{6pt}
\subsection{iNaturalist 2017 \& 2018 Performance Benchmarking}
iNaturalist is a large species dataset, and the relevant research results of the 2017 and 2018 versions are given in Tab.~\ref{table:inat_bench}. Since there is less research on iNaturalist 2017, we mainly analyze the studies on iNaturalist 2018. Same as Tab.~\ref{table:im_places_lt_bench}, three best scores of overall top-1 accuracy on 2020 and 2021 are marked in red, blue, and green, respectively.

Among 2020 studies, \emph{Remix-DRW}~\cite{chou2020remix}, \emph{MBJ}~\cite{liu2020memory} and \emph{BBN}~\cite{zhou2020bbn} rank in the best three on the overall top-1 accuracy, achieve 70.4\%, 70.0\% and 69.6\% respectively. \emph{PaCo}~\cite{cui2021paco} still stand out in 19 studies in 2021, with 73.2\% overall top-1 accuracy, an increase of 2.8 points compared with the best method in 2020. On the whole, iNaturalist 2018 is still not saturated, and its importance in the long-tailed research community is basically equal to ImageNet-LT. This domain-specific long-tailed benchmark enriches the diversity of research objectives and has great potential in industrial vision, retail, medical, \etc

\begin{table*}
	\centering
	\caption{
		{\textbf{Performance (\%) summarization of some representative methods on LVIS v0.5 benchmark.} All the methods adopt Mask R-CNN with ResNet-50-FPN. In the 'Epoch' column, values making additive operations represents the two-stage training strategy. (In the 2020 and 2021 studies, the three best scores of AP$^\text{mask}$, AP$^\text{mask}_{r}$ and AP$^\text{bbox}$ are marked in \textcolor{scorered}{red}, \textcolor{scoreblue}{blue}, and \textcolor{scoregreen}{green}, respectively.)}
	}
	\vspace{-5pt}
	\label{table:lvis_v05_bench}
	\begin{threeparttable}
		\resizebox{0.85\textwidth}{!}{
			\setlength\tabcolsep{8pt}
			\renewcommand\arraystretch{1.05}
			\begin{tabular}{|c|c|c|c|c|c|c|c|c|c|c|c|c|c|c|}
				\hline\thickhline
				\rowcolor{purple0}
			         & & &  & \multicolumn{7}{c|}{{LVIS v0.5 (mAP)}}  \\
			         \cline{5-11}
			         \rowcolor{purple0}
			         & & &  & \multicolumn{6}{c|}{{AP$^\text{mask}$}} &   \\  
			         \cline{5-10}
			         \rowcolor{purple0}
				 \multirow{-3}{*}{{Year}} & \multirow{-3}{*}{{Method}}  & \multirow{-3}{*}{{Pub.}}  & \multirow{-3}{*}{{Epoch}}  &   AP  &  AP$_{50}$  &  AP$_{75}$  & AP$_{r}$  &  AP$_{c}$  &  AP$_{f}$ &  \multirow{-2}{*}{{AP$^\text{bbox}$}}   \\
				\hline
				\hline
				
				\hline
				\multirow{4}{*}{{}}
				& Softmax Loss~\cite{he2016deep} & CVPR & 25  & 20.2  & 32.6  & 21.3 & 4.5  & 20.8 & 25.6  &  20.7  \\
				& Sigmoid Loss  & - & 25  & 20.1  & 32.7  & 21.2  & 7.2  & 19.9 & 25.4  &  20.5  \\
				& CAS~\cite{shen2016relay} & ECCV & 25  & 18.5  & 31.1  & 18.9  & 7.3  & 19.3  & 21.9 & 18.4  \\
				& Focal Loss~\cite{lin2017focal} & ICCV & 25  & 21.0  & 34.2  & 22.1  & 9.3  & 21.0  & 25.8 & 21.9  \\
				
				\hline
				\multirow{3}{*}{{2019}}
				& RFS~\cite{gupta2019lvis} & CVPR & 25  & 24.4  & -  & -  & 14.5  & 24.3 & 28.4  &  -  \\
				& CB Loss~\cite{cui2019class} & CVPR & 25  & 20.9  & 33.8  & 22.2  & 8.2  & 21.2 & 25.7  &  21.0  \\
				& LDAM~\cite{cao2019learning} & NeurIPS   &  25  & 24.1  & -  & -  & 14.6  & 25.3 & 26.3  &  24.5  \\
				
				\hline
				\multirow{8}{*}{{2020}}
				& EQL~\cite{tan2020equalization} & CVPR & 25   & 22.8  & 36.0  & 24.4  & 11.3  & 24.7  & 25.1 & 23.3   \\
				& LST~\cite{hu2020learning} & CVPR & 10 + 10   & 23.0  & 36.7  & 24.8  & -  & -  & - & 22.6   \\
				& BAGS\cite{li2020overcoming} & CVPR & 12  + 12   & 26.2  & -  & -  & 17.9  & 26.9  & 28.7 & \textcolor{scoregreen}{\textbf{25.7}}   \\
				& Forest R-CNN\cite{wu2020forest} & ACM MM & 25   & 25.6  & 40.3  & 27.1  & \textcolor{scoregreen}{\textbf{18.3}}  & 26.4  & 27.6 & 25.9   \\
				& TFA-cos\cite{li2020overcoming} & ICML & -   & -  & -  & -  & -  & -  & - & 22.7   \\
				& BALMS\cite{ren2020balanced} & NeurIPS & 25   & \textcolor{scoreblue}{\textbf{27.0}}  & -  & -  & \textcolor{scorered}{\textbf{19.6}}  & 28.9  & 27.5 & \textcolor{scorered}{\textbf{27.6}}   \\
				& SimCal~\cite{wang2020devil} & ECCV & -   & 23.4  & -  & -  & 16.4  & 22.5  & 27.2 & -   \\
				& LWS~\cite{cui2019class} & ArXiv  &  25  & 23.8  & -  & -  & 14.4  & 24.4 & 26.8  &  24.5 \\
				
				\hline
				\multirow{5}{*}{{2021}}
				& DisAlign~\cite{zhang2021distribution} & CVPR  & 25 + 2.5  & 26.3  & -  & -  & 14.9  & 27.6 & 29.2  &  25.6  \\
				& EQL v2~\cite{tan2021equalization} & CVPR  & 24  & \textcolor{scorered}{\textbf{27.1}}  & -  & -  & \textcolor{scoreblue}{\textbf{18.6}}  & 27.6 & 29.9  &  \textcolor{scoreblue}{\textbf{27.0}}  \\
				& ACSL\cite{wang2021adaptive} & CVPR & 12  + 12   & \textcolor{scoregreen}{\textbf{26.4}}  & 42.3  & 28.6  & \textcolor{scoreblue}{\textbf{18.6}}  & 26.4  & 29.3 & -   \\
				& Drop Loss\cite{hsieh2021droploss} & CVPR & 25  & 25.5  & 38.7  & 27.2 & 13.2  & 27.9 & 27.3  &  25.1  \\
				& Simp-Effe\cite{zhang2021simple} & ArXiv & 25   & -  & -  & -  & -  & -  & - & 24.5   \\

				\hline
			\end{tabular}
		}
	\end{threeparttable}
	\vspace{-1mm}
\end{table*}

\begin{table*}
	\centering
	\caption{
		{\textbf{Performance (\%) summarization of some representative methods on LVIS v1.0 benchmark.} All the methods adopt Mask R-CNN with ResNet-50-FPN. $^\dagger$ denotes that the result is reproduced by us. s(In the 2020 and 2021 studies, the three best scores of AP$^\text{mask}$, AP$^\text{mask}_{r}$ and AP$^\text{bbox}$ are marked in \textcolor{scorered}{red}, \textcolor{scoreblue}{blue}, and \textcolor{scoregreen}{green}, respectively.)}
	}
	\vspace{-5pt}
	\label{table:lvis_v10_bench}
	\begin{threeparttable}
		\resizebox{0.85\textwidth}{!}{
			\setlength\tabcolsep{8pt}
			\renewcommand\arraystretch{1.05}
			\begin{tabular}{|c|c|c|c|c|c|c|c|c|c|c|c|c|c|c|}
				\hline\thickhline
				\rowcolor{purple0}
			         & & &  & \multicolumn{7}{c|}{{LVIS v1.0 (mAP)}}  \\
			         \cline{5-11}
			         \rowcolor{purple0}
			         & & &  & \multicolumn{6}{c|}{{AP$^\text{mask}$}} &   \\  
			         \cline{5-10}
			         \rowcolor{purple0}
				 \multirow{-3}{*}{{Year}} & \multirow{-3}{*}{{Method}}  & \multirow{-3}{*}{{Pub.}}  & \multirow{-3}{*}{{Epoch}}  &   AP  &  AP$_{50}$  &  AP$_{75}$  & AP$_{r}$  &  AP$_{c}$  &  AP$_{f}$ &  \multirow{-2}{*}{{AP$^\text{bbox}$}}   \\
				\hline
				\hline
				
				\hline
				\multirow{3}{*}{{}}
				& Softmax Loss~\cite{he2016deep}$^\dagger$ & CVPR & 25  & 18.2  & 28.6  & 19.1 & 1.2 & 15.3 & 28.9  &  18.6  \\
				& Focal Loss~\cite{lin2017focal} & ICCV & 24  & - & -  & -  & -  & - & -  &  18.5  \\
				& RFS~\cite{gupta2019lvis}$^\dagger$ & CVPR & 25  & 22.2 & 34.9  & 23.5  & 11.1  & 20.8 & 28.6  &  22.9  \\
				
				\hline
				\multirow{2}{*}{{2020}}
				& EQL~\cite{tan2020equalization, tan2021equalization} & CVPR & 24   & 21.6  & -  & -  & 3.8  & 21.7  & 29.2 & 22.5   \\
				& BAGS\cite{li2020overcoming}$^\dagger$ & CVPR & 25  & \textcolor{scoregreen}{\textbf{23.7}}  & 37.7  & 25.1  & \textcolor{scoregreen}{\textbf{15.4}}  & 22.8 & 28.3  & 23.4  \\
				
				\hline
				\multirow{6}{*}{{2021}}
				& Drop Loss\cite{hsieh2021droploss} & AAAI & 25  & 22.3  & 34.5  & 23.6  & 12.4  & 22.3 & 26.5  & 22.9  \\
				& EQL v2~\cite{tan2021equalization} & CVPR & 24   & \textcolor{scoreblue}{\textbf{25.5}}  & -  & -  & \textcolor{scoreblue}{\textbf{17.7}}  & 24.3  & 30.2 & 26.1   \\
				& Seesaw Loss\cite{wang2021seesaw} & CVPR & 24  & \textcolor{scorered}{\textbf{25.7}}  & -  & -  & \textcolor{scorered}{\textbf{19.1}}  & 25.0 & 29.4  & \textcolor{scoreblue}{\textbf{26.8}}  \\
				& LDA~\cite{peng2021lda} & ACM MM & 24  & \textcolor{scorered}{\textbf{25.7}}   & -  & -  & -  & - & -  & \textcolor{scoregreen}{\textbf{26.6}}  \\
				& FASA\cite{zang2021fasa} & ArXiv & 24  & 22.6  & -  & -  & 10.2  & 21.6 & 29.2  & 22.6  \\
				& Fed Loss\cite{zhou2021probabilistic}$^\dagger$ & ArXiv & 25  & 21.4  & 33.4  & 22.6  & 4.5  & 20.5 & 29.7  & 22.2  \\
			
				\hline
				\multirow{1}{*}{{2022}}
				& EOD\cite{Li2022eod} & ArXiv & 24  & -  & -  & -  & -  & - & -  & \textcolor{scorered}{\textbf{27.5}}   \\
					
				\hline
			\end{tabular}
		}
	\end{threeparttable}
	\vspace{-1mm}
\end{table*}

\vspace{6pt}
\subsection{LVIS v0.5 \& v1.0 Performance Benchmarking}

Tab.~\ref{table:lvis_v05_bench} and Tab.~\ref{table:lvis_v10_bench} summarize the performance of some representative work on the LVIS v0.5 \& v1.0 benchmark. Due to the small number of studies, we will not discuss the studies in 2020 and 2021 separately. And as LVIS v1.0 gradually becomes the mainstream long-tailed instance segmentation benchmark, some new research does not conduct experiments on LVIS v0.5, so here we only analyze and discuss the former benchmark. The three best scores of AP$^\text{mask}$, AP$^\text{mask}_{r}$ and AP$^\text{bbox}$ are marked in red, blue, and green.

Overall, the performance of \emph{EOD}~\cite{Li2022eod}, \emph{Seesaw Loss}~\cite{wang2021seesaw}, \emph{LDA}~\cite{peng2021lda} and \emph{EQL v2}~\cite{tan2021equalization} is in the leading position in LVIS v1.0 benchmark, and the gap between them is not large. Most of these studies belong to class-level re-weighting method, which can be seen as the mainstream solution to the long-tailed object detection and instance segmentation problems. These methods also use \emph{RFS}~\cite{gupta2019lvis} to increase the sampling frequency of tail classes, and have achieved good results. In addition, the regularly held challenge competitions~\cite{lvis2019} also significantly promoted community development. \emph{EQL}~\cite{tan2020equalization, tan2021equalization}, \emph{Seesaw Loss} and \emph{Fed Loss}~\cite{zhou2021probabilistic} have all won good rankings. However, most of these studies are the extension of image-level long-tailed recognition methods, and there is still less research on the instance-level long-tailed problem. Especially for the long-tailed instance segmentation problem, the use of mask information is very small, which is a problem that researchers need to pay attention to.

\section{Analysis of Long-tailed Phenomenon}
\label{sec:phenomenon}
Although the long-tailed phenomenon is prevalent in visual recognition, the scope of current research is very limited. To solve long-tailed problems, recent research pays much attention to some long-tailed benchmark datasets which we collated in Sec.~\ref{sec:longtail datasets}, but in other tasks, the solution of the long-tailed phenomenon is not sufficient. To investigate this, we analyze some widely-used datasets in visual recognition, and quantitatively evaluate the impact of the long-tailed phenomenon.

\begin{table*}[htbp]
	\centering
	\caption{\textbf{Analysis of long-tailed phenomenon of 20 mainstream datasets}. \textcolor{purple}{Light purple} indicates which has a general long-tailed distribution (0.6 $\leq$ $\delta$ $\textless$ 0.8), and \textcolor{purple1}{dark purple} indicates which has a severe long-tailed distribution (0.8 $\leq$ $\delta$). $\dagger$ denotes few-shots datasets.}
	\begin{threeparttable}
		\resizebox{0.99\textwidth}{!}{
			\setlength\tabcolsep{5pt}
			\renewcommand\arraystretch{1.05}
			\begin{tabular}{|c|c|c|c|c|c|c|c|c|}
				\hline\thickhline
				\rowcolor{purple0}
				Dataset                                                                                      & Venue                        &  Fields                                                 & \tabincell{c}{Anno. Types} & Training Samples & Classes & Max Size & Min Size       & Gini Coef. $\delta$ \\ 
				\hline \hline
				ImageNet-1K~\cite{deng2009imagenet}                                   &  \tabincell{c}{CVPR 2009 \\ IJCV 2015}    & Object-centric     & Classification                             & 1,281,167               & 1,000                        & 1,300     & 732           & 0.013     \\
				Sports1M~\cite{karpathy2014sports1m}                                   & CVPR 2014               & Human-centric                                    & Classification    & 958,827  & 487  & 2,385  &  694        & 0.09    \\ 
				COCO~\cite{lin2014microsoft}                                             & ECCV 2014               & Object-centric            & \tabincell{c}{Bounding-box \\ Instance-mask} & 118,287 & 80  & 262,465     & 198        & 0.564        \\ 
				Market1501~\cite{zheng2015market1501}                               & ICCV 2015                & Human-centric         & Person-identity    & 12,937   & 752  & 72   &  1   & 0.329          \\
				MS1M~\cite{guo2016ms}                                       & ECCV 2016                & Face-centric              & Face-identity  & 5,822,653 &    85,742   & 602    & 2     & 0.314                   \\ 
				DukeMTMC~\cite{ristani2016duke}                                           & ECCV 2016              & Human-centric        & Person-identity    & 16,522   & 702  & 426  & 6          & 0.268       \\ 
				\rowcolor{purple1}
				YouTube8M~\cite{sami2016youtube8m}                                   & ArXiv 2016               & Human-centric          & Classification    & 5,786,881   & 3,862  & 788,288  & 123      & 0.839        \\ 
				Place365~\cite{zhou2017places}                                              & PAMI 2017                 & Scene-centric            & Classification    & 1,803,460               & 365                           & 5,000     &  3,068        & 0.011           \\ 
				\rowcolor{purple1}
				ADE20K~\cite{zhou2017ade20k}                                              & CVPR 2017               & Scene-centric        & Segmentation    & 20,210   & 150  & 3014.9  &  3.84          & 0.801         \\ 
				Sth-Sth v2~\cite{goyal2017sthsth}                                             & ICCV 2017                & Human-centric         & Classification    & 168,913   & 174  & 3,284  &  91       & 0.352         \\ 
				\rowcolor{purple} 
				COCO-stuff~\cite{caesar2018cocostuff}                                    & CVPR 2018               & Scene-centric       & Segmentation    & 118,287  & 171  & 10,021.1  &  3.81         & 0.653       \\ 
				\rowcolor{purple}
				MHP v2~\cite{zhao2018mhp}                                                     &  ACM MM 2018                  & Human-centric      & Human-parsing    & 15,403   & 59  & 840.7  &  0.823     & 0.747          \\
				\rowcolor{purple1}
				OID v4~\cite{kuznetsova2020open}                                           &  \tabincell{c}{ArXiv 2018 \\ IJCV 2020}              & Object-centric      & Bounding-box    & 1,743,042   & 500  & 1,395,645  &  4       & 0.902         \\ 
				\rowcolor{purple1}
				Object365~\cite{shao2019objects365}                                      & ICCV 2019                & Object-centric           &  Bounding-box  & 608,606 & 365                   & 2,120,895 & 28      & 0.845           \\ 
				\rowcolor{purple}
				GLD v2~\cite{weyand2020gldv2}                                               & CVPR 2020               & Scene-centric            & Classification  & 4,132,914               & 203,094                           & 10,247     &  1   & 0.655                \\ 
				FSOD$\dagger$~\cite{fan2020fsod}                                          & CVPR 2020               &  Object-centric           &  Bounding-box  & 52,350 & 800                   & 2,114 & 26       & 0.361            \\ 
				FSS-1000$\dagger$~\cite{li2020fss1000}                                  & CVPR 2020               & Object-centric       & Segmentation    & 20,006   & 1,000  & 21  &  20       & 0.00029      \\  
				\rowcolor{purple}
				Glint360K~\cite{an2020glink360}                                               & ArXiv 2020                & Face-centric              & Face-identity   & 17,091,657 & 360,232                  & 1,868    & 3       & 0.647           \\ 
				\rowcolor{purple}
				VSPW~\cite{miao2021vspw}                                                      & CVPR 2021              & Scene-centric       & Segmentation    & 197,253   & 124  & 17,191.6  &  13.5      & 0.742         \\ 
				\rowcolor{purple}
				LaST~\cite{shu2021last}                                                            & ArXiv 2021                & Human-centric        & Person-identity    & 71,248   & 5,000  & 140  &  5           & 0.427         \\ 
				\hline
			\end{tabular}
		}
	\end{threeparttable}
	\vspace{-1mm}
	\label{table:alldata}
\end{table*}

\subsection{Long-tailed Phenomenon in Widely-used Datasets}
\label{sec:lt_pheno}
Long-tailed distribution is a common phenomenon. Datasets without artificial balance will basically follow this distribution. In order to study the long-tailedness in visual recognition, we compiled and analyzed other 20 widely-used large-scale datasets covering the fields of image classification~\cite{deng2009imagenet, zhou2017places, weyand2020gldv2}, object detection~\cite{lin2014microsoft, kuznetsova2020open, shao2019objects365, fan2020fsod}, semantic segmentation~\cite{zhou2017ade20k, caesar2018cocostuff, li2020fss1000, miao2021vspw}, person re-identification~\cite{zheng2015market1501, ristani2016duke, shu2021last}, face recognition~\cite{guo2016ms, an2020glink360}, human parsing~\cite{zhao2018mhp}, video / action recognition~\cite{karpathy2014sports1m, sami2016youtube8m, goyal2017sthsth}, \etc as shown in Tab.~\ref{table:alldata}. Based on the Gini coefficient proposed in Sec.~\ref{sec:longtail datasets}, we measure the long-tailedness in these datasets and mark each of them with general long-tailed distribution in light purple and those with severe long-tailed distribution in dark purple. From the perspective of both release time and annotation type, we can observe the trend of the long-tailed phenomenon in visual recognition approximately.

In terms of release time, a number of datasets showing long-tailed phenomenons have emerged since 2016, and the proportion of long-tailed datasets has gradually increased with each year. People are almost no longer controlling the balance of datasets artificially as CIFAR, ImageNet-1K, Places365 and COCO did.

In terms of annotation type, although we have improved the long-tailed criteria of the classification datasets, the classification task has a larger proportion of balanced datasets than the object detection and segmentation tasks in our statistics. It can be seen that for the classification task, the balance of datasets is easier to control compared to object detection and segmentation task. Such as ImageNet-1K~\cite{deng2009imagenet}, Sport1M~\cite{karpathy2014sports1m}, Sth-Sth v2~\cite{goyal2017sthsth}, Market1501~\cite{zheng2015market1501}, DukeMTMC~\cite{ristani2016duke}, MS1M~\cite{guo2016ms}, and Places365~\cite{zhou2017places}, many datasets in classification task demonstrate balance for which Gini coefficients are less than 0.4 and are considered as balanced datasets. Nevertheless, there are still many long-tailed datasets for classification tasks without adding artificial control over the balance, such as scene classification dataset GLD v2~\cite{weyand2020gldv2} ($\delta$=0.655), face recognition dataset Glint360K~\cite{an2020glink360} ($ \delta $ = 0.647) and person re-identification dataset LaST~\cite{shu2021last}($\delta$=0.427) which are considered as general long-tailed datasets. The large video understanding dataset YouTube8M~\cite{sami2016youtube8m} ($\delta$=0.839) is a severe long-tailed dataset. In the field of object detection and segmentation, there are fewer balanced datasets. Except COCO~\cite{lin2014microsoft}, only few-shot learning datasets FSOD~\cite{fan2020fsod} and FSS-1000~\cite{li2020fss1000} show to be balanced, which also reflects the difference between few-shot learning and long-tailed problems. Moreover, from the perspective of the Gini coefficient, the Gini coefficient for the object detection and segmentation task is generally higher than that for the classification tasks and is generally higher than 0.7. For example, for object detection task, OID v4~\cite{kuznetsova2020open} ($\delta$=0.902) and Object365~\cite{shao2019objects365} ($\delta$=0.845) both have Gini coefficients greater than 0.8, which belong to severe long-tailed datasets. And for segmentation task, human body parsing dataset MHP v2~\cite{zhao2018mhp} ($\delta$=0.747) and video semantic segmentation dataset VSPW~\cite{miao2021vspw} ($\delta$=0.742) are general long-tailed datasets. The scene parsing dataset ADE20K~\cite{zhou2017ade20k} ($\delta$=0.801) is a severe long-tailed dataset. This indicates that the long-tailed phenomenon is much more serious for object detection task and segmentation task, which requires more attention.

We find that the long-tailed phenomenon of the dataset is prevalent in the statistical process. Tab.~\ref{table:alldata} shows that the Gini coefficient of the dataset is gradually increasing in recent years, accompanied by a more severe long-tailed phenomenon in the dataset. We attribute this phenomenon to the fact that as people's research is more and more invested in large-scale datasets, it is increasingly difficult for researchers to control the balance of datasets, so the long-tailed phenomenon will inevitably become more and more serious, which leads to more and more long-tailed datasets in recent years.

\subsection{Analysis of Performance}


We select two severe long-tailed datasets in Tab.~\ref{table:alldata} (Object365~\cite{shao2019objects365} and ADE20K~\cite{zhou2017ade20k}), and evaluate their performance to investigate whether the long-tailed problem is practically shown in visual recognition. In order to evaluate the impact of the long-tailed phenomenon, we split classes in descending order into three groups: the first 20\% as head classes, the middle 60\% as body classes, and the last 20\% as tail classes. We collect some mainstream solutions without long-tailed methods on these three datasets and analyze their performance on head, body and tail classes, respectively.

\begin{table}[htbp]
	\centering
	\caption{\textbf{Long-tailed performance analysis of object detection on Object365~\cite{shao2019objects365}}. All the methods adopt ResNet-50-FPN as backbone and are trained by us on \emph{Detectron2}~\cite{detectron2019}.}
	\begin{threeparttable}
		\resizebox{0.48\textwidth}{!}{
			\setlength\tabcolsep{5pt}
			\renewcommand\arraystretch{1.05}
			\begin{tabular}{|c|c|ccc|}
				\hline\thickhline
				\rowcolor{purple0}
				Method                                                                &  AP                  &  AP$_{tail}$           & AP$_{body}$   & AP$_{head}$  \\ 
				\hline \hline
				Faster R-CNN~\cite{ren2015faster}                    & 19.8                     & 3.7                           & 21.5                      & 31.0                     \\
				RetinaNet~\cite{lin2017focal}                              & 18.5                     & 3.5                           & 19.9                      & 29.4                     \\
				FCOS~\cite{tian2019fcos}                                  & 20.6                     & 4.8                           & 22.2                      & 31.7                     \\
				\hline
			\end{tabular}
		}
	\end{threeparttable}
	\vspace{-1mm}
	\label{table:long_tail_det}
\end{table}

For object detection task, we investigate the generic object detection dataset Object365 and evaluate the performance of Faster R-CNN~\cite{ren2015faster}, RetinaNet~\cite{lin2017focal} and FCOS~\cite{tian2019fcos} on the head, body and tail classes. We adopt ResNet-50-FPN as backbone based on \emph{Detectron2}~\cite{detectron2019}, and take AP$_{box}$ as the evaluation metric, the performance results are shown in Tab.~\ref{table:long_tail_det}. For Faster R-CNN, RetinaNet and FCOS, the head classes accuracy is 1.44$\times$, 1.48$\times$ and 1.43$\times$ higher than the body classes, and 8.38$\times$, 8.40$\times$ and 6.60$\times$ higher than the tail classes.

\begin{table}[htbp]
	\centering
	\caption{\textbf{Long-tailed performance analysis of semantic segmentation on ADE20K~\cite{zhou2017ade20k}}. All the methods adopt ResNet-50 as backbone, the SemSegFPN and PSPNet models are taken from \emph{mmsegmentation}~\cite{mmsegmentation2020}, the MaskFormer model is taken from the officially published.}
	\begin{threeparttable}
		\resizebox{0.48\textwidth}{!}{
			\setlength\tabcolsep{5pt}
			\renewcommand\arraystretch{1.05}
			\begin{tabular}{|c|c|ccc|}
				\hline\thickhline
				\rowcolor{purple0}
				Method                                                                &  mIoU                  &  mIoU$_{tail}$           & mIoU$_{body}$   & mIoU$_{head}$  \\ 
				\hline \hline
				SemSegFPN~\cite{alexander2019semsegfpn}   & 37.48                    & 24.03                       & 35.33                   & 57.40                  \\
				PSPNet~\cite{zhao2017pspnet}                          & 42.47                    & 29.16                       & 40.96                   & 60.32                  \\
				MaskFormer~\cite{cheng2021maskformer}        & 44.50                    & 33.19                       & 41.96                   & 61.96                  \\
				\hline
			\end{tabular}
		}
	\end{threeparttable}
	\vspace{-1mm}
	\label{table:long_tail_semseg}
\end{table}

For semantic segmentation task, we investigate the performance of mainstream models on the large scene parsing dataset ADE20K~\cite{zhou2017ade20k}. We take mIoU as the evaluation metric, with the backbone of ResNet-50, the performance of SemSegFPN~\cite{alexander2019semsegfpn}, PSPNet~\cite{zhao2017pspnet}, and MaskFormer~\cite{cheng2021maskformer} is shown in Tab.~\ref{table:long_tail_semseg}. For SemSegFPN, PSPNet, and MaskFormer, their head classes accuracy exceeds the body classes by 1.62$\times$, 1.47$\times$, and 1.47$\times$, and exceeds the tail classes by 2.39$\times$, 2.06$\times$, and 1.87$\times$.

From the performance of the above model on the head, body and tail classes, it can be found that the accuracy of model is strongly related to the instance number of classes, and their performance becomes worse as the instance number decreases, which indicates that the long-tailed problem of these datasets needs to be solved. There is no doubt that the long-tailed phenomenon is prevalent, and as people increasingly analyze large-scale datasets, people should realize the importance of solving it. Although some studies are aware of the datasets' long-tailed distribution, this problem has not been widely studied. For example, for the pixel-level semantic segmentation task, only few studies~\cite{zhang2021distribution} have addressed the long-tailed phenomenon. Even for many other fields, none of their mainstream approaches has a targeted design for the long-tailed problem, so we believe that more effort needs to be devoted to the analysis of the long-tailed phenomenon.

\section{Future Directions}
\label{sec:guidelines}

As a contemporary survey for long-tailed visual recognition using deep learning, this paper has discussed the problems caused by the long-tailed distribution, summarized existing popular long-tailed datasets, provided some structural taxonomy for various methods as well as analyzed their advantages and limitations, we also find that the long-tailed phenomenon is widespread, and pointed out some valuable research areas of the long-tailed problem. Despite great progress, there are still many unsolved problems. Thus in this section, we will point out these problems and introduce some promising trends for future research. We hope that this survey not only provides a better understanding of long-tailed visual recognition for researchers but also stimulates future research activities.

\vspace{6pt}
\noindent\textbf{Large model with Large-scale Data}.
Large-scale Pre-trained Language Models (PLMs) have become the new paradigm for Natural Language Processing (NLP)~\cite{jacob2019bert, zhi2019xlnet, tom2020gpt3}. Large model with large-scale data has demonstrated strong performances on natural language understanding and generation with zero-shot and few-shot learning. In the field of visual recognition, model parameters and data scale are limited by the characteristics of visual tasks, which is relatively small compared with NLP, but some recent studies have begun to work in this direction~\cite{dosovitskiy2021vit, liu2021swin, carlos2021vitmoe}. Compared with the existing methods, large models and big data do not need to explicitly model the label frequency, but learn the general representation of images through a large amount of data, so as to solve the long-tailed problem.

\vspace{6pt}
\noindent\textbf{Long-tailed Adversarial Learning}.
Adversarial learning aims to deceive the model by providing deceptive input, the main research work can be simply divided into two parts: attack~\cite{alesander2018towards, gui2019model, yang2021attacks} and defense~\cite{zhong2019adversarial, wang2021augmax}. The research of adversarial learning has greatly promoted the safety and standardization of machine learning. However, with the blowout development of the defense mode and attack mode of the model, it also gradually presents a long-tailed distribution. Therefore, it seems to be a problem worth exploring to solve the tail classed in adversarial learning through the idea of long-tailed learning.

\vspace{6pt}
\noindent\textbf{Self-supervised Long-tailed Learning}.
Self-supervised learning methods regard each sample as an individual class, which can alleviate the label shifts problems and learn a relatively complete feature representation for all classes~\cite{he2020momentum, chen2020improved, chen2020simple}. However, there is less work that draws on self-supervised learning with long-tailed datasets, such as ~\cite{yang2020rethinking} uses self-supervised learning to improve the performance on long-tailed datasets by considering ignoring the value of labels. SSD~\cite{li2021self} uses the self-supervision guide feature learning method to improve the ability of the feature extractor. As work in the field of self-supervised has matured, there is great hope that self-supervision learning can surpass traditional supervised learning methods, thus making the learning of models more intelligent and automated. On this basis, the label bias problem for long-tailed datasets will hopefully be greatly improved.

\vspace{6pt}
\noindent\textbf{Vision-Language Long-tailed Learning}.
Vision-Language tasks require a model to understand the visual world and to ground natural language to the visual observations~\cite{antol2015vqa, anderson2016spice, jiang2020defense, zhang2021vinvl}. Vision-Language dataset contains two modes of annotation, and its long-tailed phenomenon is difficult to avoid~\cite{thomee2016yfcc100m, krishna2017genome}. Recently, CLIP~\cite{alec2021clip} proposes visual representation learning via natural language supervision in a similar contrastive learning setting~\cite{raia2006contra}, and shows amazing results on zero-shot and few-shot image classification. ViLD~\cite{gu2021vild} extends CLIP to zero-shot object detection task through knowledge distillation and prompt, and goes beyond the supervised learning method in the novel class of LVIS. These studies show that Vision-Language model can learn knowledge from multiple modes and improve the representation ability of few samples, which may be the next breakthrough of the long-tailed problem.

\vspace{6pt}
\noindent\textbf{More Task Settings}.
In addition to the well-known long-tailed tasks and some research directions proposed in this section, there are more long-tailed visual recognition tasks waiting to be mined. As analyzed in Sec.~\ref{sec:lt_pheno}, data naturally satisfy the long-tailed distribution in many fields, so solving the long-tailed problem may be able to improve the performance of models in these fields. However, according to the available research results, only a few work have studied from the perspective of the long-tailed distribution in their research field, such as long-tailed distribution of object classes in UAV images~\cite{yu2021towards}, long-tailed distribution of driving behavior in autonomous driving~\cite{narayanan2018semi} , and topic in the field of visual story telling~\cite{li2020topic}, content-related words for video captioning tasks~\cite{zhang2020object}, pose inclusion in datasets for 3D human pose estimation~\cite{zeng2020srnet}, dermatological categories in dermatological diagnosis~\cite{prabhu2018prototypical}, \etc We believe that for many research fields, the existing work to analyze and solve the long-tailed distribution problem is still not enough. For future researchers, the long-tailed problem can be taken into account to solve the problem of extreme data imbalance and thus improve the performance of the task.

\section{Conclusions}
\label{sec:concl}

In this survey, we comprehensively reviewed the long-tailed visual recognition according to the datasets, methods, long-tailed phenomenon and future directions. We provided the necessary background knowledge for readers, summarize the long-tailed studies into ten categories from the perspective of representational learning, and summarized the highlights and limitations of each category. We also compiled some generalized long-tailed datasets and benchmarked the results on 8 datasets. To study the long-tailed phenomenon extensively, we also conducted a structured survey of 20 widely-used datasets and found that the long-tailed phenomenon is widespread and that many areas' mainstream studies are not aware of it. Based on the analysis of the universality of the long-tailed phenomenon, we also gave the potential innovation and future research direction. We expect this survey to provide an effective way to understand current state-of-the-arts and speed up the development of this research field.

\section*{Acknowledgements}
This work was supported by the National Key Research and Development Program of China (Grant No. 2021YFF0500900).


%
%

\bibliographystyle{spmpsci}      
\bibliography{egbib}   


\end{document}